\documentclass[10pt,twocolumn,letterpaper]{article}

\usepackage[pagenumbers]{cvpr} %

\usepackage[dvipsnames]{xcolor}

\definecolor{cvprblue}{rgb}{0.21,0.49,0.74}
\usepackage[pagebackref,breaklinks,colorlinks,citecolor=cvprblue]{hyperref}
\usepackage{pifont}
\usepackage{multirow}
\usepackage{xcolor}
\usepackage{longtable}
\usepackage[toc,page]{appendix}

\definecolor{darkgreen}{RGB}{0,100,0}
\definecolor{darkred}{RGB}{139,0,0}
\definecolor{darkblue}{RGB}{0,0,139}

\title{Scaling Laws of Synthetic Images for Model Training ... for Now}

\author{
   Lijie Fan${^{1, \dagger, }}$\thanks{Work done while at Google.} \hspace{2mm} Kaifeng Chen$^2$\hspace{2mm} Dilip Krishnan$^2$\hspace{2mm} Dina Katabi$^1$\hspace{2mm} Phillip Isola$^1$\hspace{2mm} Yonglong Tian${^{2, \dagger}}$\\[3mm]
  $^1$MIT CSAIL, \hspace{3pt}
  $^2$Google Research, \hspace{3pt}
  $^\dagger$equal contribution\\[2mm]
  {\small \hspace{3pt} Github Repo: 
  \url{https://github.com/google-research/syn-rep-learn}}
}

\begin{document}
\maketitle
\begin{abstract}
Recent significant advances in text-to-image models unlock the possibility of training vision systems using synthetic images, potentially overcoming the difficulty of collecting curated data at scale. It is unclear, however, how these models behave at scale, as more synthetic data is added to the training set. In this paper we study the scaling laws of synthetic images generated by state of the art text-to-image models, for the training of supervised models: image classifiers with label supervision, and CLIP with language supervision. We identify several factors, including text prompts, classifier-free guidance scale, and types of text-to-image models, that significantly affect scaling behavior. After tuning these factors, we observe that synthetic images demonstrate a scaling trend similar to, but slightly less effective than, real images in CLIP training, while they significantly underperform in scaling when training supervised image classifiers. Our analysis indicates that the main reason for this underperformance is the inability of off-the-shelf text-to-image models to generate certain concepts, a limitation that significantly impairs the training of image classifiers. Our findings also suggest that scaling synthetic data can be particularly effective in scenarios such as: (1) when there is a limited supply of real images for a supervised problem (e.g., fewer than 0.5 million images in ImageNet), (2) when the evaluation dataset diverges significantly from the training data, indicating the out-of-distribution scenario, or (3) when synthetic data is used in conjunction with real images, as demonstrated in the training of CLIP models.

\end{abstract}    
\section{Introduction}
\label{sec:intro}

\begin{figure}[t]
\centering
\includegraphics[width=\linewidth]{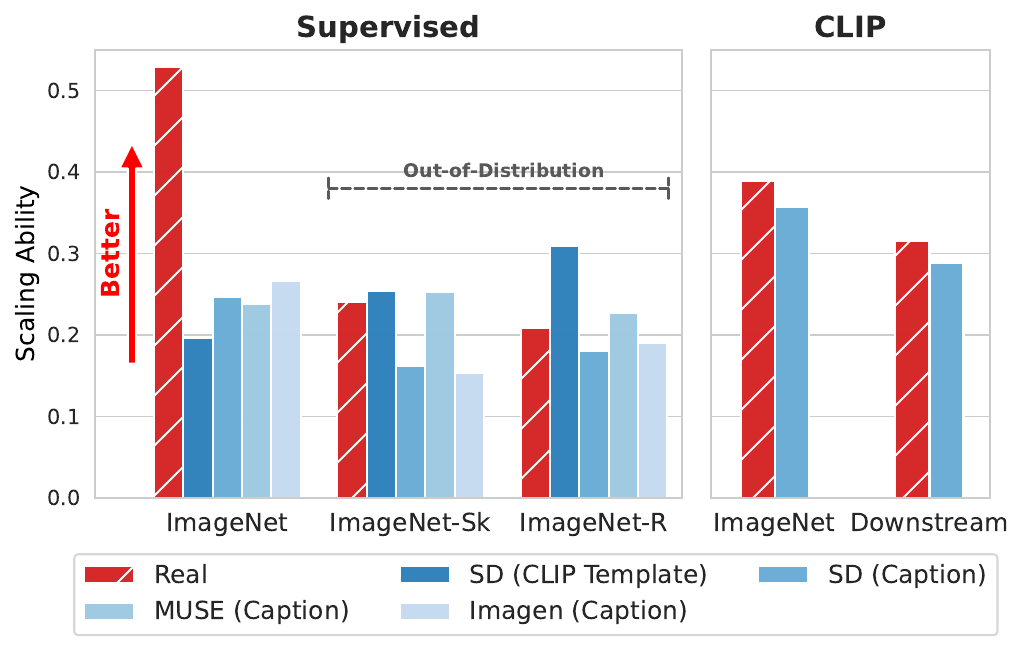}\\
\vspace{-2mm}
\caption{
\small Scaling ability (\ie, the slope of the power law curve between loss and dataset size fitted in the log space, see Eq.~\ref{equ:scaling}) comparison between real and synthetic images on supervised classifier and CLIP training. Red bars represent real images and blue bars represent synthetic images generated with different text-to-image models. Supervised models are trained on real or synthetic ImageNet, and text in parentheses is the text prompt used to generate the images (details in Section~\ref{sec:3factor}). ImageNet-Sketch and ImageNet-R are out-of-distribution tests. CLIP models are trained on LAION-400M with real or synthetic images. We see that: (1) scaling ability of synthetic data is \emph{slightly worse} than that of real data for CLIP training; (2) robustness on ImageNet-Sketch and ImageNet-R datasets can be \emph{better} when training on synthetic data.}
\label{fig:teaser}
\vspace{-1mm}
\end{figure}
The quality and quantity of data play a crucial role in training vision models. Historically, the emphasis has been on creating large, meticulously curated image datasets with categorical labels at the image level for training supervised models~\cite{krizhevsky2009learning,deng2009imagenet,sun2017revisiting,ridnik2021imagenet}. Prominent examples include CIFAR \citep{krizhevsky2009learning} and ImageNet \citep{deng2009imagenet}. While creating these datasets is effective on a smaller scale, their expansion to hundreds of millions of samples presents significant challenges. These challenges include the intensive labor required for curation at scale, as well as the increasing potential for noise and quality issues as the datasets scale up.

Recently, there has been an increasing interest in training vision models using language supervision~\cite{jia2021scaling,radford2021learning}. This shift is exemplified by models like CLIP \citep{radford2021learning}, which move beyond the fixed, predefined categories typical of datasets like ImageNet. Training these models requires extensive image-text pair datasets. Developments ranging from the creation of the Conceptual Captions dataset \cite{sharma2018conceptual}, which comprises millions of image-text pairs, to the LAION dataset \citep{schuhmann2022laion}, encompassing billions of pairs, are examples of this growing trend. 
However, this approach is not without its challenges. The massive scale of data sourcing, often through web scraping, introduces significant noise. Scalability issues also persist. Moreover, the immense size of these datasets presents practical difficulties in terms of storage and data transfer. 
For instance, LAION-2B requires tens of terabytes of disk space and could take days, if not weeks, to download.

Fortunately, recent breakthroughs in text-to-image models have introduced exciting new possibilities in the realm of synthetic data generation. These models, capable of producing high-quality images from textual descriptions, offer several significant advantages. Firstly, they allow precise control over image content through input texts, which could provide categorical labels or paired text supervision for free. Secondly, they are bandwidth-efficient, as only the model needs to be transferred, not the entire dataset. For instance, models like Stable Diffusion~\cite{rombach2022high} occupy merely 5 GB of disk space, which is 2000$\times$ more efficient compared to the massive LAION-2B dataset. Thirdly, they facilitate easier scalability with markedly reduced human labor for dataset curation. These benefits naturally lead to the question of whether it's feasible to scale up vision datasets with synthetic images for training supervised models.

However, the use of synthetic images is also not without its drawbacks. When scaled to tens or hundreds of millions of images, these models may produce images of lower quality or with misaligned concepts, and might also struggle with maintaining diversity.
In this paper, we tackle a pivotal question: \textit{How effective is the scaling of synthetic images, specifically generated for training supervised vision models?} We examine the scaling behavior of synthetic images created by cutting-edge text-to-image models, comparing their efficacy to real images in two key scenarios: the training of supervised classifiers and the training of vision models with language supervision, such as CLIP. Additionally, we explore a range of factors that markedly impact the scaling efficiency of synthetic images. These include the choice of text-to-image model, the classifier-free guidance scale employed, and the nature of text prompts used for generating training images. 
A summarized comparison of the scaling ability between real and synthetic images is shown in Figure~\ref{fig:teaser}.

We present our key findings as follows:

\begin{itemize}
\item An empirical study on the scaling behavior of images synthesized by three major text-to-image models (Stable Diffusion~\cite{rombach2022high}, Imagen~\cite{saharia2022photorealistic}, and Muse~\cite{chang2023muse}) shows that model performance exhibits power law scaling~\cite{kaplan2020scaling} as a function of the number of synthetic images they are trained on. This trend holds until computation budget and model size become limiting factors~\cite{kaplan2020scaling}.

\item We identify several factors that can significantly alter the scaling ability of synthetic data, including prompt design, classifier free guidance, and the choice of models.

\item In supervised settings, synthetic data does not scale as effectively as real data. However, there are exceptions where synthetic data demonstrates better scaling: (1) with classes that text-to-image models are particularly adept at generating, and (2) when the test data deviates significantly from the training data, \eg, out of distribution data.
\item In CLIP training, the disparity in scaling performance between synthetic and real data is less pronounced. Incorporating synthetic data with real data leads to enhanced zero-shot performance in most scenarios.
\end{itemize}

\section{Related Work}
\label{sec:related}

\noindent\textbf{Text to image models.}
Recent breakthroughs in text-to-image models, primarily driven by advances in diffusion models \citep{ho2020denoising, song2020score, yang2022diffusion}, have enabled the generation of high-quality, photo-realistic images using neural networks. Key examples of such models include Imagen \citep{saharia2022photorealistic}, which performs diffusion in pixel space, and Stable Diffusion \citep{rombach2022high}, which operates in the latent space of an autoencoder. DALL-E 3 \citep{betker2023improving} also exemplifies this category.
An alternative family of models, based on visual tokens, utilizes VQGAN \citep{van2017neural} and Transformers \citep{vaswani2017attention}. Prominent examples within this category include Parti \citep{yu2022scaling} and Muse \citep{chang2023muse}. Additionally, recent advancements have been exploring the scaling Generative Adversarial Networks (GANs) \citep{goodfellow2014generative} for text-to-image generation, as demonstrated in works such as \citep{kang2023scaling}.

\noindent\textbf{Learning from synthetic data.}
Synthetic data has proven to be effective in improving performance across various domains~\cite{tucker2020generating,dan2020generative,rosenberg2019speech,rossenbach2020generating,mimura2018leveraging,kumar2020data,taori2023alpaca,yang2020generative,he2022generate,meng2022generating}. Synthetic images, in particular, have been extensively utilized in a range of different computer vision tasks, including object detection~\cite{rozantsev2015rendering,peng2015learning}, semantic segmentation~\cite{chen2019learning,ros2016synthia}, autonomous driving~\cite{abu2018augmented}, and robotics~\cite{yen2022nerf,moreau2022lens}. More recently, there has been evidence that combining synthetic images generated by text-to-image models with real images can improve the performance on supervised learning tasks~\cite{azizi2023synthetic,he2022synthetic}. Particularly,~\cite{azizi2023synthetic} has fine-tuned the text-to-image model using the target dataset, \eg ImageNet, while this paper studies the capabilities of off-the-shelf text-to-image models. Additionally, there are efforts developing methods for learning transferable representations from synthetic images~\cite{ren2018cross,liu2022palm,baradad2021learning,jahanian2021generative,he2022synthetic,tian2023stablerep,sariyildiz2023fake}.

\noindent\textbf{Neural scaling laws.}
Scaling up model size, data amount, and training budget has unlocked new capabilities of deep models~\cite{chowdhery2022palm,gpt4,rae2021scaling,vit22b,vitscale}. Recent studies~\cite{hestness2017deep,kaplan2020scaling} suggest the testing loss behaves as a power low with respect to each of these three resources when the other two are proper, in large language models (LLMs), machine translation~\cite{gordon2021data}, auto-regressive generative models~\cite{henighan2020scaling}, and transfer learning~\cite{hernandez2021scaling}. Similar behavior is observed in multi-modal models~\cite{aghajanyan2023scaling}. Chinchilla~\cite{hoffmann2022training} suggests scaling up data proportionally to model size, to obtain compute-optimal LLMs. ~\cite{alabdulmohsin2022revisiting} propose to fit scaling laws by extrapolating training curves. Of particular interest, \cite{sorscher2022beyond} theoretically shows one can break the power law with respect to data size with an ideal data pruning strategy. In this paper, we focus on the scaling behavior of synthetic data for training models.

\section{Preliminaries}
\label{sec:preliminaries}

We first study the scaling behavior of synthetic images generated with state-of-the-art text-to-image models under the ImageNet supervised training setting. 
\subsection{Three Factors on T2I Generation}
\label{sec:3factor}
There are three primary factors influencing the generated images used for supervised training: (1) choice of text-to-image model, (2) the classifier-free guidance scale, and (3) the class-specific prompt used for the text input. We will now provide a detailed description of each of these factors:

\noindent\textbf{Text-to-Image Models.} We conducted the study on three state-of-the-art text-to-image models of different types:
\begin{itemize}
    \item \textit{Stable Diffusion~\cite{rombach2022high}}, a  model that drives the diffusion process in the latent space of a pre-trained autoencoder.
    \item \textit{Imagen~\cite{saharia2022photorealistic}}, a model that drives the diffusion process directly in the raw pixel space.
    \item \textit{Muse~\cite{chang2023muse}}, a visual token-based generation model trained with masked generative modeling, that performs discrete diffusion in the latent space of an autoencoder. 
\end{itemize}
These models have distinct architectural designs, but are all capable of generating photo-realistic images. Since Imagen~\cite{saharia2022photorealistic} and Muse~\cite{chang2023muse} are not publicly available, we base our work on a version trained on internal data sources.

\noindent\textbf{Guidance Scale.}
All modern text-to-image models primarily rely on the classifier-free guidance (CFG) technique to generate images based on textual input \citep{ho2022classifier}. Increasing the CFG scale typically improves the alignment between the generated images and the input text, resulting in higher-quality output images. However, this also tends to reduce the diversity of content in the generated images.
Through empirical analysis, we determined that when generating images for training supervised classifiers, it is advisable to use a relatively \textbf{\textit{lower}} CFG scale compared to the default value used in generation. This ensures that the generated images exhibit a higher degree of diversity, particularly when generating images from texts describing the same class.
We conducted a detailed analysis and determined the optimal CFG scale ranges for different models: [1.5, 10.0] for Stable Diffusion, [1.0, 2.0] for Imagen, and [0.1, 1.0] for Muse.

\noindent\textbf{Class-specific Prompts.}\label{sec:prompts}
To generate images for each class in ImageNet, we employed different techniques to create corresponding text prompts. This allows us to generate images conditioned on the specific ImageNet class via the prompts. Take the class `Tench' as example, we can have prompts as:
\begin{itemize}
    \item \textit{Classnames:} Directly use the ImageNet class name. (`Tench')
    \item \textit{Classnames + Description:} Combine class name with its WordNet \citep{miller1995wordnet} description. (`tench, freshwater dace-like game fish of Europe and western Asia ...')
    \item \textit{Classnames + Hypernyms:} Combine ImageNet class name with its Wordnet hypernyms. (`Tench, Tinca tinca, cyprinid, cyprinid fish')
    \item \textit{Word2Sen:} Use a pre-trained T5 model~\cite{raffel2020exploring} as used in~\cite{he2022synthetic} to convert the ImageNet class name into a sentence. We generate 100 sentences for each class. (`a tench with fish in the distance.')
    \item \textit{CLIP templates:} Generate either 7 or 80 sentences with the text templates CLIP used for zero-shot classification task. (`a photo of the large tench')
    \item \textit{IN-Captions:} Combine the class name with captions from ImageNet(IN) training images. Captions are generated by BLIP2~\cite{li2023blip}.  (`Tench, a man holding a fish')
\end{itemize}

\begin{table*}[t]
\begin{tabular}{cc}
\begin{minipage}[t]{0.36\linewidth}
\centering
\includegraphics[width=\linewidth]{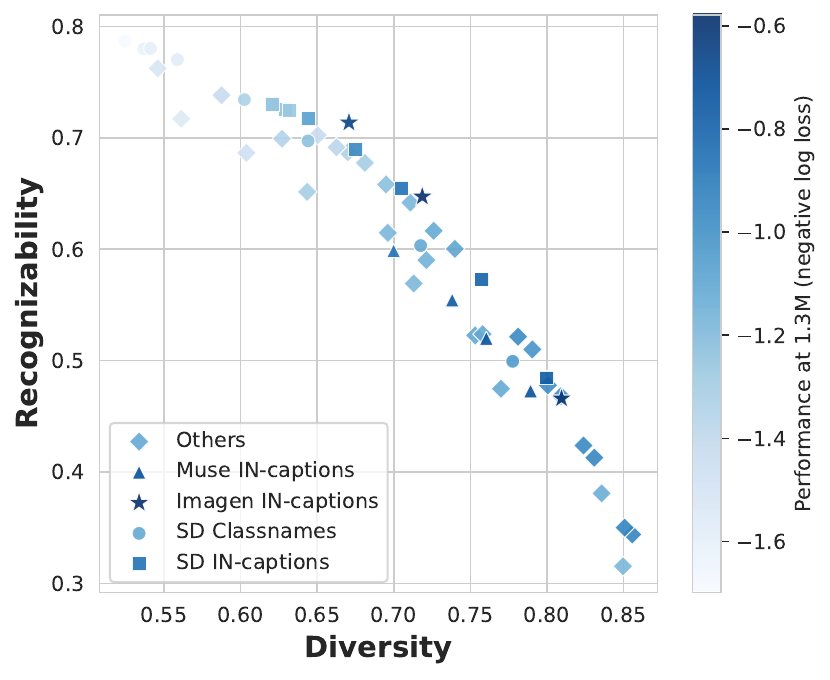}
\captionof{figure}{
\small
Recognizability vs. diversity plot for various synthetic image generation configurations (as in Section~\ref{sec:1.3m_performance}), colored by the performance at 1.3M on ImageNet validation set (measured by negative log loss). Deeper color stands for smaller loss and better performance.}
\label{fig:rec_div}
\end{minipage}
&
\begin{minipage}[t]{0.59\linewidth}
\centering
\includegraphics[width=\linewidth]{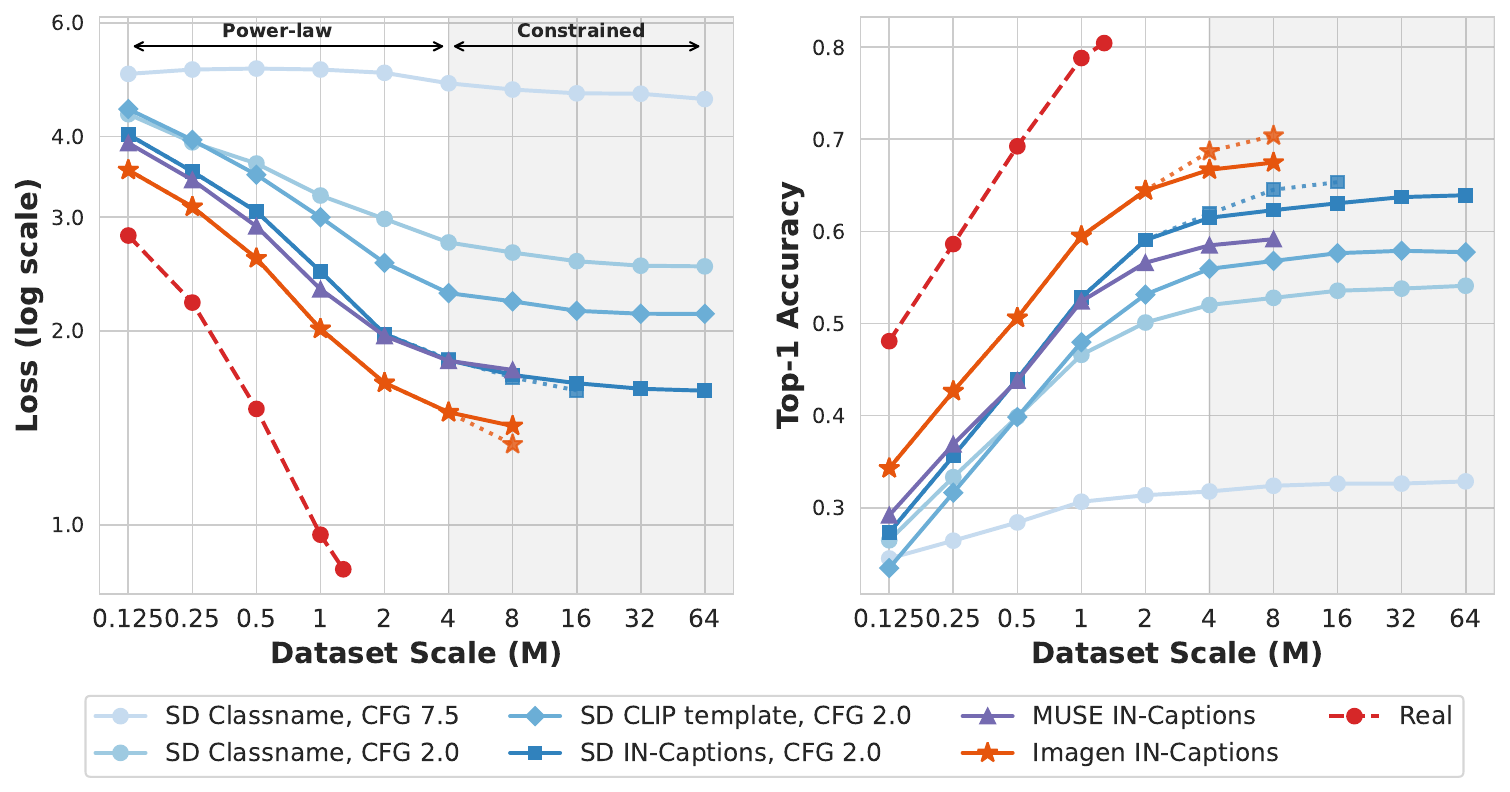}
\captionof{figure}{
\small
Scaling on ImageNet validation set for various configurations as in Section~\ref{sec:scaling_ability}. Loss and data scale follows the power-law (as in Equation~\ref{equ:scaling}) with varied $k$ when data is less than 4M. By tuning the CFG scale, text prompts and text-to-image models, the scaling behavior for synthetic images can be significantly improved (from light blue to orange). Red dashed line is for real images. Orange and blue dotted lines are ViT-L backbones, extending the power-law to 8M.}
\label{fig:all_scaling}
\end{minipage}

\end{tabular}
\end{table*}

\subsection{Metrics: Recognizability and Diversity}
\label{sec:metric-rec-div}
The above factors give us a number of configurations to generate synthetic data. We now proceed to define metrics to analyze the resulting images, and then analyze the scaling behavior exhibited by the images generated under this configuration. The generated images should possess two crucial attributes: (1) Recognizabilty: Synthetic images should exhibit high precision, meaning they correctly represent the intended class, and high recall, implying that images for other classes should not mistakenly contain elements of this class. (2) Diversity: It is essential that the generated images are diverse from each other to improve generalization. 

We define two measures to quantify the recognizability and diversity of images generated under a specific configuration. We generate 50 images for each ImageNet class, resulting in a synthetic test set comprising 50,000 images. Subsequently, we define the two metrics as follows:
\begin{itemize}
    \item \textit{Recognizabiliy:} Use a pre-trained ImageNet classifier (a ViT-B with 86.2$\%$ accuracy from~\cite{rw2019timm}) to classify the generated images and compute the F1 score for each class. The final metric is given by averaging F1 score across all classes.
    \item \textit{Diversity\footnote{We also tried replacing the diversity metric with FID~\cite{heusel2017gans} or LPIPS~\cite{zhang2018unreasonable}, please refer to Appendix~\ref{sec:appendix-fid} for details.}:} Following~\cite{boutin2023diffusion}, we extract features from the same pre-trained model~\cite{rw2019timm} and compute the standard deviation on the feature space for images from every class, and then compute the average score across all classes.
\end{itemize}

\subsection{Scaling Law for Synthetic Data}
Prior works on scaling laws, such as ~\cite{kaplan2020scaling}, have observed that, for \emph{sufficiently large} models, the test loss $L$ and dataset scale $D$, approximately follow a power-law relationship:
\begin{equation}
    L_D \propto (1/D)^k
\label{equ:powerlaw}
\end{equation}
where $k$ is a constant. Thus $L_D$ exhibits linear dependence on $D$ in log space. Let $D_I$ be 1.3 million, roughly the size of the ImageNet training set with real data. We re-write Equation \ref{equ:powerlaw} as:
\begin{equation}
\label{equ:scaling}
    \log L_D = \underbrace{- k}_{\text{${k}$: Scaling Ability}}(\log D- \log D_I) - \underbrace{(-\log L_{D_I})}_\text{Performance at 1.3M}
\end{equation}
The slope $-k$ and y-intercept $-\log L_{D_I}$ would determine a unique scaling curve in log space. With this, we provide quantitative definitions for two key metrics for scaling:
\begin{itemize}
    \item \textbf{Scaling Ability}: Quantifies the scaling effectiveness of synthetic images generated by a particular text-to-image configuration. Stepper curves means loss scales better with data, therefore we represent scaling ability by the negative of the \textbf{slope}: $k$.
    \item \textbf{Performance at 1.3M}: Measures the classification performance of models (as negative log loss) when trained on a dataset with a scale equivalent to 1.3M, the size of the ImageNet training set. It is represented by the \textbf{y-intercept} $-\log L_{D_I}$.
\end{itemize}

\section{Scaling on Supervised Training}

\subsection{Setup}
We train supervised classification models exclusively using the images generated by text-to-image models and evaluate their performance by computing cross-entropy loss and top-1 accuracy on the ImageNet validation set, which contains real images. Training iterations are scheduled linearly based on the training data size in logarithmic space. All generated images are resized to a resolution of 256x256 pixels. Unless stated otherwise, we employ the ViT-B model \citep{dosovitskiy2020image} with a patch size of 16 as our backbone architecture. Training hyperparameters details are provided in Appendix~\ref{sec:implementation_details}.

\subsection{Performance at 1.3M}
\label{sec:1.3m_performance}

\begin{figure*}[t]
\centering
\includegraphics[width=1.0\linewidth]{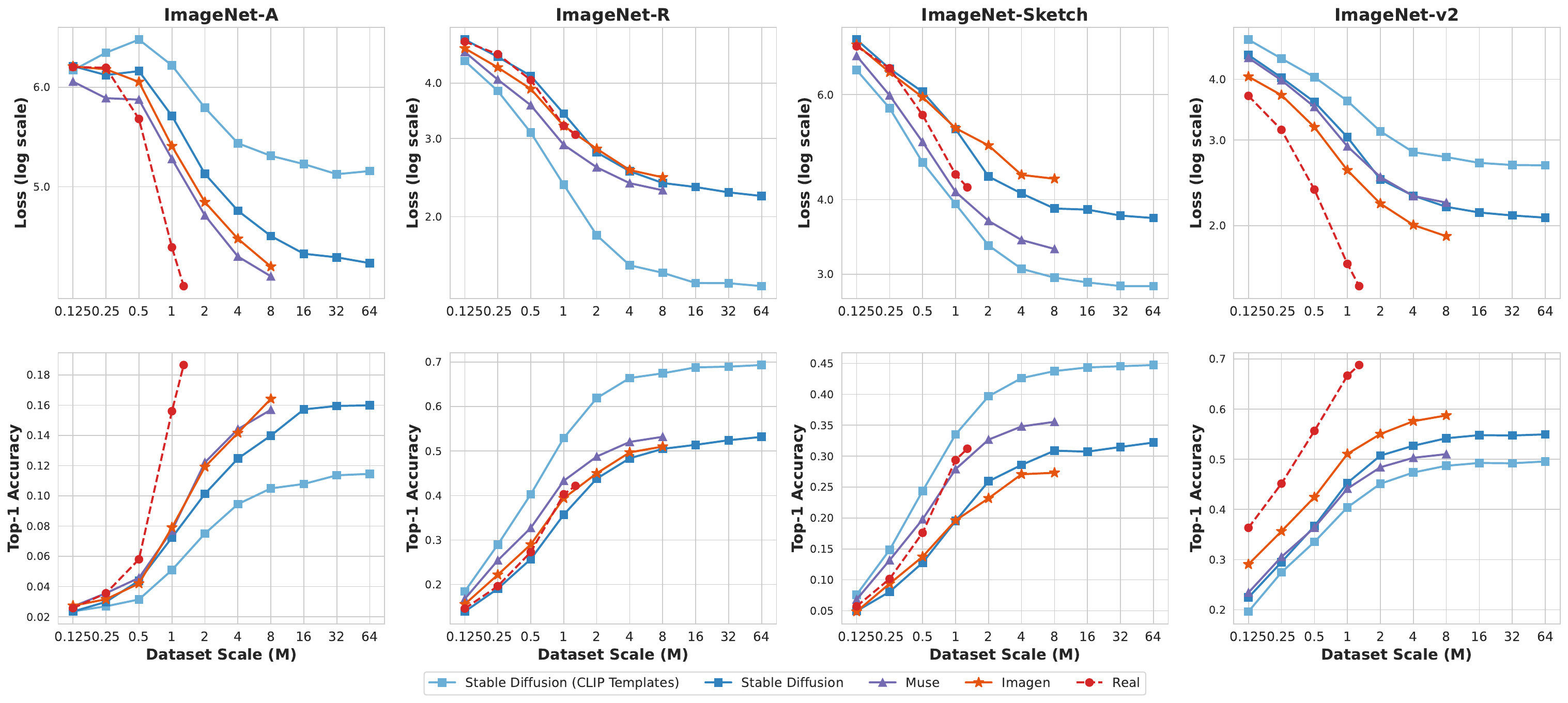}
\caption{
\small
Scaling behavior on four different out-of-distribution validation sets. We compare synthetic images generated with optimal CFG scales by Stable Diffusion (with 80 CLIP templates or IN-Captions prompt), Imagen and Muse (all with IN-Caption prompt) with real images. Scaling synthetic data is useful and can surpass real images when the domain gap between the training and testing is significant, e.g. when evaluated on ImageNet-R and ImageNet-Sketch.}
\label{fig:ood_scaling}
\vspace*{-1mm}
\end{figure*}

We commenced by generating synthetic ImageNet datasets, each containing 1.3 million synthetic images, using various configurations of text-to-image models, CFG scales, and prompts as outlined in Section \ref{sec:preliminaries}. In total, we created synthetic ImageNets in 54 distinct configurations, with detailed information provided in Appendix~\ref{sec:appendix-1.3m}. Figure~\ref{fig:rec_div} displays the validation loss on the real ImageNet validation set, represented by $-\log L_{D_I}$ as defined in Equation~\ref{equ:scaling}. A higher value correlates with increased classification accuracy, signaling better performance. Comparisons focusing on classification accuracy are also included in Appendix~\ref{sec:appendix-1.3m}.

Within this study, we investigate the impact of different prompt sets (Section \ref{sec:preliminaries}) within the Stable Diffusion model. {Squares} (\ding{110}), {Circles} (\ding{108}), and {Diamonds} (\ding{117})  represent prompt configurations involving IN-captions, Classnames, and all other prompt setups, respectively. 
For Muse and Imagen configurations, we maintain the prompt set as IN-Captions and vary the CFG scale within the ranges [1, 2] and [0.1, 1], respectively. {Triangles} (\ding{115}) represent the performance of images generated with Muse, while {Stars} (\ding{72}) represent the performance of images generated with Imagen. Several key findings emerge from the results:

    \noindent \textbf{Diversity and Recognizability trade-off}: Across different configurations, we observe a trade-off between diversity and recognizability. The top-right corner of the figure represents the best performance, indicating configurations that can generate both accurate and diverse images. Configurations perform poorly when either recognizability or diversity falls below a certain threshold. Also see Figure~\ref{fig:appendix_rec_all} bottom left for scattering colorized by accuracy.
    
    \noindent \textbf{Effect of Prompt Sets: }Choosing different prompt sets can impact performance. Using a more diverse prompt set shifts the configuration towards the bottom-right of the figure. Transitioning from Classname to IN-captions for text-to-image prompts may contribute to this shift, likely due to the increased diversity on the text side, which inherently leads to more diverse generated images.

    \noindent \textbf{Impact of CFG Scale:} When prompts are fixed, controlling the CFG scale also affects the performance of the classification model. Increasing the CFG scale shifts the configuration towards the upper-left part of the figure, where recognizability is increased, but diversity decreases. This initially leads to improved performance, followed by a decrease.
    
    \noindent \textbf{Text-to-Image Model Performance:} In terms of text-to-image models, 
    when all configured to use IN-Captions as prompts, Stable Diffusion, Imagen, and Muse demonstrate a quite similar trend in balancing recognizability and diversity. This similarity in their trade-off is reflected in their close proximity to each other in the plot.

\subsection{Scaling Ability}
\label{sec:scaling_ability}

We next proceed to analyze the scaling behavior of different models, as well as the difference between training supervised models on synthetic images and on the real ImageNet training set. Figure~\ref{fig:all_scaling} illustrates the scaling behavior across various configurations. Specifically, for Stable Diffusion, we depict the scaling behavior of different configurations with various prompts and CFG scales. We select the optimal configuration for Muse and Imagen from Section~\ref{sec:1.3m_performance}, using IN-Caption as prompts and the corresponding optimal CFG scale for each model.
From the figure, several observations can be made:

\noindent\textbf{Power-law Relationship:} Training on synthetic images follows a power-law relationship from 0.125 million to 4 million training images. Validation loss and training data size exhibit a linear correlation when analyzed in log space.

\noindent\textbf{Scaling Disparity:} Training on synthetic images does not scale as effectively as training on the real ImageNet training set images, and typically has a smaller scaling ability. This difference can be attributed to the curation of ImageNet training images and performing validation under an in-domain setting.

\noindent\textbf{Impact of Prompts and CFG Scale:} Using default prompt sets and CFG scale for image generation results in poor scaling ability, i.e. a very flat slope and smaller $k$ value. However, by tuning the prompts and CFG scale properly, the generated images become much more diverse, leading to an increased scaling behavior for synthetic images, bringing it closer to the scaling ability observed with real images. Nevertheless, the best scaling configuration is still significantly worse than scaling with real data.

\subsection{Scaling beyond 4M}
We naturally wonder about the scaling behavior when the dataset size exceeds 4 million images and whether the validation loss will continue to decrease. In Figure~\ref{fig:all_scaling}, we also illustrate the scaling curve for Stable Diffusion up to 64 million images, and for Muse and Imagen up to 8 million training images (in gray background). The results indicate that the relationship changes when the dataset scale exceeds around 4 million images.

We hypothesize that this could be due to the loss being constrained by insufficient model capacity. According to~\cite{kaplan2020scaling}, the power-law relationship between validation loss and training dataset size requires the model to have sufficient capacity to fit the dataset and converge. Therefore, when the dataset size exceeds 4 million images, and if we continue to use ViT-B as the backbone architecture, the validation loss in log space no longer exhibits a linear trend.
To address this, we retrain the supervised models with ViT-L as the backbone architecture for the best Stable Diffusion and Imagen configuration, as shown in the dotted lines. This improvement in model capacity could achieve a lower validation loss and maintain a roughly linear ratio up to the 8 million scale and slightly postpones the inflection point. 

\subsection{Out-Of-Distribution Scaling}
We also investigate the scaling behavior on out-of-distribution (OOD) validation sets to determine whether it differs from the in-domain setup on the ImageNet validation set. We employ the supervised ImageNet classifiers and test them on four OOD validation sets, which include 
ImageNet-A~\cite{hendrycks2021nae}, ImageNet-R~\cite{hendrycks2021many}, ImageNet-Sketch~\cite{wang2019learning}, and Imagenet-V2~\cite{recht2019imagenet}.
The scaling curves for validation loss and top-1 validation accuracy are presented in Figure~\ref{fig:ood_scaling}.

Our empirical results indicate that in scenarios where the domain gap is relatively small, such as testing on ImageNet-v2, the scaling behavior mirrors the observation in in-domain setups, with real images showing superior scaling performance. However, a much more intriguing observation emerges when the domain shift is more pronounced, as seen in tests on ImageNet-R and ImageNet-Sketch. In these instances, the disparity in scaling capabilities between synthetic and real images narrows. Consequently, scaling up synthetic images becomes particularly beneficial and useful. Remarkably, in situations with sufficiently large dataset scales, synthetic images can even outperform real images from ImageNet training set (e.g. for ImageNet-R and ImageNet-Sketch with Muse), highlighting the potential of synthetic images in bridging significant domain gaps. Interestingly, when images are generated with 80 CLIP templates as text prompt (the light blue line in the plot) instead of IN-Captions, the improvements over real images on ImageNet-R and ImageNet-Sketch are more significant, although the scaling ability on the ImageNet validation set is worse (as shown in Figure~\ref{fig:all_scaling}). This suggests that carefully crafting text prompts can unlock further potential in increasing the efficacy of synthetic images, particularly for OOD scenarios.

\begin{figure}[t]
\centering
\includegraphics[width=1.0\linewidth]{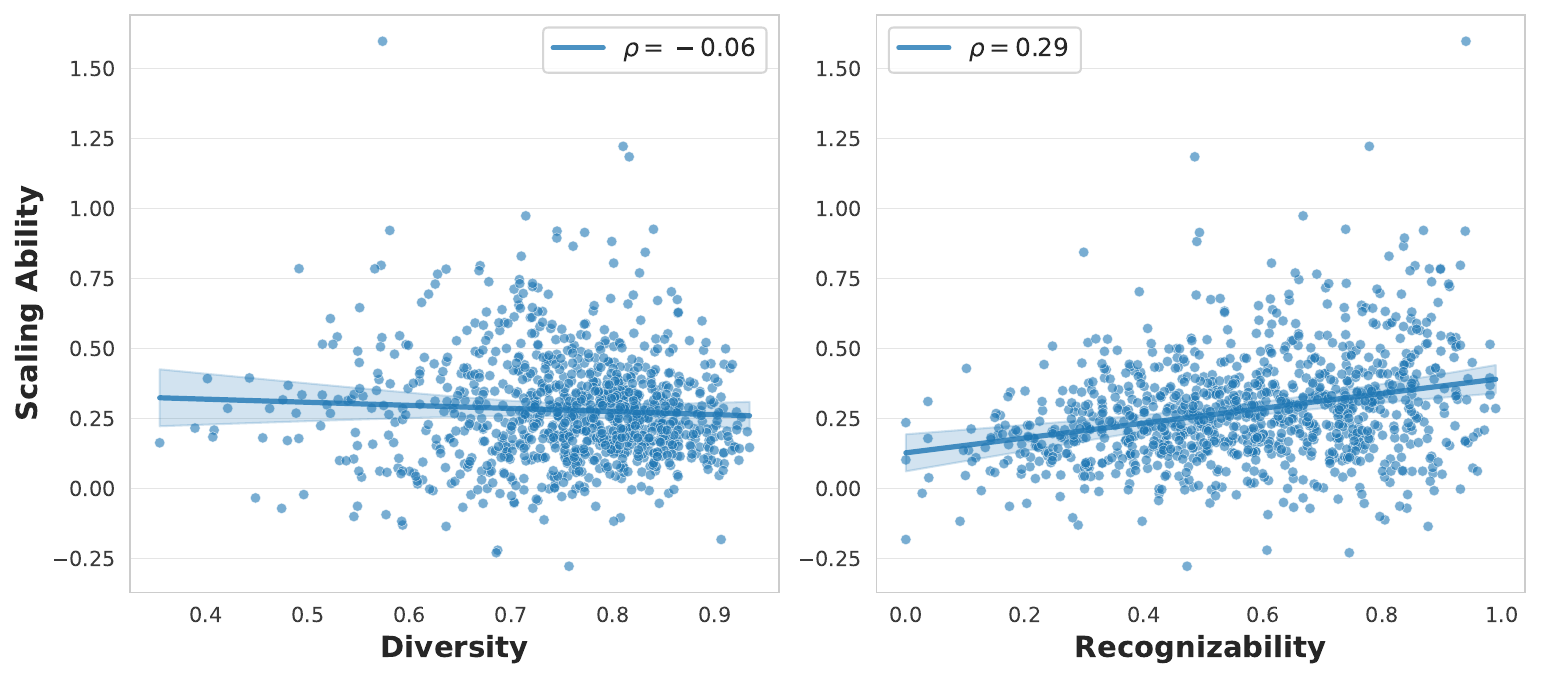}
\caption{
\small
Per class analysis on the relationship between scaling ability (defined as $k$ in Equation~\ref{equ:scaling}) and both diversity and recognizability. Within each specific class, the plots indicate a positive correlation between recognizability and scaling ability. However, the correlation between diversity and scaling ability appears to be negligible.}
\label{fig:scaling_div_rec}
\end{figure}
\begin{figure*}[h]
\centering

\includegraphics[width=0.98\linewidth]{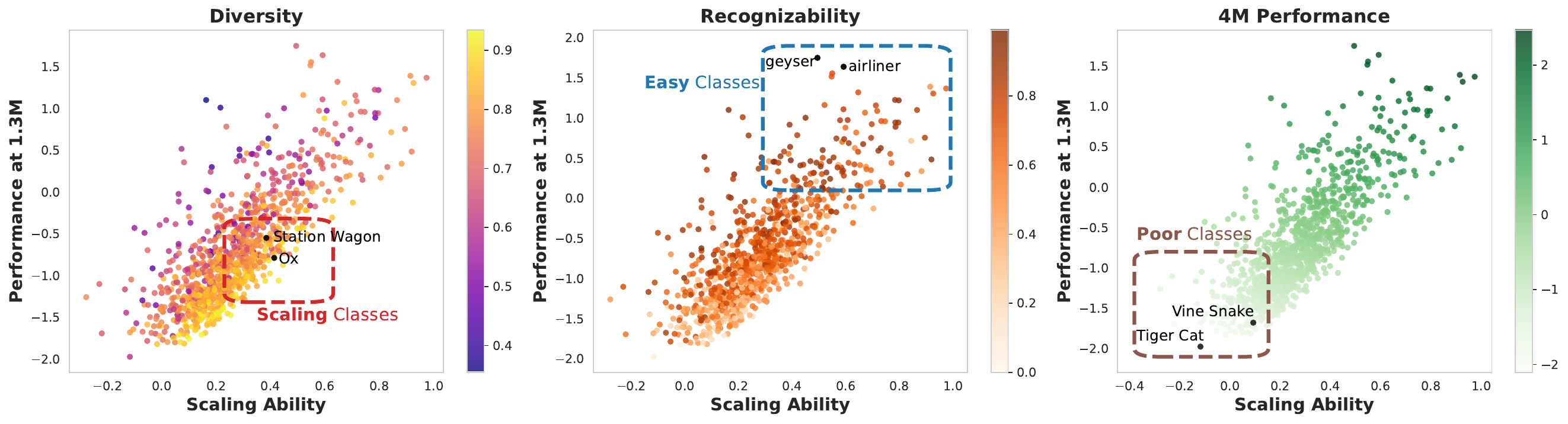}
\caption{
\small
Scaling ability vs. Performance at 1.3M plot for synthetic data. Each point represents one of the 1,000 ImageNet classes. Classes are colored by their diversity, recognizability, and their final performance at 4M scale in the three sub-figures respectively. The scaling ability is measured by $k$ defined in Equation~\ref{equ:scaling}. The performances at Y-axis is measured by the validation loss: $-\log(L)$, and higher numbers indicate lower loss and better performance. We choose two classes in each of the `Scaling', `Easy' and `Poor' class categories, and their detailed scaling behavior and visualization can be found in Figure~\ref{fig:visualization}.
}
\label{fig:per_class_distinction}
\vspace*{4mm}
\includegraphics[width=\linewidth]{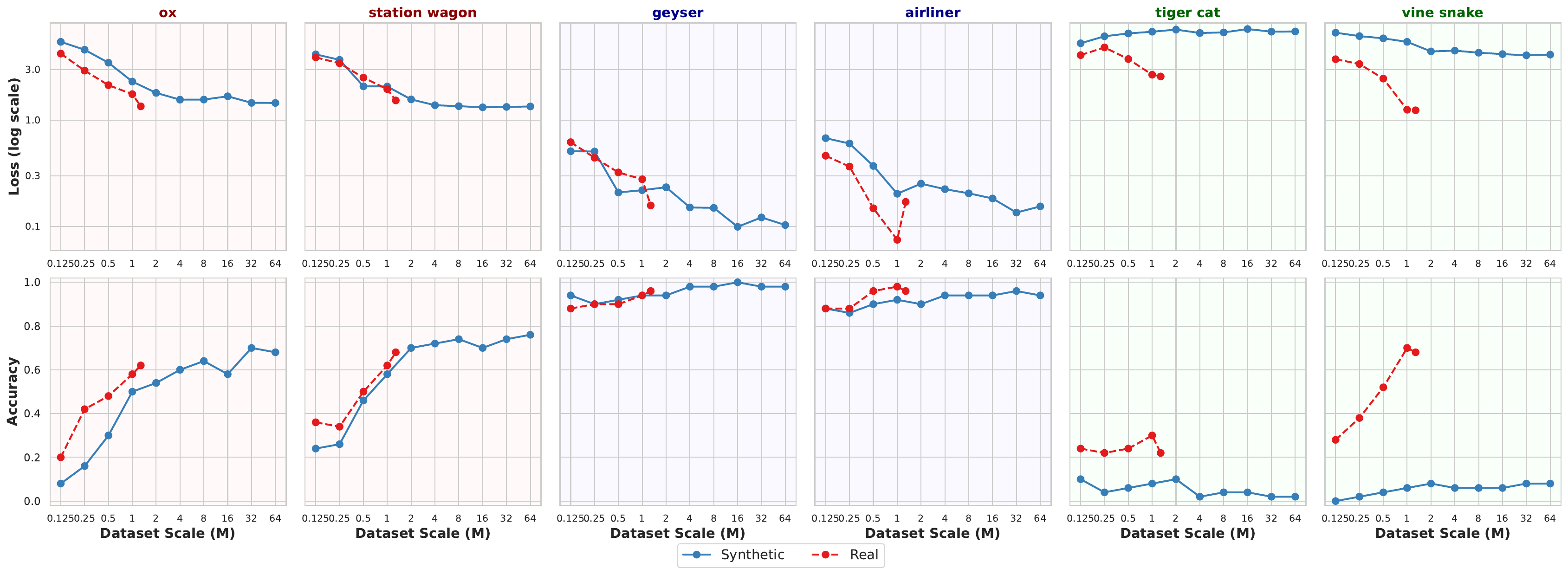}\\
\vspace*{-4mm}
\caption{
\small
Scaling behavior for classes from `Scaling (\textcolor[rgb]{0.77,0.2,0.22}{Red})', `Easy (\textcolor[rgb]{0.22,0.47,0.69}{Blue})' and `Poor (\textcolor[rgb]{0.32,0.65,0.24}{Green})' categories. Easy classes have a good initial accuracy with limited amounts of data, while Poor classes do not scale well. Scaling classes scale the best, and can achieve better performances than real images as the data amount goes up.
}
\label{fig:per_class_loss_acc}
\vspace*{4mm}
\includegraphics[width=\linewidth]{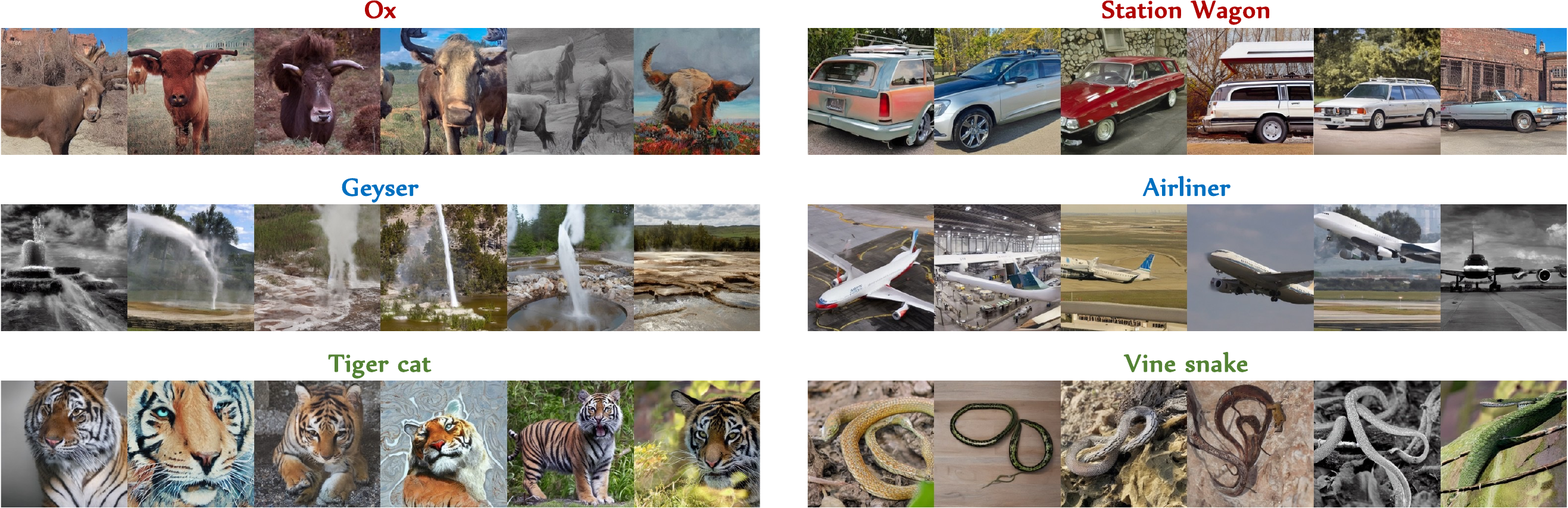}\\
\vspace*{-2mm}
  \caption{\small{Visualization of different class categories generated by Stable Diffusion. Top row are the `Scaling' classes that scales well. Middle row are the `Easy' classes that has a good initial performance. Bottom row are the `Poor' classes that has poor scaling ability and performance.}}
  \label{fig:visualization}

\end{figure*}

\section{Zoom-in: Per Class Analysis}
\label{sec:perclass}
In addition to our general analysis of scaling behavior and its impact on overall performance across all classes, we take a more detailed approach to gain a better understanding of the key factors affecting class-specific scaling behavior. Our analysis involves assessing the scaling ability of each specific class in the 1,000 categories in ImageNet. We aim to establish connections between the scaling ability of each class and the characteristics of the images generated by text-to-image models, aiming to figure out potential reasons why synthetic data does not scale as good as real images.
For this analysis, we focus on images generated by Stable Diffusion, using the optimal CFG of 2.0 and IN-Caption prompts.

\subsection{What affects Scaling Ability}
\label{sec:affect-scaling-ability}
When we fix the generation configuration of specific text-to-image model, CFG scales and prompts as described in Section~\ref{sec:3factor},
text-to-image models could still exhibit varying degrees of recognizability and diversity when generating images for different object classes. To explore how these factors influence the scaling ability of each class,  we conducted an analysis focusing on the correlation between scaling ability and both recognizability and diversity, all computed for each class individually. These correlations, and their implications for scaling efficiency, are depicted in Figure~\ref{fig:scaling_div_rec}.
Our analysis underscores the potential positive role of recognizability in determining the scaling ability for the synthetic images, for each specific class, generated by text-to-image models. We identified a positive correlation between recognizability and scaling ability, indicating that the precision in generating the intended class significantly enhances the scaling effectiveness of synthetic images. In contrast, the influence of diversity within each class seems to be more limited. Our findings reveal only a negligible correlation between diversity and scaling ability. This might be attributed to the increased noise introduced when computing diversity for specific categories, as opposed to the overall dataset.

\subsection{What makes a `Poor' class}
\label{sec:poor-class}

To gain a deeper insight into how scaling ability is distributed across different classes, and identify classes that do not scale well, we created a scatter plot with scaling ability on the X-axis and 1.3M performance on the Y-axis, as shown in Figure~\ref{fig:per_class_distinction}. Each class is represented as a dot in this plot, with top-right positions indicating better performance when scaled up to 4 million images. Points are colored based on either diversity or recognizability, or their final performance at 4M dataset scale.

Based on their positioning in the scatter plot, classes can be categorized into three distinct groups. Points in the bottom-left section represent `Poor' classes, which are marked by both limited scaling ability and poor overall performance. In contrast, classes located in the upper-right section are deemed `Easy' characterized by strong initial performance as well as robust scaling ability. Lastly, classes situated in the mid-right section can be described as `Scaling'. These classes may exhibit poor initial performance but demonstrate considerable improvement as the dataset size increases.

In Figure~\ref{fig:per_class_loss_acc}, we showcase two classes from each of the `Scaling', `Easy', and `Poor' categories to illustrate and compare their scaling behaviors against real images within the same class. This analysis highlights an intriguing finding: certain `Scaling' classes demonstrate a scaling ability that surpasses that of real images, thereby emphasizing the potential utility of synthetic images in these scenarios. We present additional results for `Scaling' classes in Appendix~\ref{sec:appendix-scaling-classes}.

Additionally, we present visualizations of the generated images from these categories in Figure~\ref{fig:visualization}. Our findings show that text-to-image models adeptly generate images for `Scaling' and `Easy' classes with commendable accuracy and diversity. However, these models face challenges in accurately rendering the correct concepts for `Poor' classes.

\section{Scaling on CLIP}
\begin{table*}[t]
\begin{center}
\caption{\small{Zero-shot transfer performance on 15 downstream datasets. Models are trained on LAION-400M subsets at various scales from 1M to the total 371M, with images from synthetic, real or synthetic+real. Combining synthetic images with real images can improve performance, especially when data amount is limited.
}}
\vspace*{-2mm}
\label{table:clip_result}
\resizebox{.95\textwidth}{!}{
\begin{tabular}
{ccccccccccccccccc@{\hspace{1.0em}}|@{\hspace{1.0em}}c@{\hspace{1.0em}}}
\toprule[1.2pt]
\bf Scale&\bf Data&
\rotatebox[origin=lb]{90}{\smash{\small Food-101}} & \rotatebox[origin=lb]{90}{\smash{\small CIFAR-10}} & \rotatebox[origin=lb]{90}{\smash{\small CIFAR-100}} & \rotatebox[origin=lb]{90}{\smash{\small SUN397}} &
\rotatebox[origin=lb]{90}{\smash{\small Cars}} & \rotatebox[origin=lb]{90}{\smash{\small Aircraft}} & \rotatebox[origin=lb]{90}{\smash{\small DTD}} & \rotatebox[origin=lb]{90}{\smash{\small Pets}} & \rotatebox[origin=lb]{90}{\smash{\small Caltech-101}} &
\rotatebox[origin=lb]{90}{\smash{\small Flowers}} & \rotatebox[origin=lb]{90}{\smash{\small STL-10}} & \rotatebox[origin=lb]{90}{\smash{\small EuroSAT}} &
\rotatebox[origin=lb]{90}{\smash{\small RESISC45}} & \rotatebox[origin=lb]{90}{\smash{\small GTSRB}} & \rotatebox[origin=lb]{90}{\smash{\small Country211}}  & \rotatebox[origin=lb]{90}{\smash{\small \bf Average}} \\
\midrule
\multirow{3}{1.3em}{\rotatebox[origin=c]{0}{\small{1M}}}   & \small Syn & 5.2 & 12.8 & 3.3 & 5.9 & 1.7 & 0.9 & 5.5 & 6.7 & 17.8 & 3.5 & 29.4 & 9.0 & 9.7 & 5.4 & 1.2 & 7.9  \\
 & \small Real & 5.2 & 25.4 & 7.6 & 5.0 & 2.1 & 1.0 & 5.4 & 5.4 & 18.0 & 5.0 & 36.4 & 14.7 & 9.3 & 6.6 & 1.0 & 9.9 \\
 &\small Syn+Real & \textbf{10.9} & \textbf{32.2} & \textbf{13.0} & \textbf{13.1} & \textbf{4.6} & \textbf{1.4} & \textbf{9.4} & \textbf{12.0} & \textbf{36.0} & \textbf{8.9} & \textbf{62.5} & \textbf{19.8} & \textbf{14.7} & \textbf{7.5} & \textbf{1.9} & \textbf{16.5} \\ \midrule

\multirow{3}{1.3em}{\rotatebox[origin=c]{0}{\small{2M}}}   & \small Syn &11.0 & 15.3 & 3.8 & 14.5 & 6.2 & 1.7 & 10.3 & 15.6 & 36.2 & 7.2 & 36.3 & 15.6 & 14.5 & 3.4 & 1.7 & 12.9  \\
 & \small Real & 13.4 & 39.0 & 16.8 & 13 & 6.6 & 1.3 & 10.5 & 13.0 & 40.1 & 12.4 & 57.1 & \textbf{17.0} & 14.9 & \textbf{6.5} & 1.7 & 17.6  \\
 &\small Syn+Real & \textbf{22.2} & \textbf{59.7} & \textbf{27.0} & \textbf{23.4} & \textbf{18.2} & \textbf{2.1} & \textbf{15.4} & \textbf{24.8} & \textbf{55.7} & \textbf{13.1} & \textbf{73.9} & \textbf{22.4} & \textbf{20.6} & \textbf{4.2} & \textbf{3.0} & \textbf{25.7}\\ \midrule
 
 \multirow{3}{1.3em}{\rotatebox[origin=c]{0}{\small{4M}}}   & \small Syn & 19.6 & 19.7 & 7.1 & 23.0 & 22.4 & 2.1 & 17.0 & 30.1 & 53.4 & 13.9 & 64.4 & 12.8 & 21.1 & 5.3 & 3.1 & 21.0  \\
 & \small Real & 30.8 & 63.8 & 33.4 & 26.2 & 27.7 & 1.8 & 18.8 & 33.9 & 61.7 & 18.9 & 79.2 & \textbf{40.2} & 21.5 & 10.0 & 3.4 & 31.4 \\
 &\small Syn+Real & \textbf{40.3} & \textbf{67.1} & \textbf{39.3} & \textbf{35.8} & \textbf{40.9} & \textbf{2.3} & \textbf{22.9} & \textbf{45.4} & \textbf{70.1} & \textbf{23.1} & \textbf{88.2} & 33.5 & \textbf{27.7} & \textbf{12.3} & \textbf{4.5} & \textbf{36.9}
 \\ \midrule
 \multirow{3}{1.3em}{\rotatebox[origin=c]{0}{\small{8M}}} &\small Syn & 34.2 & 23.8 & 9.5 & 32.6 & 39.9 & 3.5 & 20.3 & 46.3 & 63.0 & 20.7 & 78.7 & 9.8 & 19.1 & 4.9 & 4.3 & 27.4 \\
 &\small Real & 48.7 & 79.6 & 47.9 & 38.5 & 48.9 & 3.9 & 25.5 & 52.8 & 74.9 & \textbf{31.2} & 88.0 & 27.4 & 32.8 & \textbf{16.7} & 5.2 & 41.5  \\
 &\small Syn+Real & \textbf{54.5} & \textbf{82.9} & \textbf{53.1} & \textbf{46.3} & \textbf{57.3} & \textbf{5.1} & \textbf{29.6} & \textbf{61.4} & \textbf{78.2} & 31.1 & \textbf{92.5} & \textbf{29.6} & \textbf{41.1} & 14.5 & \textbf{6.5} & \textbf{45.6} \\ \midrule

\multirow{3}{1.8em}{\rotatebox[origin=c]{0}{\small{16M}}}   & \small Syn &44.2 & 32.4 & 11.5 & 41.6 & 51.3 & 4.9 & 27.4 & 58.3 & 72.1 & 24.8 & 83.6 & 16.7 & 29.5 & 4.6 & 5.9 & 33.9  \\
 & \small Real & 62.9 & 85.2 & 58.1 & 49.0 & 60.6 & \textbf{5.0} & 30.4 & 61.9 & 81.5 & \textbf{40.9} & 93.1 & \textbf{43.2} & 39.4 & \textbf{28.0} & 7.4 & 49.8  \\
 &\small Syn+Real & \textbf{64.8} & \textbf{87.5} & \textbf{61.0} & \textbf{53.7} & \textbf{63.3} & \textbf{4.9} & \textbf{36.5} & \textbf{67.7} & \textbf{82.8} & \textbf{38.6} & \textbf{94.5} & \textbf{37.6} & \textbf{48.2} & \textbf{28.6} & \textbf{8.2} & \textbf{51.9}\\ \midrule

\multirow{3}{2.3em}{\rotatebox[origin=c]{0}{\small{128M}}} &\small Syn & 63.7 & 45.1 & 15.9 & 52.3 & 67.1 & 9.3 & 37.8 & 75.7 & 80.5 & 39.1 & 93.2 & 8.0 & 35.7 & 10.1 & 9.5 & 42.9 \\
 &\small Real & \textbf{81.9} & 90.5 & \textbf{70.9} & 62.5 & 78.7 & 10.7 & 46.0 & \textbf{85.9} & 88.7 & \textbf{60.4} & 96.0 & \textbf{48.3} & 57.8 & 42.7 & \textbf{14.2} & 62.3 \\
 &\small Syn+Real & 81.6 & \textbf{91.0} & 70.4 & \textbf{64.0} & \textbf{79.4} & \textbf{11.9} & \textbf{52.5} & 85.1 & \textbf{90.2} & 59.5 & \textbf{97.0} & 47.3 & \textbf{61.1} & \textbf{45.3} & 14.1 & \textbf{63.4} \\ \midrule
 \multirow{3}{2.3em}{\rotatebox[origin=c]{0}{\small{371M}}} &\small Syn & 70.1 & 51.9 & 26.2 & 55.5 & 70.8 & 12.3 & 41.5 & 79.6 & 83.6 & 45.5 & 95.7 & 28.8 & 39.3 & 20.6 & 10.9 & 48.8  \\
 &\small Real & \textbf{85.7} & \textbf{93.9} & \textbf{75.6} & \textbf{67.5} & \textbf{83.3} & 14.2 & 50.1 & \textbf{88.8} & 91.1 & \textbf{67.0} & 97.0 & 43.9 & \textbf{66.6} & 42.8 & \textbf{17.5} & 65.7 \\
 &\small Syn+Real & 84.6 & 92.4 & 73.2 & 67.1 & 82.0 & \textbf{17.2} & \textbf{56.8} & 86.4 & \textbf{91.7} & 61.6 & \textbf{97.3} & \textbf{52.2} & 65.9 & \textbf{46.7} & 16.0 & \textbf{66.1} \\

\bottomrule[1.2pt]
\end{tabular}}
\end{center}
\vspace{-5mm}
\end{table*}

\subsection{Setup}
We investigated the scaling behavior of synthetic data in CLIP training using the extensive LAION-400M dataset. The synthetic images were generated using Stable Diffusion. We compare across different CFG scales and choose the optimal one ($1.5$) for CLIP training. For evaluation, we followed the prompt templates from~\cite{radford2021learning} and conduct zero-shot classification on ImageNet and 15 different fine-grained classification datasets, including Food-101~\cite{bossard2014food}, Stanford Cars~\cite{krause20133d}, Oxford Pets~\cite{parkhi2012cats} etc. The training scale begins with 1 million image-text pairs, progressively scaling up to encompass the full dataset of $371$\footnote{The LAION-400M dataset we used contains slightly less samples compared to the orignal one because of link rot.} million samples. All models use ViT-B as the backbone with a patch size of $16$, and are trained for $32$ epochs across all dataset scales. Detailed training hyper-parameters are in Appendix~\ref{sec:appendix-clip-training-detail}. Comparisons on different CFG scales are also available in Appendix~\ref{sec:appendix-clip}.

\subsection{Scaling Analysis}

\begin{figure}[t]
\centering
\includegraphics[width=1.0\linewidth]{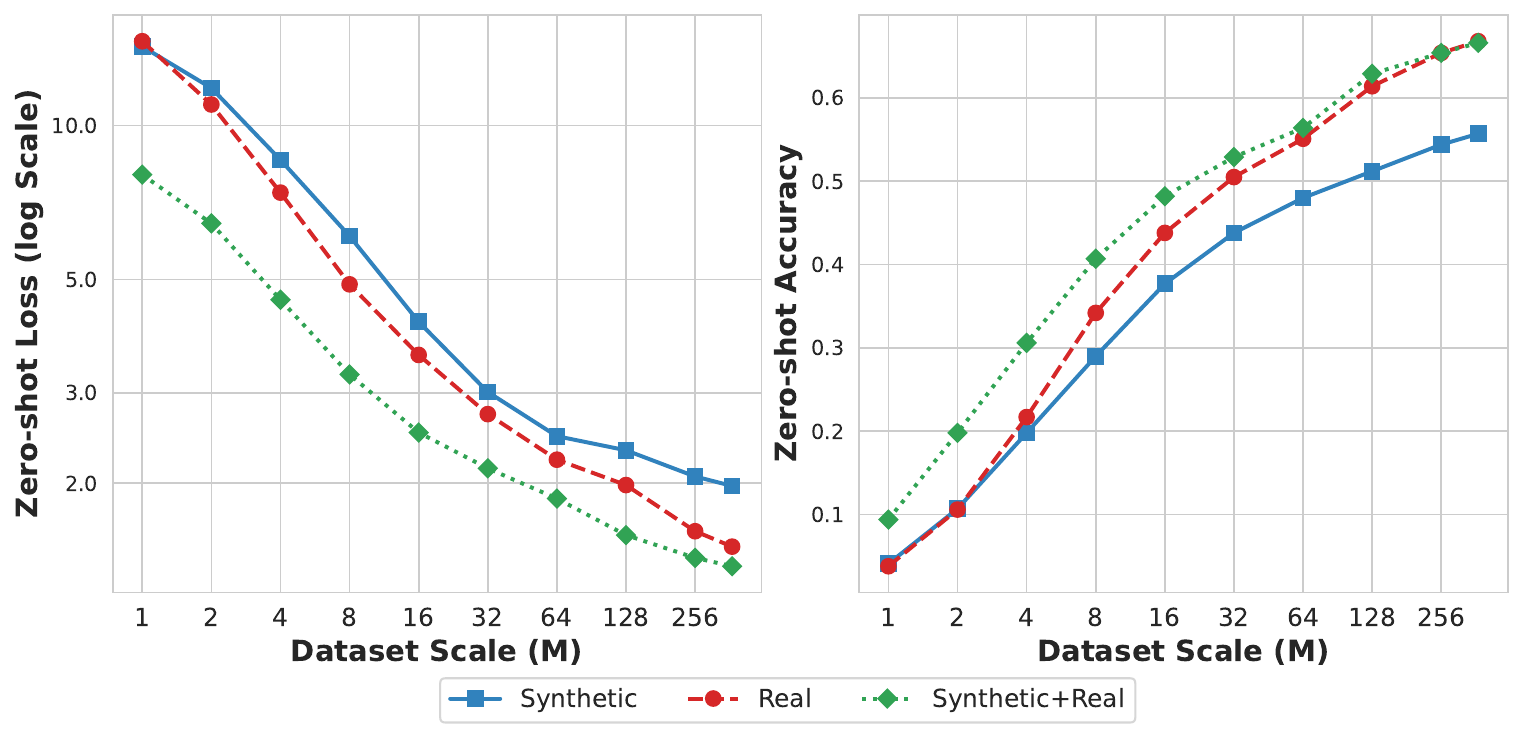}
\caption{
\small
Scaling behavior for CLIP models trained on LAION-400M subsets of different scales. Models are trained with synthetic, real, or a combination of synthetic and real images, and are evaluated with ImageNet zero-shot accuracy. Dataset scale here refers to the number of captions.}
\label{fig:clip_scaling}
\end{figure}

We evaluated the scaling behavior across three different data setups: (1) using only synthetic images, (2) using only real images, and (3) using a combination of both synthetic and real images.  Dataset scale here refers to the number of captions. When combining synthetic and real images for training, we maintained a consistent text scale throughout. During each training iteration, we randomly selected one image, either real or synthetic, for use. The comparative analysis of these setups, evaluated on zero-shot classification loss and accuracy on ImageNet validation set, is depicted in Figure~\ref{fig:clip_scaling}.

The analysis revealed that for all three scenarios, zero-shot classification loss adheres to the power-law relationship when the data amount is under around 64 million, compared to the 4 million scale in supervised training. In this range, the loss and data scale maintain a linear relationship in logarithmic space. Additionally, while the scaling efficiency (reflected in the slope of the curve) of synthetic data is somewhat lower than that of real data, this discrepancy is less pronounced than in the supervised classifier settings. However, a noticeable performance gap persists between synthetic and real images, which is likely attributable to concept mismatches between generated images and corresponding texts in certain classes, as discussed in Section~\ref{sec:poor-class}.

Moreover, our results indicate that combining synthetic and real images during CLIP training can significantly enhance zero-shot performance, particularly when the dataset is limited. For instance, in training scenarios with fewer than 10 million image-text pairs, integrating synthetic images with real data can boost performance by up to $5\%$.

\subsection{Scaling on downstream datasets}
We followed the same setup and extended our comparison to include the scaling behavior of synthetic versus real images on 15 fine-grained classification datasets, detailed in Table~\ref{table:clip_result}. This analysis indicates a scaling behavior in these datasets that is consistent with our findings from the ImageNet evaluations. Notably, a combination of synthetic and real images demonstrated superior performance in most scenarios, particularly when the total dataset size was under 100 million samples. In cases with extremely limited data availability, such as with just 1 million samples, training on synthetic images occasionally yielded better performance than with real images, for some certain tasks, such as Pets~\cite{parkhi2012cats} and SUN397~\cite{xiao2010sun}.

\section{Discussion}

In this paper, we investigate the scaling laws of synthetic data in model training and identify three key factors that significantly influence scaling behavior: the choice of \emph{models}, the \emph{classifier-free guidance scale}, and the selection of \emph{prompts}. After optimizing these elements and increasing the scale of training data, we find that, as expected, synthetic data still does not scale as effectively as real data, particularly for supervised classification on ImageNet. This limitation largely stems from the inability of standard text-to-image models to accurately generate certain concepts. However, our study also highlights several scenarios where synthetic data proves advantageous: (1) In certain classes, synthetic data demonstrates better scaling behavior compared to real data; (2) Synthetic data is particularly effective when real data is scarce, for instance, in CLIP training with limited datasets; (3) Models trained on synthetic data may exhibit superior generalization to out-of-distribution data. We hope our findings will pave the way for further research in this field.

\noindent \textbf{Acknowledgements.} 
The authors would like to thank Shobhita Sundaram, Julia Chae, Sara Beery, and the VisCam team for fruitful discussions, Yuanzhen Li for helping with computation resources, and Jason Baldridge and Sergey Ioffe for guidance and check on publication policy.
\clearpage
{
    \small
    \bibliographystyle{ieeenat_fullname}
    \bibliography{main}
}
\clearpage

\begin{appendices}
\setcounter{table}{0}
\renewcommand{\thetable}{A\arabic{table}}
\setcounter{figure}{0}
\renewcommand{\thefigure}{A\arabic{figure}}

\section{Details on Supervised Training}
\label{sec:implementation_details}
\subsection{Training Hyper-parameters}
Supervised training was conducted on both the real ImageNet training set and various synthetic ImageNet generated by text-to-iamge models at different dataset scales. To ensure fair comparisons across different setups, identical training hyper-parameters were used for both real and synthetic images. Our training approach aligns with the setup described in \cite{steiner2021train}, utilizing binary cross-entropy loss. The number of total training iterations and warm-up iterations were adjusted in proportion to the dataset scale in logarithmic space. For instance, models at the 1 million scale were trained for 95k iterations with a 10k iteration warm-up period. At the 2 million scale, training was extended to 190k iterations with a 20k iteration warm-up, and for the 4 million scale, the training and warm-up periods were increased to 285k and 30k iterations, respectively. More detailed descriptions of the training hyper-parameters are provided in Table~\ref{table:hyperparam-supervised}.
\begin{table}[h]
\centering
\caption{
\small Detailed pre-training hyper-parameters for supervised training on both real ImageNet training set and synthetic ImageNet generated by text-to-image models.}
\label{table:hyperparam-supervised}

\centering
\begin{minipage}{0.9\linewidth}{\begin{center}
\resizebox{0.98\textwidth}{!}{
\begin{tabular}{l@{\hspace{2.5em}}|@{\hspace{.5em}}l@{\hspace{2.5em}}}
\toprule
Config & Value \\
\midrule
Batch size & $4096$ \\
Optimizer & Adam~\cite{kingma2014adam} \\
Learning rate & $3\times10^{-3}$ \\
Weight decay & $0.1$ \\
Adam $\beta$ & $\beta_1, \beta_2=(0.9, 0.999)$\\
Total iterations & $95$k for 1M \\
Warm up iterations & $10$k for 1M \\
Learning rate schedule & cosine decay \\
Mixup & 0.5 \\
Dropout & 0.1 \\
Stochastic depth & 0.1 \\
Augmentation & RandAug(2, 15)~\cite{cubuk2020randaugment} \\
\bottomrule
\end{tabular}}
\end{center}}\end{minipage}

\end{table}
\subsection{Details on Text Prompts}
In this section, we provide more details on the different configurations of text prompts used for Class-specific Prompts, as outlined in Section~\ref{sec:3factor}.
For the \textit{Classnames + Hypernyms} configuration, we utilized all hypernyms associated with each specific ImageNet category, separated by commas.
Regarding \textit{CLIP templates}, we employed two sets of prompt templates with different number of sentences. The first one includes the 80 distinct sentence originally used in the CLIP paper \cite{radford2021learning} and its inference code\footnote{https://github.com/openai/CLIP/tree/main/notebooks}. The second set includes a subset of 7 templates, as recommended in \cite{li2023scaling}.
Additionally, we incorporated two more prompt configurations for comparison, following the approach in \cite{sariyildiz2023fake}:
(1) \textit{Classnames + Description + Places}, which combines ImageNet class names with their WordNet descriptions, followed by a background category sampled from the Places dataset~\cite{zhou2014learning}.
(2) \textit{Classnames + Hypernyms + Places}, which is similar to the previous configuration but replaces the descriptions with WordNet hypernyms, also incorporating a background category from Places.

Together with the configurations described in Section~\ref{sec:3factor} of the main paper, these methods result in a total of $8$ different configurations for text prompts when generating images for ImageNet categories. Additional visualizations of images produced by each these prompt configurations are included in Appendix~\ref{sec:appendix-prompt}.
\subsection{Evaluation on Downstream Datasets}
\label{sec:appendix-ds-linear}
In addition to evaluating the trained supervised classifiers directly, we also conducted linear probing on 15 different fine-grained classification datasets. Detailed descriptions of these datasets can be found in Appendix~\ref{sec:appendix-ds}.
To perform linear probing on these datasets, we first removed the linear classification head from the classifier trained on ImageNet. Then, we extracted features from both the training and testing sets of each dataset, without applying any data augmentation. Subsequently, logistic regression was employed on these extracted features. The logistic regression layer was optimized using L-BFGS, with a maximum number of iterations equals $500$.
For a detailed comparison of these results, please refer to Appendix~\ref{sec:appendix-scaling-ds}.

\begin{table*}[t]
\begin{center}
\caption{\small{Detailed metrics and number of training and testing images of the downstream classification datasets. Only test images are used in the zero-shot classification task for CLIP evaluation.}}
\label{table:appendix-datasets}
\resizebox{0.9\textwidth}{!}{
\begin{tabular}{@{\hspace{2.5em}}l@{\hspace{5.5em}}c@{\hspace{3.5em}}c@{\hspace{4.0em}}c@{\hspace{5.0em}}c@{\hspace{2.5em}}}
\toprule[1.2pt]
Dataset          & $\#$ Categories & $\#$ Train Images & $\#$ Test Images & Val Metric       \\
\midrule
Food-101~\cite{bossard2014food}          & 101 & 75,750 & 25,250 & Top-1 Accuracy    \\
CIFAR-10~\cite{krizhevsky2009learning}             & 10  & 50,000 & 10,000 & Top-1 Accuracy  \\
CIFAR-100~\cite{krizhevsky2009learning}     & 100 & 50,000 & 10,000    & Top-1 Accuracy     \\
SUN397~\cite{xiao2010sun}     & 397 & 19,850 & 19,850       & Top-1 Accuracy     \\
Stanford Cars~\cite{krause20133d}     & 196 & 8,144  & 8,041    & Top-1 Accuracy  \\
FGVC Aircraft~\cite{maji2013fine}& 100 & 6,667  & 3,333   & Mean per class  \\
DTD~\cite{cimpoi2014describing}     & 47  & 3,760  & 1,880          & Top-1 Accuracy      \\
Oxford Pets~\cite{parkhi2012cats} & 37  & 3,680  & 3,669     & Mean per class \\
Caltech-101~\cite{fei2006one} & 102 & 3,060  & 6,085    & Mean per class  \\
Oxford Flowers~\cite{nilsback2008automated}& 102 & 2,040  & 6,149  & Mean per class  \\
STL-10~\cite{coates2011analysis}   & 10  & 1,000  & 8,000        & Top-1 Accuracy       \\
EuroSAT~\cite{helber2019eurosat}   & 10  & 10,000 & 5,000         & Top-1 Accuracy     \\
RESISC45~\cite{cheng2017remote}     & 45  & 25,200 & 6,300       & Top-1 Accuracy    \\
GTSRB~\cite{stallkamp2011german}      & 43  & 26,640 & 12,630      & Top-1 Accuracy     \\
Country211~\cite{radford2021learning,thomee2016yfcc100m}      & 211 & 42,200 & 21,100    & Top-1 Accuracy   \\
\bottomrule[1.2pt]
\end{tabular}
}
\end{center}
\vspace{-5mm}
\end{table*}
\section{Details on CLIP Training}
\label{sec:appendix-clip-training-detail}
\subsection{Hyper-Parameters}
Previous studies \cite{mu2022slip,radford2021learning} along with our empirical analysis, indicate the necessity of using different training parameters according to dataset scale. Specifically, for smaller-scale datasets, a larger learning rate and weight decay are recommended to mitigate overfitting. Conversely, for larger datasets, both the learning rate and weight decay should be reduced. Accordingly, we have followed two distinct sets of hyper-parameters within the CLIP training pipeline, one tailored for datasets with fewer than 100 million captions following the parameter in~\cite{mu2022slip}, and another for those exceeding this threshold following the parameter in~\cite{radford2021learning}. The specific parameters for both configurations are outlined in Table~\ref{table:appendix-clip-hyperparam}. 
Models are trained for 32 epochs across all data scales. The number of warmup steps was set to 600 for the 1 million scale, 1200 for the 2 million scale, and 2000 for scales of 4 million or greater.
It is important to note that we maintained consistent training hyper-parameters across all three different types of data sources (synthetic, real, synthetic+real) at the same data scale to ensure fair comparisons. For an in-depth comparison of the effects of different hyper-parameters in CLIP training at different scales, please refer to the experimental details provided in Appendix~\ref{sec:appendix-clip-hyperselect}.
\begin{table}[h]
\centering
\caption{
\small Detailed pre-training hyper-parameters for CLIP at different dataset scales.}
\label{table:appendix-clip-hyperparam}

\subfloat[
\small
Hyper-parameter for CLIP with $< 100M$ samples.
]{
\centering
\begin{minipage}{0.9\linewidth}{\begin{center}
\resizebox{0.98\textwidth}{!}{
\begin{tabular}{l@{\hspace{2.5em}}|@{\hspace{.5em}}l@{\hspace{2.5em}}}
\toprule
Config & Value \\
\midrule
Batch size & $8192$ \\
Optimizer & AdamW~\cite{loshchilov2017decoupled} \\
Learning rate & $1\times10^{-3}$ \\
Weight decay & $0.5$ \\
Adam $\beta$ & $\beta_1, \beta_2=(0.9, 0.98)$\\
Adam $\epsilon$ & $1\times10^{-8}$ \\
Total epochs & $32$ \\
Warm up iterations & $600, 1200, 2000$ \\
Learning rate schedule & cosine decay \\
\bottomrule
\end{tabular}}
\end{center}}\end{minipage}
}
\\
\vspace{1.0em}
\subfloat[
\small
Hyper-parameter for CLIP with $\geq 100M$ samples.
]{
\centering
\begin{minipage}{0.9\linewidth}{\begin{center}
\resizebox{0.98\textwidth}{!}{
\begin{tabular}{l@{\hspace{2.5em}}|@{\hspace{.5em}}l@{\hspace{2.5em}}}
\toprule
Config & Value \\
\midrule
Batch size & $32768$ \\
Optimizer & AdamW~\cite{loshchilov2017decoupled} \\
Learning rate & $5\times10^{-4}$ \\
Weight decay & $0.2$ \\
Adam $\beta$ & $\beta_1, \beta_2=(0.9, 0.98)$\\
Adam $\epsilon$ & $1\times10^{-6}$ \\
Total epochs & $32$ \\
Warm up iterations & $2000$ \\
Learning rate schedule & cosine decay \\
\bottomrule
\end{tabular}}
\end{center}}\end{minipage}
}\\
\end{table}

\subsection{Downstream Dataset}
\label{sec:appendix-ds}
For all the pre-trained CLIP models, we conducted zero-shot evaluations on ImageNet and 15 other widely used downstream classification datasets. These datasets include Food-101~\cite{bossard2014food}, Stanford Cars~\cite{krause20133d}, SUN397~\cite{xiao2010sun}, Oxford Pets~\cite{parkhi2012cats}, among others. Detailed information about these evaluation datasets can be found in Table~\ref{table:appendix-datasets}. It's important to note that for zero-shot evaluations, only the test images from these datasets are used.

\subsection{Zero-shot Evaluation Details}
We employed the same text prompt templates as referenced in \cite{radford2021learning}, following a similar text ensembling strategy. For each category, text features were computed for every single template, and the mean average of these features across all templates was used to represent the final text feature for that specific category.
Given that CLIP training involves a trainable temperature parameter, $\tau$, it is necessary to incorporate this parameter during zero-shot evaluation to accurately compute the zero-shot classification loss. 
Let $z_{img}$ be the image feature from the visual encoder, $z_{txt}$ denote the aggregated text feature. Assuming a total of $C$ classes, with $z_{txt_c}$ as the text feature for $c-th$ category, the zero-shot classification loss is calculated as follows:
$$
L = -\log \frac{\exp (\text{sim} (z_{img}, z_{txt_c})\cdot\tau)}{\sum_{c'=1}^{C} \exp(\text{sim} (z_{img}, z_{txt_{c'}})\cdot\tau)}
$$
Here $\text{sim} (z_{img}, z_{txt_c})$ calculates the dot product, measuring the similarity between image feature and text features for each category.

\section{Details for Performance at 1.3M}
\label{sec:appendix-1.3m}
\begin{table*}[t]
\begin{center}
\caption{\small{Detailed comparison on ImageNet Validation performance and recognizability, diversity, FID and LPIPS for different CFG scale and prompt configurations with Stable Diffusion as the text-to-image model.}}
\label{table:appendix-prompt}

\resizebox{\textwidth}{!}{
\begin{tabular}{@{\hspace{2em}}c@{\hspace{1em}}|@{\hspace{1em}}l@{\hspace{2em}}|@{\hspace{1em}}cc@{\hspace{1em}}|@{\hspace{.5em}}c@{\hspace{1em}}|c@{\hspace{2em}}c@{\hspace{2em}}c@{\hspace{2em}}}
\toprule[1.2pt]
\multicolumn{8}{l}{\textit{Text-to-Image Model: \textbf{Stable Diffusion}, main comparison on Prompt Config}}\\
\midrule
\midrule
\textbf{CFG Scale}
& \textbf{Prompt Config} & IN loss($\downarrow$) & IN Top1 & Recognizability & Diversity & FID($\downarrow$) & LPIPS($\downarrow$) \\
\midrule
\multirow{9}{*}{2}
& Word2Sen & 3.27 & 38.42 & 0.315 & \bf 0.850 & 3.566 & 0.717 \\
 & CLIP Templates (7) & 2.63 & 49.26 & 0.522 & 0.781 & 3.297 & 0.714 \\
 & CLIP Templates (80) & 2.76 & 49.88 & 0.510 & 0.790 & 3.437 & 0.719 \\
 & Classnames+Hypernym & 3.28 & 45.19 & 0.569 & 0.713 & 2.553 & 0.698 \\
 & Classnames+Description & 3.27 & 45.05 & \bf 0.615 & 0.696 & 2.678 & 0.697 \\
 & Classnames+Hypernym+Places & 3.14 & 43.91 & 0.381 & 0.836 & 4.441 & 0.712 \\
 & Classnames+Description+Places & 3.06 & 46.18 & 0.524 & 0.758 & 2.890 & 0.704 \\
 & Classnames & 3.05 & 47.82 & 0.603 & 0.718 & 2.589 & \bf 0.702 \\
 & IN-Captions & \bf 2.23 & \bf 55.04 & 0.573 & 0.757 & \bf 2.450 & 0.714 \\
\midrule
\multirow{4}{*}{7.5}
& CLIP Templates (7) & 4.17 & 38.39 & 0.702 & 0.650 & 5.681 & 0.731 \\
 & CLIP Templates (80) & 3.86 & 40.26 & 0.687 & \bf 0.670 & 5.619 & 0.739 \\
 & Classnames & 4.96 & 31.21 & \bf 0.780 & 0.541 & 4.113 & \bf 0.707 \\
 & IN-Captions & \bf 3.56 & \bf \bf 40.38 & 0.725 & 0.632 & \bf 3.641 & 0.737 \\
 \bottomrule[1.2pt]
\end{tabular}
}
\end{center}
\end{table*}

\begin{table*}[t]
\begin{center}
\caption{\small{Detailed comparison on ImageNet Validation performance and recognizability, diversity, FID and LPIPS for different text-to-image models and CFG scales. All configurations use IN-Captions as prompts.}}
\label{table:appendix-t2i}

\resizebox{\textwidth}{!}{
\begin{tabular}{@{\hspace{2em}}c@{\hspace{1em}}|@{\hspace{2em}}c@{\hspace{2em}}|@{\hspace{1em}}cc@{\hspace{1em}}|@{\hspace{.5em}}c@{\hspace{1em}}|c@{\hspace{2em}}c@{\hspace{2em}}c@{\hspace{2em}}}
\toprule[1.2pt]
\multicolumn{8}{l}{\textit{Text Prompt: \textbf{IN-Captions}, main comparison on Text-to-image models}}\\
\midrule
\midrule
\textbf{Text-to-Image Model }& \textbf{CFG Scale} & IN loss($\downarrow$) & IN Top1 & Recognizability & Diversity & FID($\downarrow$) & LPIPS($\downarrow$) \\
\midrule
\multirow{8}{*}{Stable Diffusion} & 1.5 & 2.14 & 54.66 & 0.484 & \textbf{0.800} & \textbf{2.403} & \textbf{0.710} \\
 & 2 & 2.23 & \textbf{55.04} & 0.573 & 0.757 & 2.450 & 0.714 \\
 & 3 & 2.38 & 54.10 & 0.655 & 0.705 & 2.790 & 0.722 \\
 & 4 & 2.61 & 51.13 & 0.690 & 0.675 & 3.100 & 0.728 \\
 & 6 & 2.92 & 46.99 & 0.717 & 0.644 & 3.483 & 0.734 \\
 & 7.5 & 3.56 & 40.38 & 0.725 & 0.632 & 3.641 & 0.737 \\
 & 8 & 3.47 & 40.87 & 0.726 & 0.629 & 3.655 & 0.738 \\
 & 10 & 3.42 & 33.59 & \textbf{0.730} & 0.621 & 3.723 & 0.740 \\ \midrule
\multirow{3}{*}{Imagen} & 1 & 1.84 & 58.52 & 0.466 & \textbf{0.810} & \textbf{3.451} & 0.713 \\
 & 1.5 & 1.78 & \textbf{61.51} & 0.647 & 0.719 & 4.546 & 0.733 \\
 & 2 & 1.93 & 60.58 & 0.714 & 0.671 & 6.867 & \textbf{0.701} \\ \midrule
\multirow{4}{*}{Muse} & 0.1 & 2.05 & 54.19 & 0.473 & \textbf{0.789} & \textbf{4.057} & 0.755 \\
 & 0.3 & 2.08 & \textbf{54.45} & 0.520 & 0.760 & 4.616 & 0.749 \\
 & 0.5 & 2.13 & 54.03 & 0.554 & 0.738 & 5.189 & 0.745 \\
 & 1 & 2.37 & 51.55 & \textbf{0.599} & 0.700 & 6.274 & \textbf{0.734} \\ 
 \bottomrule[1.2pt]
\end{tabular}
}
\end{center}
\end{table*}

\begin{table*}[t]
\begin{center}
\caption{\small{Detailed comparison on ImageNet Validation performance and recognizability, diversity, FID($\downarrow$) and LPIPS($\downarrow$) for different text-to-image models and CFG scales. All configurations use Stable Diffusion as the text-to-image model.}}
\label{table:appendix-CFG}

\resizebox{\textwidth}{!}{
\begin{tabular}{@{\hspace{2em}}c@{\hspace{2em}}|@{\hspace{2em}}c@{\hspace{2em}}|@{\hspace{1em}}cc@{\hspace{1em}}|@{\hspace{.5em}}c@{\hspace{1em}}|c@{\hspace{2em}}c@{\hspace{2em}}c@{\hspace{2em}}}
\toprule[1.2pt]
\multicolumn{8}{l}{\textit{Text-to-Image Model: \textbf{Stable Diffusion}, main comparison on CFG Scale}}\\
\midrule
\midrule
\textbf{Prompt Config }& \textbf{CFG Scale} & IN loss($\downarrow$) & IN Top1 & Recognizability & Diversity & FID($\downarrow$)& LPIPS($\downarrow$)\\
\midrule
\multirow{8}{*}{ClassNames} & 1.5 & 2.84 & \textbf{48.10} & 0.499 & \textbf{0.778} & \textbf{2.525} & \textbf{0.702} \\
 & 2 & 3.05 & 47.82 & 0.603 & 0.718 & 2.589 & \textbf{0.702} \\
 & 3 & 3.41 & 44.91 & 0.697 & 0.644 & 2.981 & 0.704 \\
 & 4 & 3.80 & 41.55 & 0.734 & 0.602 & 3.383 & 0.705 \\
 & 6 & 4.82 & 33.58 & 0.770 & 0.559 & 3.897 & 0.707 \\
 & 7.5 & 4.96 & 31.21 & 0.780 & 0.541 & 4.113 & 0.707 \\
 & 8 & 5.04 & 29.78 & 0.780 & 0.537 & 4.167 & 0.707 \\
 & 10 & 5.47 & 26.13 & \textbf{0.787} & 0.524 & 4.289 & 0.707 \\ \midrule
\multirow{5}{*}{ClassNames+Description} & 1.5 & 3.05 & \textbf{45.84} & 0.523 & \textbf{0.753} & \textbf{2.468} & \textbf{0.697} \\
 & 2 & 3.27 & 45.05 & 0.615 & 0.696 & 2.678 & \textbf{0.697} \\
 & 3 & 3.87 & 41.78 & 0.699 & 0.627 & 3.265 & 0.699 \\
 & 4 & 4.14 & 38.92 & 0.738 & 0.588 & 3.714 & 0.701 \\
 & 6 & 4.58 & 33.52 & \textbf{0.762} & 0.546 & 4.223 & 0.703 \\ \midrule
\multirow{5}{*}{ClassNames+Hypernym} & 1.5 & 3.06 & \textbf{45.22} & 0.475 & \textbf{0.770} & \textbf{2.463} & \textbf{0.698} \\
 & 2 & 3.28 & 45.19 & 0.569 & 0.713 & 2.553 & \textbf{0.698} \\
 & 3 & 3.70 & 42.52 & 0.652 & 0.643 & 2.961 & 0.700 \\
 & 4 & 4.35 & 38.02 & 0.687 & 0.604 & 3.336 & 0.701 \\
 & 6 & 4.69 & 33.38 & \textbf{0.717} & 0.561 & 3.831 & 0.703 \\ \midrule
\multirow{8}{*}{CLIP Templates (80)} & 1.25 & 2.58 & 47.59 & 0.344 & \textbf{0.856} & \textbf{3.193} & \textbf{0.712} \\
 & 1.5 & 2.61 & 49.00 & 0.413 & 0.831 & 3.200 & 0.714 \\
 & 1.75 & 2.66 & 49.77 & 0.468 & 0.809 & 3.284 & 0.717 \\
 & 2 & 2.76 & \textbf{49.88} & 0.510 & 0.790 & 3.437 & 0.719 \\
 & 3 & 3.00 & 48.76 & 0.600 & 0.740 & 4.098 & 0.726 \\
 & 4 & 3.27 & 46.56 & 0.642 & 0.711 & 4.653 & 0.731 \\
 & 6 & 3.70 & 42.24 & 0.677 & 0.681 & 5.349 & 0.736 \\
 & 7.5 & 3.86 & 40.26 & \textbf{0.687} & 0.670 & 5.619 & 0.739 \\ \midrule
\multirow{8}{*}{CLIP Templates (7)} & 1.25 & 2.57 & 47.86 & 0.350 & \textbf{0.851} & \textbf{3.029} & \textbf{0.709} \\
 & 1.5 & 2.63 & 49.24 & 0.424 & 0.824 & 3.017 & 0.711 \\
 & 1.75 & 2.69 & \textbf{50.20} & 0.478 & 0.801 & 3.116 & 0.712 \\
 & 2 & 2.63 & 49.26 & 0.522 & 0.781 & 3.297 & 0.714 \\
 & 3 & 3.07 & 48.69 & 0.617 & 0.726 & 4.025 & 0.720 \\
 & 4 & 3.42 & 45.93 & 0.658 & 0.695 & 4.661 & 0.724 \\
 & 6 & 4.00 & 41.15 & 0.692 & 0.663 & 5.393 & 0.729 \\
 & 7.5 & 4.17 & 38.39 & \textbf{0.702} & 0.650 & 5.681 & 0.731 \\ \midrule
\multirow{8}{*}{IN-Captions} & 1.5 & 2.14 & 54.66 & 0.484 & \textbf{0.800} & \textbf{2.403} & \textbf{0.710} \\
 & 2 & 2.23 & \textbf{55.04} & 0.573 & 0.757 & 2.450 & 0.714 \\
 & 3 & 2.38 & 54.10 & 0.655 & 0.705 & 2.790 & 0.722 \\
 & 4 & 2.61 & 51.13 & 0.690 & 0.675 & 3.100 & 0.728 \\
 & 6 & 2.92 & 46.99 & 0.717 & 0.644 & 3.483 & 0.734 \\
 & 7.5 & 3.56 & 40.38 & 0.725 & 0.632 & 3.641 & 0.737 \\
 & 8 & 3.47 & 40.87 & 0.726 & 0.629 & 3.655 & 0.738 \\
 & 10 & 3.42 & 33.59 & \textbf{0.730} & 0.621 & 3.723 & 0.740 \\ 
 \bottomrule[1.2pt]
\end{tabular}
}
\end{center}
\end{table*}
\begin{figure*}[t]
\centering
\includegraphics[width=\linewidth]{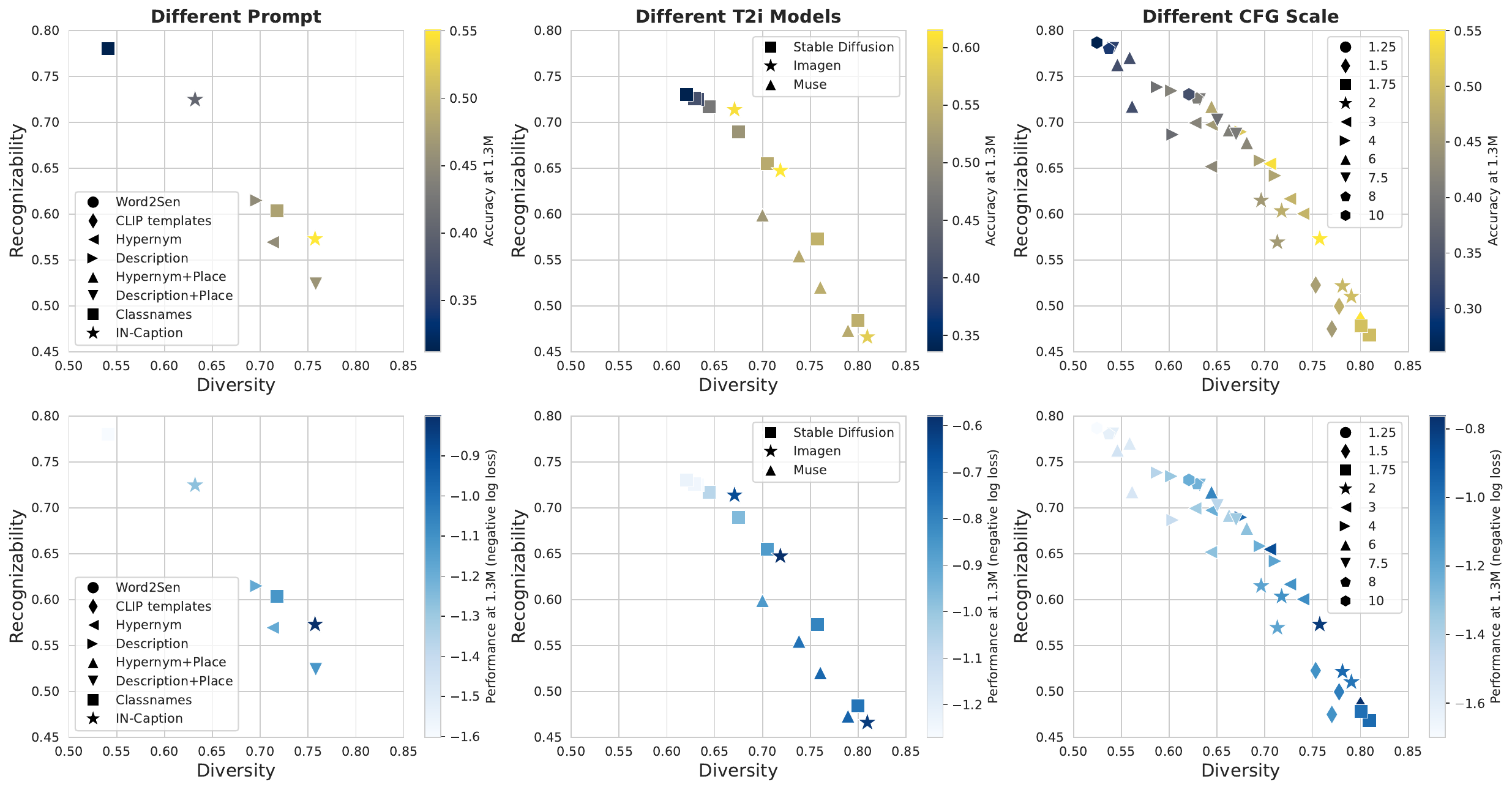}
\caption{
\small Recognizability versus diversity plot for different text-to-image configuration groups as described in Appendix~\ref{sec:appendix-1.3m}. Each column corresponds to one comparison group. The first column mainly compares on text prompts and corresponds to Table~\ref{table:appendix-prompt}. The second column compares different text-to-image models and corresponds to Table~\ref{table:appendix-t2i}. The last column mainly compares optimal CFG scale for Stable Diffusion and corresponds to Table~\ref{table:appendix-CFG}. On the top row, each point is colored by the top-1 accuracy in ImageNet validation set, on the bottom row the points are colored by the negative log loss.
}
\label{fig:appendix_rec_div_sep}

\includegraphics[width=\linewidth]{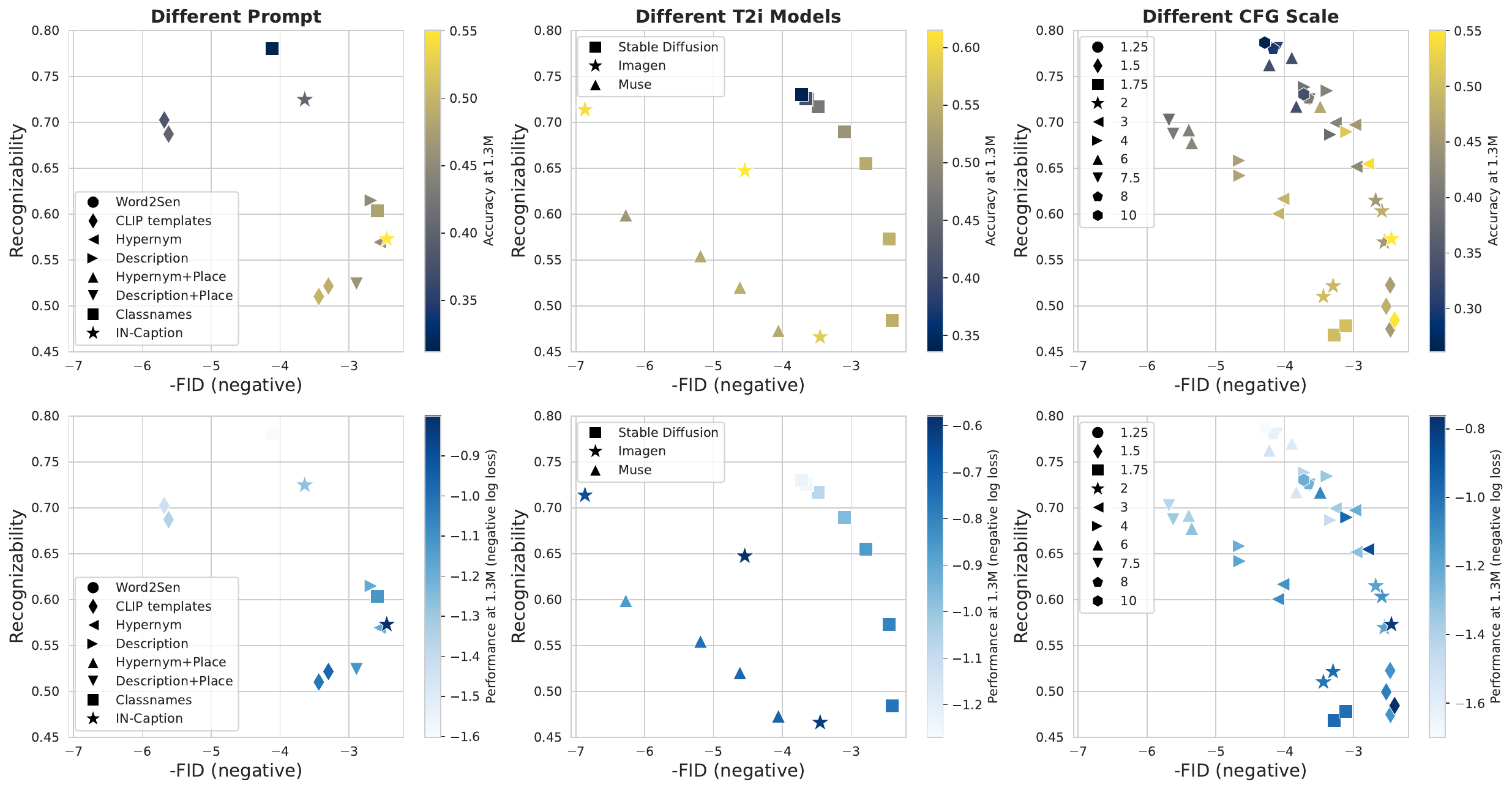}
\caption{
\small Recognizability versus FID plot for different text-to-image configuration groups as described in Appendix~\ref{sec:appendix-1.3m}. Each column corresponds to one comparison group. The detailed correspondence between the plot and tables is the same as Figure~\ref{fig:appendix_rec_div_sep}. On X-axis we take the negative of FID, upper right indicates better metric with lower FID and higher recognizability.
}
\label{fig:appendix_rec_fid_sep}
\end{figure*}

\begin{figure*}[t]
\centering
\includegraphics[width=\linewidth]{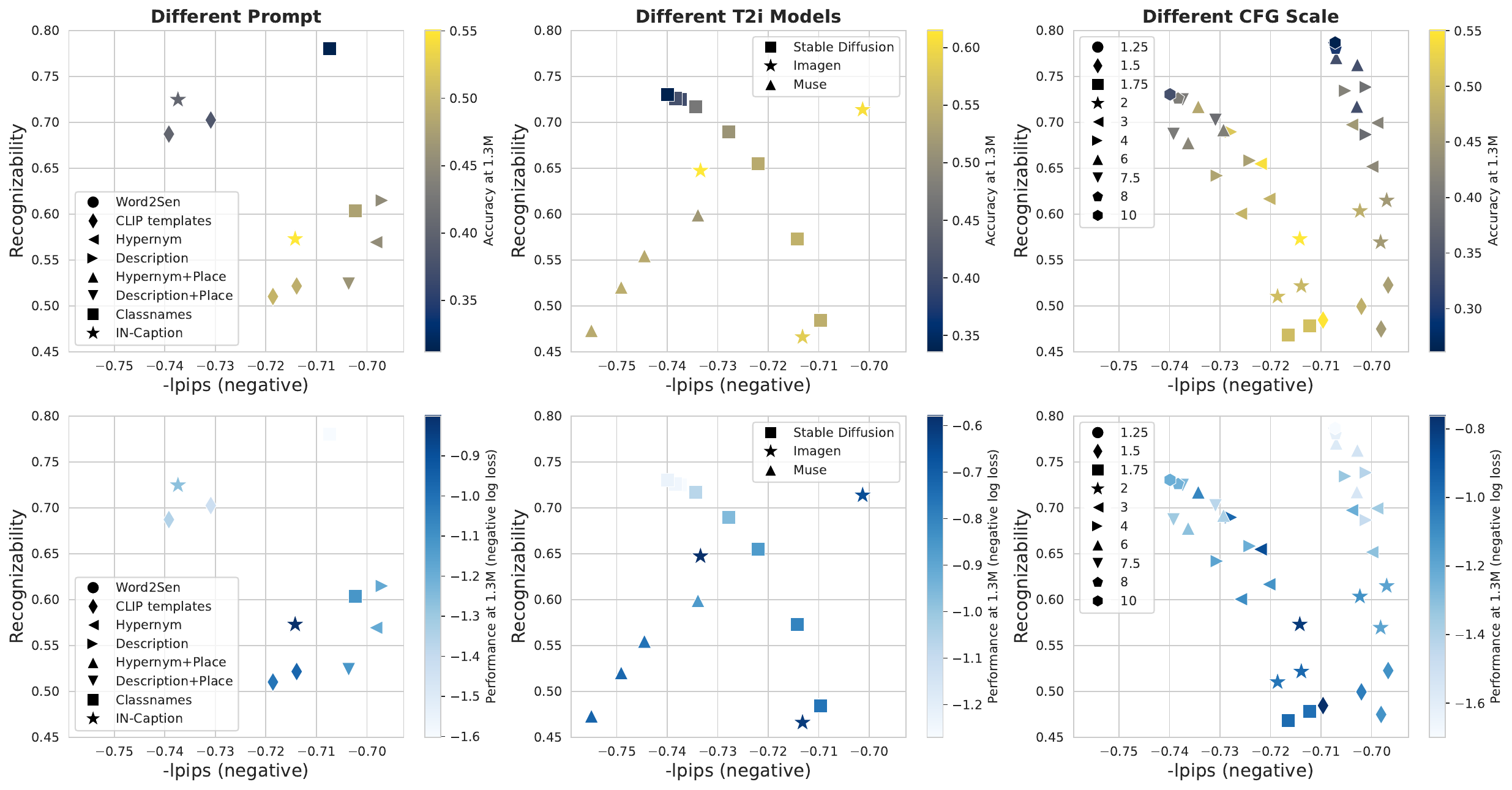}
\caption{
\small Recognizability versus LPIPS plot for different text-to-image configuration groups as described in Appendix~\ref{sec:appendix-1.3m}. Each column corresponds to one comparison group. The detailed correspondence between the plot and tables is the same as Figure~\ref{fig:appendix_rec_div_sep}. On X-axis we take the negative of LPIPS, upper right indicates better metric with lower LPIPS and higher recognizability.
}
\label{fig:appendix_rec_lpips_sep}
\includegraphics[width=\linewidth]{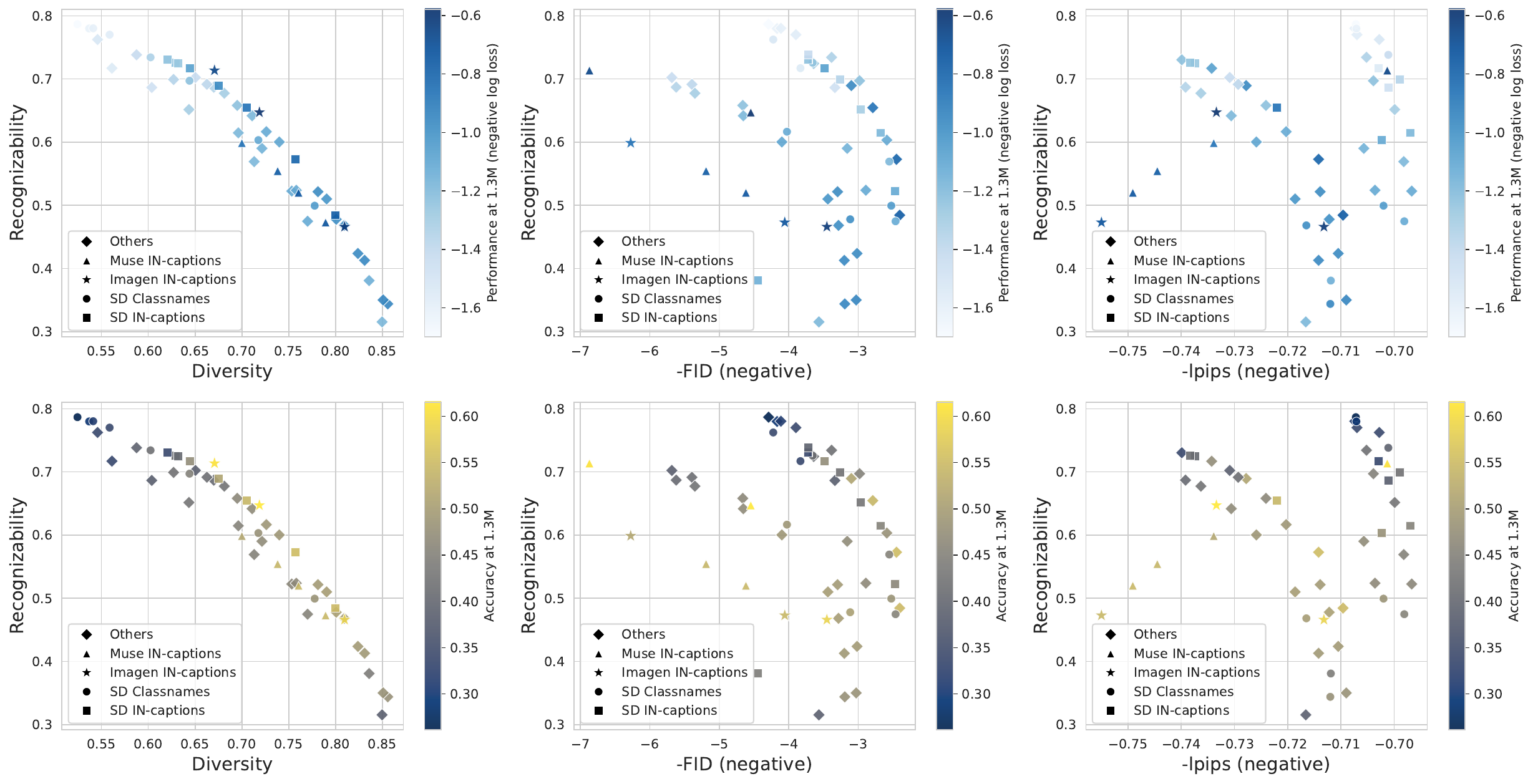}
\caption{
\small Recognizability versus diversity, FID or LPIPS plot for all the different text-to-image configurations under 1.3M scale. The first column corresponds to diversity, while the latter two correspond to FID and LPIPS, respectively. On X-axis we take the negative of FID and LPIPS. In the plots upper right indicates better metric with higher diversity, lower FID or LPIPS, and higher recognizability. Each point correspond to one configuration, and is color-coded by either the negative log loss (top row) or the top-1 accuracy (bottom) on the ImageNet validation set. 
}
\label{fig:appendix_rec_all}
\end{figure*}
As detailed in Section~\ref{sec:1.3m_performance} of the main paper, we adopted 54 different configurations encompassing distinct text-to-image models, CFG scales, and text prompts to generate 1.3 million synthetic images for each configuration. Following the generation process, a supervised model was trained on the images geenrated by each configuration. To facilitate a clearer comparison in our tables, we have categorized these 54 configurations into three distinct groups, with each group focusing on specific comparative factors:
\begin{itemize}
    \item The first group exclusively uses \textit{Stable Diffusion} as the text-to-image model. The primary comparison focus here is on the impact of varying \textbf{text prompt} configurations.
    \item The second group standardizes the text prompt configuration to \textit{IN-Caption}. This group's aim is to assess the effects of using different \textbf{text-to-image models} and to understand the behavior of CFG scales within each specific model, and to find the optimal CFG scale for each of them.
    \item The third group also exclusively uses \textit{Stable Diffusion} as the text-to-image model. Here, the comparison emphasis is on the impact of different \textbf{CFG} scales under different text prompt configurations.
\end{itemize}
By grouping the configurations into these three different groups, we aim to provide a more structured and comprehensible analysis.
In each of the three groups, we present the detailed validation loss (the negative log loss here is used to plot Figure~\ref{fig:rec_div} in the main paper) and top-1 accuracy on ImageNet validation set for models trained with different configurations, all under the scale of 1.3 million images. 

Table~\ref{table:appendix-prompt} presents the analysis for the first group. It shows that using \textbf{IN-Caption} as the text prompt yields the best performance across both CFG scales of $2$ and $7.5$. This superior performance is largely attributed to its ability to guide the text-to-image model to generate diverse images while maintaining high recognizability, thereby justifying our choice of IN-Caption for most of our experiments.
In the second group, detailed in Table~\ref{table:appendix-t2i}, we observe that different text-to-image models exhibit varying optimal CFG scales for image generation to train supervised models. Specifically, Stable Diffusion, Imagen, and Muse reach their optimal performance at CFG scales of 2, 1.5, and 0.3, respectively. These findings validate our decision to employ these specific CFG scales in our later study of scaling behavior for each model.
Table~\ref{table:appendix-CFG} covers the third group's comparisons, focusing on finding the optimal CFG scales for Stable Diffusion under different text prompts. The results indicate that for text prompts with less diversity, such as using Classnames or Classnames+Hypernym, a smaller optimal CFG scale (1.5) is better since it will lead to more diverse images during the generation process. In contrast, for more diverse text prompts, like IN-Captions and CLIP templates with 80 sentences, since there is more diversity on the text side relativly, a larger optimal CFG scale (2) is more effective.

In addition, we have included a recognizability versus diversity plot for each of the three comparison groups in Figure~\ref{fig:appendix_rec_div_sep}. Each point in these plots represents a specific configuration, and is color-coded based on either the top-1 classification accuracy or the negative log loss on ImageNet validation set.
The figures illustrate a trade-off between diversity and recognizability. Optimal performance is typically observed when there is a relatively better and more balanced trade-off between these two factors. Configurations characterized by either low diversity or low recognizability tend to result in suboptimal performance, indicating the necessity of maintaining a balance between these two factors.

\section{Evaluation under FID and LPIPS}
\label{sec:appendix-fid}
In addition to diversity, we computed two other key metrics: FID (Frechet Inception Distance)\cite{heusel2017gans} and LPIPS (Learned Perceptual Image Patch Similarity)\cite{zhang2018unreasonable}. Both of them are standard evaluation metrics for the text-to-image generation models. Our study examines the performance variations in relation to these two metrics. As detailed in Section~\ref{sec:metric-rec-div} of the main paper, these metric scores are also calculated using the synthetic test sets, which comprises $50,000$ images for each configuration:
\begin{itemize}
    \item The FID scores are derived by measuring the Frechet Inception Distance~\cite{heusel2017gans} between the synthetic test set, containing these 50,000 generated images, and the real ImageNet validation set.
    \item For LPIPS, we perform the calculation on a per-class basis. We randomly select and compute the similarity between 250 pairs of synthetic images for each class, and the final LPIPS metric is computed as the average across all classes.
\end{itemize}
The comparison of FID and LPIPS scores across each group is presented in Tables~\ref{table:appendix-prompt}, \ref{table:appendix-t2i}, and \ref{table:appendix-CFG}. Additionally, in Figures~\ref{fig:appendix_rec_fid_sep} and \ref{fig:appendix_rec_lpips_sep}, we plot a detailed comparison of the performance across different image generation configuration groups, substituting diversity with either FID or LPIPS. 
Considering that lower scores for FID indicate better distribution match and for LPIPS implies larger intra-class diversity,
we take the negative of these values for plotting purposes. This adjustment ensures consistency with the diversity plot on the X-axis, positioning better results towards the right.

Furthermore, we incorporate comparisons using diversity, FID, or LPIPS as the X-axis for all 54 text-to-image generation configurations in Figure~\ref{fig:appendix_rec_all}. Our findings reveal that while there is a moderate correlation between the FID score or LPIPS of generated images and the classification performance of models trained on them, the relationship is not definitive. In some cases, configurations with the same level of recognizability but lower FID scores or LPIPS show inferior classification performance. This suggests that while FID and LPIPS are effective metrics for evaluating the quality of the images generated by text-to-image models, their correlation with the performance of supervised classifiers trained on synthetic images is not as strong as expected. This observation underscores the need for a more specific metric tailored to evaluate the performance of supervised classifiers trained on such synthetic images.

\section{Detailed Scaling Behavior Comparison}
\label{sec:appendix-scaling-ds}
\begin{table*}[t]
\begin{center}
\caption{\small{Detailed scaling behavior on 15 different downstream classification datasets and ImageNet-A, ImageNet-R, ImageNet-Sketch and ImageNet-V2 validation set for supervised classifiers trained with real images from ImageNet training set and synthetic images from various configurations using Stable Diffusion. Dataset scale is in million.}}
\label{table:appendix_scaling}
\resizebox{\textwidth}{!}{
\begin{tabular}
{r|ccccccccccccccc|c|cccc}
\toprule[1.2pt]
\bf Scale&
\rotatebox[origin=lb]{90}{\smash{\small Food-101}} & \rotatebox[origin=lb]{90}{\smash{\small CIFAR-10}} & \rotatebox[origin=lb]{90}{\smash{\small CIFAR-100}} & \rotatebox[origin=lb]{90}{\smash{\small SUN397}} &
\rotatebox[origin=lb]{90}{\smash{\small Cars}} & \rotatebox[origin=lb]{90}{\smash{\small Aircraft}} & \rotatebox[origin=lb]{90}{\smash{\small DTD}} & \rotatebox[origin=lb]{90}{\smash{\small Pets}} & \rotatebox[origin=lb]{90}{\smash{\small Caltech-101}} &
\rotatebox[origin=lb]{90}{\smash{\small Flowers}} & \rotatebox[origin=lb]{90}{\smash{\small STL-10}} & \rotatebox[origin=lb]{90}{\smash{\small EuroSAT}} &
\rotatebox[origin=lb]{90}{\smash{\small RESISC45}} & \rotatebox[origin=lb]{90}{\smash{\small GTSRB}} & \rotatebox[origin=lb]{90}{\smash{\small Country211}}  & \rotatebox[origin=lb]{90}{\smash{\small \bf DS Average}}
& \rotatebox[origin=lb]{90}{\smash{\small ImageNet-A}} & \rotatebox[origin=lb]{90}{\smash{\small ImageNet-R}} & \rotatebox[origin=lb]{90}{\smash{\small ImageNet-Sketch}} & \rotatebox[origin=lb]{90}{\smash{\small ImageNet-V2}} 
\\
\midrule
\multicolumn{21}{c}{\textbf{Real ImageNet Training set}}\\
\midrule
0.125 & 61.4 & 80.6 & 60.9 & 47.3 & 24.0 & 31.0 & 64.7 & 77.2 & 73.4 & 86.7 & 88.0 & 95.8 & 87.4 & 57.0 & 12.0 & 63.2  & 2.6 & 14.6 & 5.7 & 36.3 \\
0.25 & 65.5 & 85.2 & 66.2 & 52.7 & 30.0 & 37.7 & 67.7 & 83.3 & 81.0 & 88.8 & 92.5 & 95.8 & 88.0 & 61.0 & 11.9 & 67.1 & 3.6 & 19.6 & 10.2 & 45.1 \\
0.5 & 71.0 & 89.7 & 72.5 & 57.6 & 44.2 & 43.4 & 70.2 & 88.8 & 87.2 & 92.1 & 96.0 & 95.8 & 89.8 & 65.2 & 12.4 & 71.7 & 5.8 & 27.3 & 17.6 & 55.7 \\
1 & 77.0 & 94.7 & 80.0 & 63.1 & 57.3 & 51.6 & 72.8 & 92.6 & 92.8 & 93.4 & 98.1 & 96.0 & 90.5 & 70.8 & 13.9 & 76.3  & 15.6 & 40.3 & 29.4 & 66.7 \\
1.3 & 77.8 & 94.6 & 81.0 & 64.3 & 62.5 & 53.1 & 74.0 & 93.5 & 93.4 & 93.4 & 98.6 & 96.1 & 90.0 & 71.7 & 14.4 & 77.2  & 18.7 & 42.2 & 31.2 & 68.8 \\
\midrule
\multicolumn{21}{c}{\textbf{Stable Diffusion, CFG scale=7.5, Classname}}\\
\midrule
0.125 & 56.4 & 75.8 & 54.1 & 43.2 & 24.7 & 31.5 & 61.9 & 73.1 & 65.4 & 83.9 & 81.3 & 94.7 & 84.6 & 53.8 & 9.4 & 59.6 & 1.7 & 15.4 & 6.8 & 20.6 \\
0.25 & 59.1 & 77.9 & 56.7 & 43.9 & 28.7 & 35.7 & 61.1 & 75.0 & 67.2 & 84.5 & 84.0 & 94.4 & 84.7 & 57.9 & 9.7 & 61.4 & 1.8 & 16.9 & 8.1 & 21.6 \\
0.5 & 60.2 & 79.4 & 58.8 & 46.5 & 31.9 & 37.6 & 64.5 & 77.7 & 73.4 & 84.7 & 87.1 & 95.3 & 86.3 & 61.1 & 10.3 & 63.7 & 2.1 & 19.6 & 10.2 & 23.4 \\
1 & 61.2 & 82.7 & 61.4 & 47.7 & 35.6 & 39.4 & 62.1 & 79.0 & 73.0 & 85.6 & 87.4 & 94.9 & 86.5 & 62.2 & 10.3 & 64.6 & 2.5 & 22.5 & 13.1 & 25.2 \\
2 & 61.7 & 82.6 & 61.4 & 48.9 & 38.5 & 39.3 & 63.6 & 79.4 & 73.8 & 85.6 & 87.9 & 94.4 & 86.6 & 64.1 & 10.5 & 65.2 & 2.6 & 24.9 & 15.2 & 25.8 \\
4 & 61.5 & 83.3 & 62.2 & 48.6 & 35.9 & 40.5 & 63.2 & 80.1 & 76.2 & 84.7 & 89.0 & 94.0 & 86.1 & 62.6 & 10.5 & 65.2 & 3.1 & 26.0 & 16.3 & 25.8 \\
8 & 62.1 & 83.8 & 63.4 & 48.5 & 38.8 & 39.7 & 63.9 & 79.6 & 76.4 & 83.2 & 89.5 & 93.8 & 86.5 & 63.3 & 10.6 & 65.5 & 2.7 & 27.3 & 16.5 & 26.2 \\
16 & 61.2 & 83.1 & 62.9 & 49.0 & 36.3 & 40.3 & 62.4 & 79.4 & 77.1 & 83.9 & 89.4 & 93.5 & 86.6 & 66.5 & 10.7 & 65.5 & 3.3 & 27.8 & 18.0 & 27.2 \\
32 & 61.2 & 84.4 & 63.8 & 49.6 & 36.8 & 38.6 & 64.0 & 78.9 & 76.6 & 82.8 & 89.5 & 93.8 & 86.0 & 63.9 & 10.5 & 65.4 & 3.1 & 28.2 & 17.9 & 26.2 \\
64 & 61.8 & 83.8 & 63.7 & 49.3 & 37.3 & 38.5 & 62.4 & 80.3 & 76.9 & 82.6 & 89.8 & 93.8 & 86.0 & 64.2 & 10.7 & 65.4 & 3.2 & 28.9 & 18.0 & 26.9 \\

\midrule
\multicolumn{21}{c}{\textbf{Stable Diffusion, CFG scale=2.0, Classname}}\\
\midrule
0.125 & 61.2 & 73.2 & 51.6 & 46.6 & 25.6 & 33.4 & 62.8 & 76.0 & 68.2 & 85.4 & 84.1 & 95.1 & 86.6 & 54.9 & 9.9 & 61.0 & 2.4 & 17.9 & 6.6 & 22.3 \\
0.25 & 65.1 & 77.5 & 56.7 & 51.1 & 33.6 & 38.6 & 66.2 & 82.3 & 76.4 & 88.6 & 87.4 & 95.1 & 88.0 & 57.2 & 10.2 & 64.9 & 2.9 & 23.9 & 10.8 & 28.7 \\
0.5 & 69.5 & 80.7 & 61.2 & 54.6 & 46.9 & 46.0 & 68.3 & 86.9 & 83.6 & 90.6 & 91.5 & 95.3 & 89.4 & 60.6 & 11.1 & 69.1 & 3.3 & 31.7 & 16.5 & 33.8 \\
1 & 72.5 & 84.9 & 65.9 & 58.8 & 55.1 & 50.6 & 70.8 & 89.1 & 87.1 & 91.9 & 94.4 & 96.0 & 90.1 & 64.3 & 12.0 & 72.2 & 5.0 & 41.0 & 23.0 & 39.1 \\
2 & 74.1 & 86.9 & 67.5 & 59.9 & 55.8 & 52.1 & 70.9 & 88.8 & 87.8 & 91.7 & 95.3 & 94.7 & 89.7 & 69.1 & 12.3 & 73.1 & 6.6 & 45.7 & 27.2 & 42.2 \\
4 & 75.4 & 87.1 & 68.4 & 60.2 & 57.2 & 52.1 & 71.2 & 89.0 & 89.2 & 91.8 & 95.8 & 95.3 & 89.6 & 67.3 & 12.1 & 73.4 & 7.9 & 49.2 & 29.6 & 44.3 \\
8 & 76.3 & 88.3 & 69.4 & 60.8 & 60.5 & 53.5 & 71.8 & 89.5 & 89.9 & 92.2 & 96.5 & 94.7 & 89.8 & 68.0 & 12.5 & 74.2 & 8.0 & 50.4 & 31.0 & 44.9 \\
16 & 76.4 & 88.6 & 69.9 & 61.6 & 59.5 & 53.9 & 72.6 & 88.2 & 89.2 & 91.8 & 96.7 & 94.5 & 90.1 & 67.5 & 12.8 & 74.2 & 8.6 & 51.8 & 31.4 & 45.6 \\
32 & 76.7 & 88.8 & 71.1 & 61.7 & 57.0 & 51.7 & 72.0 & 89.7 & 88.9 & 92.3 & 96.6 & 94.4 & 89.6 & 67.7 & 12.9 & 74.1 & 9.0 & 51.9 & 32.1 & 46.1 \\
64 & 76.6 & 88.2 & 69.6 & 61.7 & 58.5 & 53.0 & 72.0 & 89.6 & 89.8 & 91.9 & 96.6 & 95.0 & 90.1 & 68.1 & 12.9 & 74.2 & 9.4 & 52.5 & 32.4 & 46.0 \\
\midrule
\multicolumn{21}{c}{\textbf{Stable Diffusion, CFG scale=2.0, CLIP Templates(80)}}\\
\midrule
0.125 & 60.4 & 70.4 & 48.5 & 45.1 & 24.9 & 32.6 & 63.1 & 73.0 & 67.3 & 86.4 & 83.1 & 95.2 & 86.5 & 50.9 & 9.8 & 59.8 & 2.4 & 18.5 & 7.6 & 19.6 \\
0.25 & 64.0 & 74.5 & 53.7 & 49.4 & 32.2 & 37.8 & 65.1 & 80.5 & 75.3 & 88.0 & 86.7 & 95.4 & 87.2 & 56.4 & 10.1 & 63.8 & 2.7 & 29.0 & 14.8 & 27.4 \\
0.5 & 67.6 & 79.6 & 58.7 & 54.0 & 42.7 & 45.2 & 68.1 & 86.8 & 84.6 & 90.3 & 91.0 & 95.1 & 88.9 & 60.5 & 10.1 & 68.2 & 3.1 & 40.3 & 24.4 & 33.5 \\
1 & 71.7 & 85.1 & 65.5 & 58.6 & 53.6 & 50.1 & 70.6 & 88.0 & 88.8 & 92.6 & 94.7 & 96.0 & 89.3 & 65.4 & 11.3 & 72.1 & 5.1 & 52.9 & 33.5 & 40.4 \\
2 & 75.0 & 87.8 & 69.3 & 61.8 & 61.0 & 55.1 & 72.6 & 90.3 & 90.9 & 93.7 & 96.5 & 95.4 & 90.3 & 68.5 & 12.0 & 74.7 & 7.5 & 61.9 & 39.7 & 45.1 \\
4 & 76.3 & 89.7 & 71.9 & 62.9 & 63.3 & 55.3 & 74.3 & 91.8 & 92.3 & 94.3 & 96.6 & 94.9 & 91.1 & 68.9 & 12.8 & 75.8 & 9.5 & 66.4 & 42.6 & 47.4 \\
8 & 76.9 & 90.1 & 71.6 & 63.0 & 61.7 & 56.2 & 73.9 & 90.7 & 91.0 & 93.5 & 96.7 & 95.1 & 89.6 & 67.6 & 12.4 & 75.3 & 10.5 & 67.5 & 43.8 & 48.7 \\
16 & 77.2 & 89.7 & 72.1 & 63.0 & 63.5 & 56.0 & 73.8 & 90.9 & 91.3 & 92.0 & 97.0 & 94.5 & 89.7 & 67.6 & 12.4 & 75.4 & 10.8 & 68.8 & 44.3 & 49.3 \\
32 & 77.5 & 90.2 & 71.9 & 62.6 & 62.5 & 56.3 & 73.5 & 90.8 & 91.5 & 93.2 & 96.6 & 94.9 & 89.9 & 67.8 & 12.3 & 75.4 & 11.4 & 69.0 & 44.5 & 49.2 \\
64 & 77.5 & 90.5 & 73.1 & 62.8 & 61.6 & 55.7 & 73.5 & 90.9 & 91.8 & 93.0 & 97.2 & 94.7 & 90.1 & 67.3 & 12.6 & 75.5 & 11.5 & 69.3 & 44.7 & 49.6 \\

\midrule
\multicolumn{21}{c}{\textbf{Stable Diffusion, CFG scale=2.0, IN Caption}}\\
\midrule
0.125 & 59.5 & 71.3 & 49.2 & 45.7 & 22.6 & 31.6 & 59.5 & 73.5 & 64.5 & 84.3 & 82.6 & 95.0 & 86.1 & 48.3 & 9.8 & 58.9 & 2.3 & 13.9 & 5.0 & 22.5 \\
0.25 & 63.1 & 75.7 & 54.6 & 50.3 & 26.7 & 36.6 & 64.6 & 81.0 & 73.0 & 86.9 & 88.0 & 95.1 & 86.7 & 55.4 & 10.3 & 63.2 & 3.0 & 19.1 & 8.1 & 29.5 \\
0.5 & 67.6 & 79.8 & 58.5 & 54.1 & 38.3 & 41.7 & 66.6 & 86.6 & 80.4 & 90.0 & 91.2 & 95.8 & 88.4 & 60.7 & 10.9 & 67.4 & 4.4 & 25.7 & 12.8 & 36.7 \\
1 & 71.9 & 84.3 & 64.2 & 58.8 & 48.3 & 49.1 & 69.8 & 89.2 & 86.4 & 91.1 & 94.8 & 95.5 & 89.4 & 60.7 & 11.9 & 71.0 & 7.2 & 35.7 & 19.5 & 45.3 \\
2 & 76.0 & 88.0 & 68.9 & 62.4 & 57.5 & 51.4 & 72.2 & 90.5 & 88.6 & 93.0 & 96.6 & 95.5 & 90.4 & 64.5 & 13.0 & 73.9 & 10.1 & 43.8 & 25.9 & 50.7 \\
4 & 77.2 & 88.8 & 69.2 & 62.5 & 56.2 & 54.0 & 72.0 & 90.7 & 89.4 & 92.8 & 96.8 & 95.2 & 90.0 & 64.5 & 13.0 & 74.2 & 12.5 & 48.4 & 28.6 & 52.7 \\
8 & 78.1 & 88.9 & 70.4 & 63.6 & 56.3 & 53.7 & 71.3 & 90.6 & 90.5 & 93.1 & 97.2 & 94.7 & 90.4 & 66.0 & 13.4 & 74.6 & 14.0 & 50.5 & 30.9 & 54.2 \\
16 & 78.3 & 89.5 & 71.4 & 63.8 & 56.4 & 52.2 & 72.9 & 90.8 & 89.8 & 92.3 & 97.3 & 95.0 & 89.5 & 66.0 & 13.4 & 74.6 & 15.7 & 51.4 & 30.7 & 54.8 \\
32 & 78.7 & 89.9 & 71.8 & 64.1 & 58.8 & 53.6 & 72.0 & 91.2 & 90.4 & 91.8 & 97.3 & 95.1 & 89.8 & 66.6 & 13.9 & 75.0 & 15.9 & 52.4 & 31.5 & 54.8 \\
64 & 78.4 & 89.8 & 70.9 & 64.1 & 58.9 & 53.4 & 73.0 & 91.4 & 90.7 & 92.3 & 97.3 & 94.5 & 89.7 & 67.4 & 13.7 & 75.0 & 16.0 & 53.2 & 32.2 & 55.0 \\
\bottomrule[1.2pt]
\end{tabular}}
\end{center}
\end{table*}

\begin{table*}[t]
\begin{center}
\caption{\small{Detailed scaling behavior on 15 different downstream classification datasets and ImageNet-A, ImageNet-R, ImageNet-Sketch and ImageNet-V2 validation set for supervised classifiers trained with synthetic images from various configurations using Imagen and Muse. Dataset scale is in million.}}
\label{table:appendix_scaling_imagen_muse}
\resizebox{\textwidth}{!}{
\begin{tabular}
{r|ccccccccccccccc|c|cccc}
\toprule[1.2pt]
\bf Scale&
\rotatebox[origin=lb]{90}{\smash{\small Food-101}} & \rotatebox[origin=lb]{90}{\smash{\small CIFAR-10}} & \rotatebox[origin=lb]{90}{\smash{\small CIFAR-100}} & \rotatebox[origin=lb]{90}{\smash{\small SUN397}} &
\rotatebox[origin=lb]{90}{\smash{\small Cars}} & \rotatebox[origin=lb]{90}{\smash{\small Aircraft}} & \rotatebox[origin=lb]{90}{\smash{\small DTD}} & \rotatebox[origin=lb]{90}{\smash{\small Pets}} & \rotatebox[origin=lb]{90}{\smash{\small Caltech-101}} &
\rotatebox[origin=lb]{90}{\smash{\small Flowers}} & \rotatebox[origin=lb]{90}{\smash{\small STL-10}} & \rotatebox[origin=lb]{90}{\smash{\small EuroSAT}} &
\rotatebox[origin=lb]{90}{\smash{\small RESISC45}} & \rotatebox[origin=lb]{90}{\smash{\small GTSRB}} & \rotatebox[origin=lb]{90}{\smash{\small Country211}}  & \rotatebox[origin=lb]{90}{\smash{\small \bf DS Average}} 
& \rotatebox[origin=lb]{90}{\smash{\small ImageNet-A}} & \rotatebox[origin=lb]{90}{\smash{\small ImageNet-R}} & \rotatebox[origin=lb]{90}{\smash{\small ImageNet-Sketch}} & \rotatebox[origin=lb]{90}{\smash{\small ImageNet-V2}} 
\\

\midrule
\multicolumn{21}{c}{\textbf{Imagen, CFG scale=2.0, IN Caption}}\\
\midrule
0.125 & 58.5 & 71.9 & 50.7 & 45.7 & 22.6 & 31.5 & 59.8 & 78.8 & 65.7 & 84.4 & 83.9 & 94.9 & 84.7 & 52.6 & 11.0 & 59.8 & 2.7 & 15.6 & 4.9 & 29.1 \\
0.25 & 61.8 & 78.3 & 56.8 & 50.1 & 28.7 & 36.3 & 63.9 & 84.3 & 74.6 & 86.4 & 89.8 & 94.6 & 85.3 & 54.6 & 11.1 & 63.8 & 3.2 & 22.2 & 9.4 & 35.6 \\
0.5 & 66.6 & 81.9 & 61.1 & 54.8 & 38.4 & 43.4 & 66.1 & 88.1 & 80.1 & 89.5 & 93.4 & 95.4 & 86.1 & 58.2 & 11.3 & 67.6 & 4.2 & 29.0 & 13.7 & 42.4 \\
1 & 71.3 & 86.3 & 66.8 & 59.6 & 49.5 & 50.3 & 69.6 & 90.9 & 85.9 & 90.0 & 96.0 & 95.4 & 88.0 & 61.3 & 12.6 & 71.6 & 7.9 & 39.4 & 19.6 & 51.1 \\
2 & 73.8 & 90.0 & 70.9 & 60.2 & 50.2 & 52.1 & 68.0 & 90.9 & 87.2 & 89.7 & 97.2 & 95.1 & 86.3 & 62.0 & 12.9 & 72.4 & 11.9 & 45.0 & 23.2 & 55.0 \\
4 & 75.1 & 90.2 & 71.8 & 61.5 & 49.5 & 53.7 & 69.1 & 91.2 & 87.8 & 89.6 & 97.4 & 94.7 & 87.1 & 62.8 & 13.0 & 73.0 & 14.2 & 49.7 & 27.1 & 57.6 \\
8 & 75.0 & 91.2 & 72.8 & 62.0 & 52.4 & 51.7 & 67.5 & 91.7 & 88.0 & 88.3 & 98.1 & 95.0 & 86.4 & 62.1 & 13.4 & 73.0 & 16.4 & 51.0 & 27.3 & 58.7 \\

\midrule
\multicolumn{21}{c}{\textbf{Muse, CFG scale=2.0, IN Caption}}\\
\midrule
0.125 & 58.3 & 76.6 & 54.9 & 45.3 & 20.5 & 29.6 & 62.0 & 73.2 & 66.6 & 84.8 & 87.5 & 95.2 & 83.3 & 52.2 & 10.7 & 60.0 & 2.7 & 16.8 & 6.8 & 23.4 \\
0.25 & 63.3 & 84.1 & 63.5 & 50.2 & 27.0 & 35.9 & 65.3 & 80.3 & 76.5 & 87.5 & 92.5 & 95.5 & 85.3 & 59.4 & 11.2 & 65.2 & 3.6 & 25.4 & 13.2 & 30.5 \\
0.5 & 67.0 & 87.0 & 68.1 & 54.0 & 37.3 & 42.4 & 67.7 & 85.1 & 82.2 & 89.1 & 94.5 & 95.5 & 86.4 & 64.8 & 11.4 & 68.8 & 4.6 & 32.7 & 19.8 & 36.4 \\
1 & 71.2 & 91.3 & 72.8 & 58.9 & 48.1 & 47.6 & 70.4 & 87.2 & 87.0 & 92.1 & 96.8 & 95.4 & 87.2 & 66.9 & 11.8 & 72.3 & 7.7 & 43.3 & 27.9 & 44.1 \\
2 & 73.9 & 92.0 & 74.3 & 60.3 & 49.7 & 49.3 & 70.6 & 89.4 & 87.6 & 91.0 & 97.1 & 94.6 & 87.3 & 67.9 & 12.7 & 73.2 & 12.2 & 48.8 & 32.7 & 48.4 \\
4 & 74.9 & 92.9 & 74.4 & 60.7 & 51.4 & 51.0 & 71.1 & 89.6 & 88.4 & 91.5 & 97.7 & 94.7 & 86.9 & 68.3 & 12.6 & 73.7 & 14.4 & 52.0 & 34.8 & 50.3 \\
8 & 75.3 & 91.7 & 73.3 & 61.1 & 51.4 & 51.9 & 70.5 & 89.6 & 87.7 & 91.6 & 98.1 & 93.5 & 86.9 & 66.1 & 12.8 & 73.4 & 15.7 & 53.2 & 35.6 & 51.0 \\
\bottomrule[1.2pt]
\end{tabular}}
\end{center}
\end{table*}

In Tables~\ref{table:appendix_scaling} and~\ref{table:appendix_scaling_imagen_muse}, we present a comparison of the scaling behavior of supervised models trained under various configurations. This comparison is based on linear probing performed on 15 fine-grained classification datasets, as detailed in Appendix~\ref{sec:appendix-ds-linear}. Our findings indicate that, in general, the scaling behavior observed in linear probing on these downstream datasets aligns with the trends seen in the ImageNet validation set. However, there are  instances where training on synthetic images surpasses the performance of training on real images, in the Food-101 dataset for example.

Additionally, we have also included the detailed comparison on the out-of-distribution (OOD) validation sets, including ImageNet-A~\cite{hendrycks2021nae}, ImageNet-R~\cite{hendrycks2021many}, ImageNet-Sketch~\cite{wang2019learning}, and Imagenet-V2~\cite{recht2019imagenet}. The results from these comparisons demonstrate that training on synthetic images can yield improved performance on OOD test sets, exemplified by the results on ImageNet-R.

\section{Visualization on Generated Images}
\label{sec:appendix-prompt}
To better understand the impact of various text prompts used in generating training images, we provide additional visualizations of images created using different text prompt configurations for specific ImageNet categories. These visualizations were generated using Stable Diffusion, with the CFG scale set to 2.

\begin{figure*}[t]
  \centering
  \begin{minipage}[b]{\textwidth}
    \centering
              \small \textbf{(a) Classnames}
    \begin{tabular}{@{\hspace{0em}}p{0.32\textwidth}@{\hspace{0.5em}}p{0.32\textwidth}@{\hspace{0.5em}}p{0.32\textwidth}}
    \small{goldfish} &  \small{Golden Retriever} & \small{shopping cart}\\
    \includegraphics[width=\linewidth]{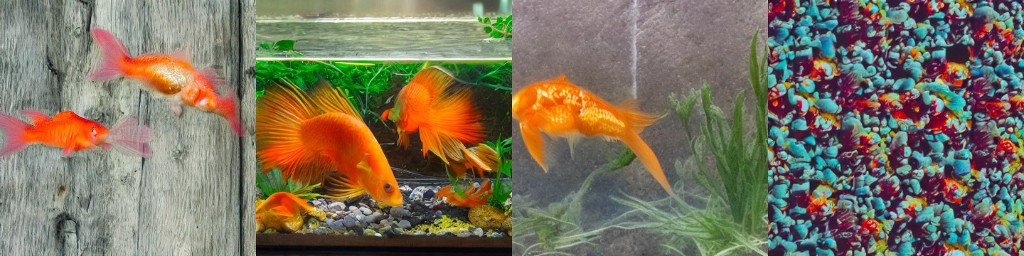}
    & \includegraphics[width=\linewidth]{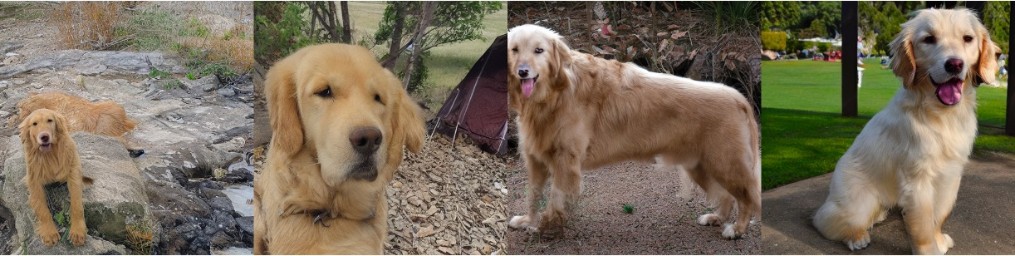} & \includegraphics[width=\linewidth]{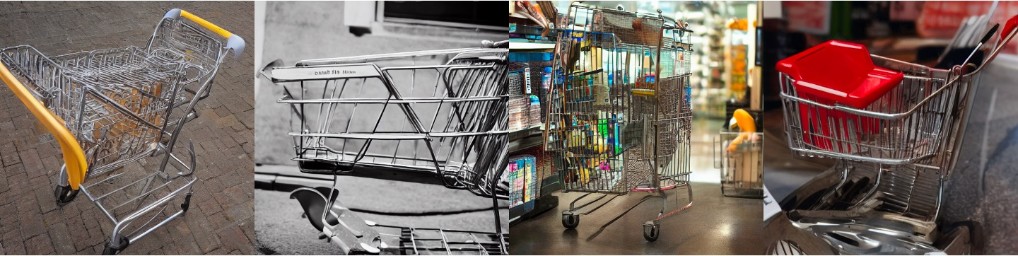}\\
        \end{tabular}
    \end{minipage}
  
  \begin{minipage}[b]{\textwidth}
    \centering
                 \small \textbf{(b) Classname + Description}
    \begin{tabular}  {@{\hspace{0em}}p{0.32\textwidth}@{\hspace{0.5em}}p{0.32\textwidth}@{\hspace{0.5em}}p{0.32\textwidth}}
    \small{goldfish, Carassius auratus, small golden or orange-red freshwater fishes of Eurasia used as pond or aquarium fishes} &  \small{Golden Retriever, an English breed having a long silky golden coat} & \small{shopping cart, a handcart that holds groceries or other goods while shopping}\\
    \includegraphics[width=\linewidth]{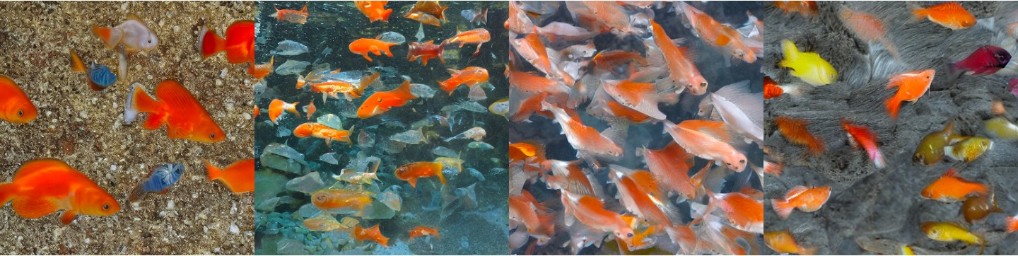}
    & \includegraphics[width=\linewidth]{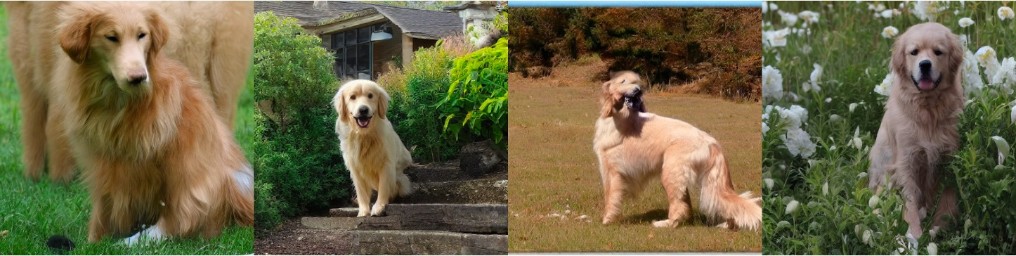} & \includegraphics[width=\linewidth]{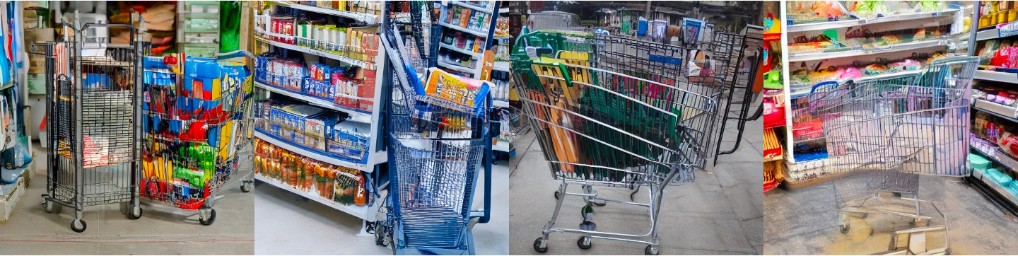}\\
    \end{tabular}
    \end{minipage}

      \begin{minipage}[b]{\textwidth}
    \centering
                  \small \textbf{(c) Classname + Description + Places}
    \begin{tabular}  {@{\hspace{0em}}p{0.32\textwidth}@{\hspace{0.5em}}p{0.32\textwidth}@{\hspace{0.5em}}p{0.32\textwidth}}
    \small{goldfish, Carassius auratus, small golden or orange-red freshwater fishes of Eurasia used as pond or aquarium fishes inside ice skating rink} &  \small{Golden Retriever, an English breed having a long silky golden coat inside gas station} & \small{shopping cart, a handcart that holds groceries or other goods while shopping inside basketball court}\\
    \includegraphics[width=\linewidth]{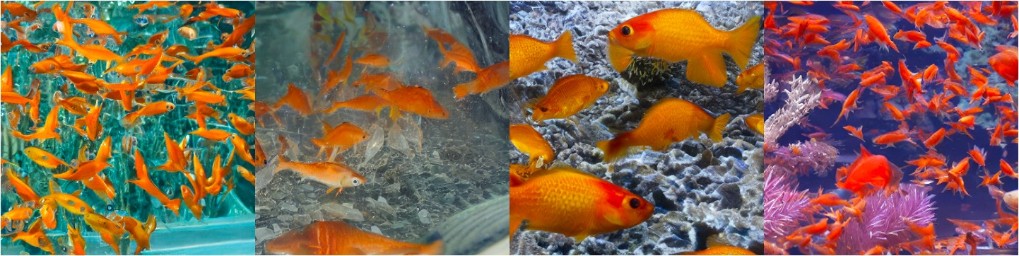}
    & \includegraphics[width=\linewidth]{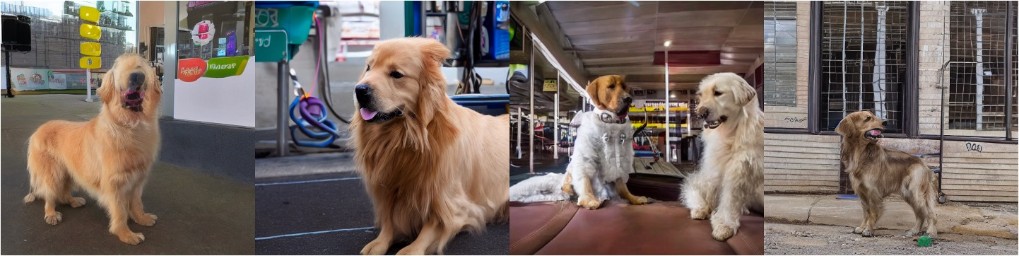} & \includegraphics[width=\linewidth]{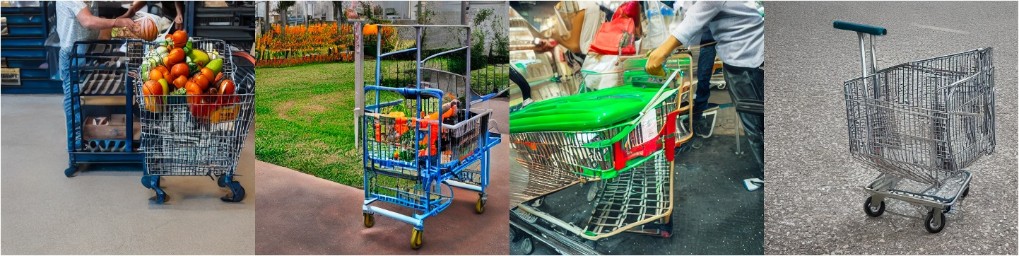}\\
    \end{tabular}
    \end{minipage}

      \begin{minipage}[b]{\textwidth}
    \centering
                  \small \textbf{(d) Classname + Hypernym}
    \begin{tabular}  {@{\hspace{0em}}p{0.32\textwidth}@{\hspace{0.5em}}p{0.32\textwidth}@{\hspace{0.5em}}p{0.32\textwidth}}
    \small{goldfish, Carassius auratus, cyprinid} &  \small{Golden Retriever, retriever} & \small{shopping cart, handcart, pushcart, cart, go-cart}\\
    \includegraphics[width=\linewidth]{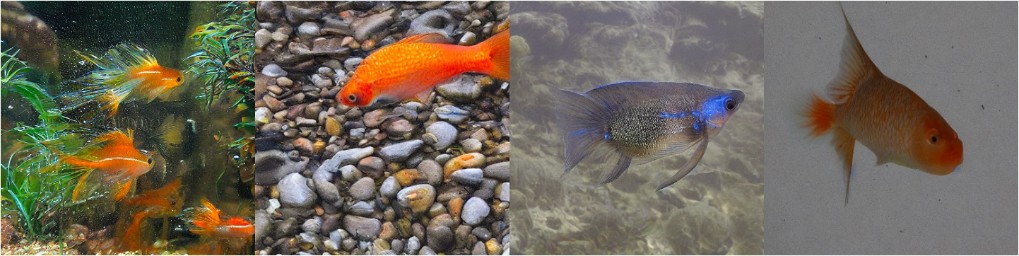}
    & \includegraphics[width=\linewidth]{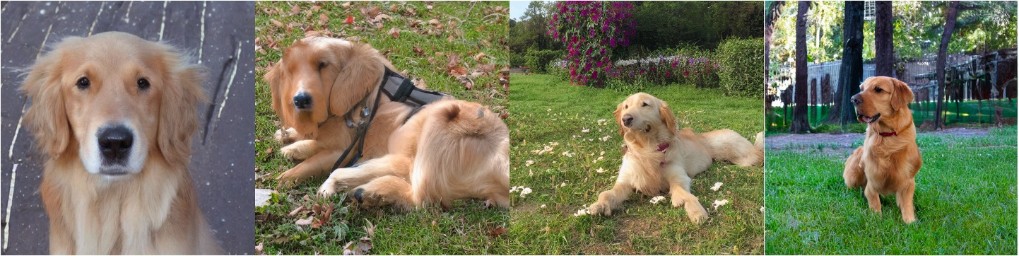} & \includegraphics[width=\linewidth]{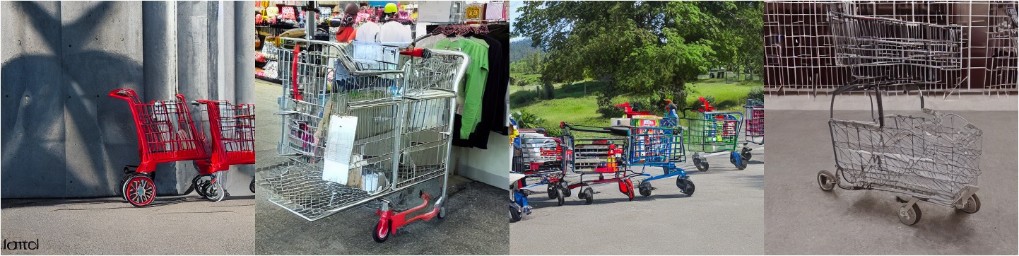}\\
    \end{tabular}
    \end{minipage}

      \begin{minipage}[b]{\textwidth}
    \centering
                  \small \textbf{(e) Classname + Hypernym + Places}
    \begin{tabular}  {@{\hspace{0em}}p{0.32\textwidth}@{\hspace{0.5em}}p{0.32\textwidth}@{\hspace{0.5em}}p{0.32\textwidth}}
    \small{goldfish, Carassius auratus, cyprinid inside ice skating rink} &  \small{Golden Retriever, retriever inside amusement park} & \small{shopping cart, handcart, pushcart, cart, go-cart inside airfield}\\
    \includegraphics[width=\linewidth]{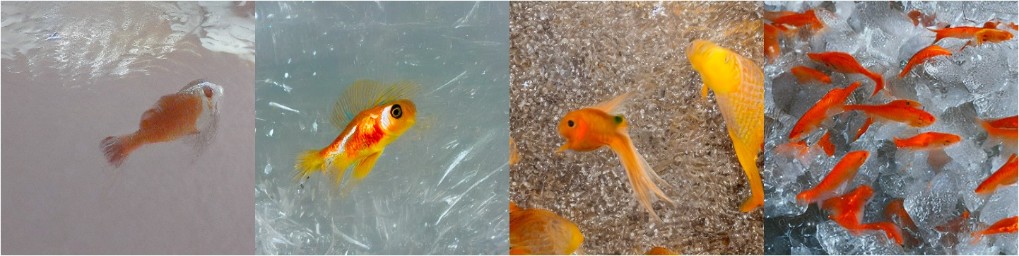}
    & \includegraphics[width=\linewidth]{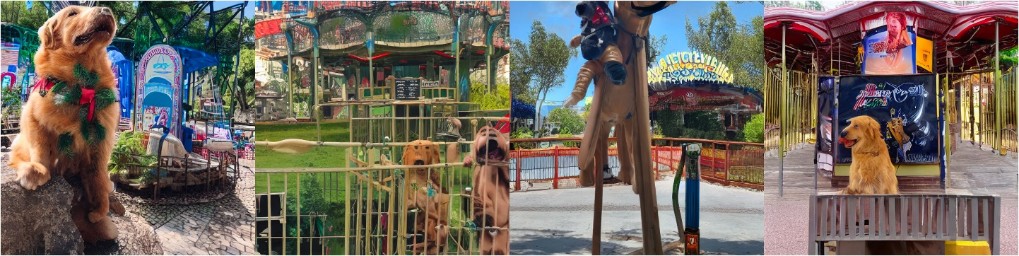} & \includegraphics[width=\linewidth]{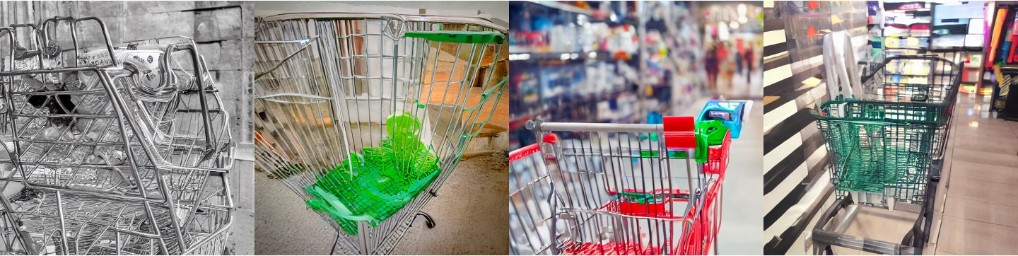}\\
    \end{tabular}
    \end{minipage}

      \begin{minipage}[b]{\textwidth}
    \centering
                  \small \textbf{(f) Word2Sen}
    \begin{tabular}  {@{\hspace{0em}}p{0.32\textwidth}@{\hspace{0.5em}}p{0.32\textwidth}@{\hspace{0.5em}}p{0.32\textwidth}}
    \small{goldenfish on a shelf} &  \small{The golden retriever on a run} & \small{the cart is a perfect place for shopping}\\
    \includegraphics[width=\linewidth]{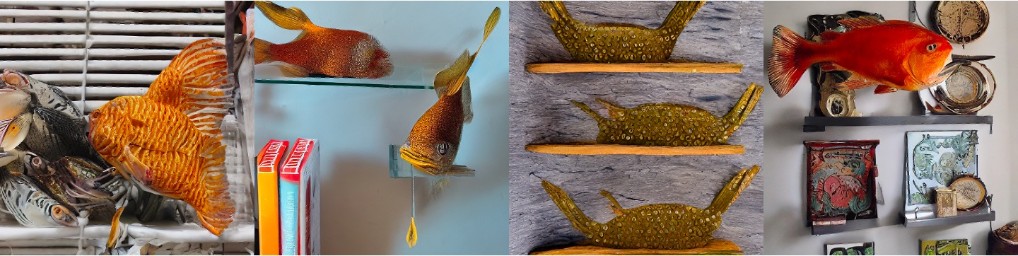}
    & \includegraphics[width=\linewidth]{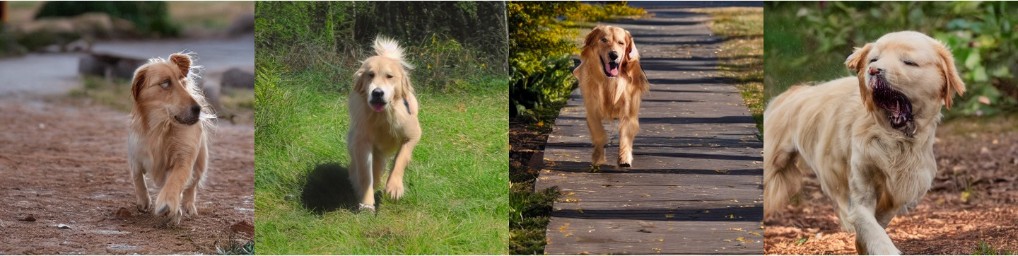} & \includegraphics[width=\linewidth]{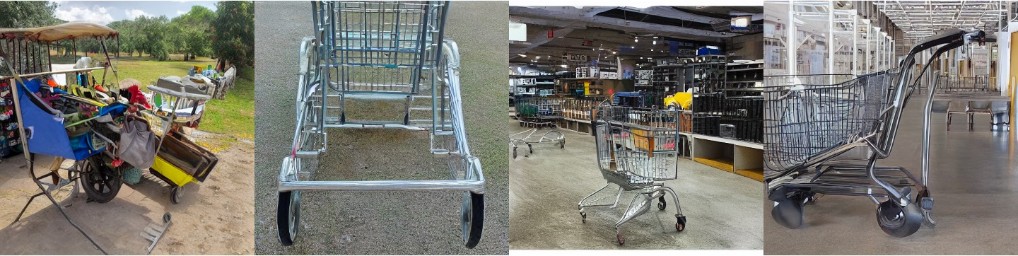}\\
    \end{tabular}
    \end{minipage}

      \begin{minipage}[b]{\textwidth}
    \centering
                  \small \textbf{(g) CLIP templates}
    \begin{tabular}  {@{\hspace{0em}}p{0.32\textwidth}@{\hspace{0.5em}}p{0.32\textwidth}@{\hspace{0.5em}}p{0.32\textwidth}}
    \small{A good photo of the golden fish} &  \small{A close-up photo of a Golden Retriever} & \small{A photo of the cool shopping cart}\\
    \includegraphics[width=\linewidth]{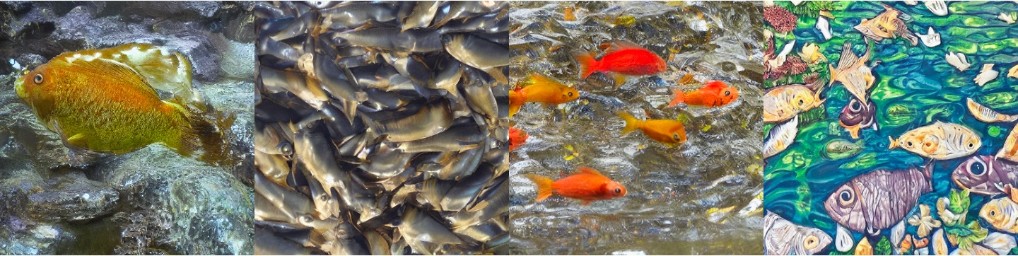}
    & \includegraphics[width=\linewidth]{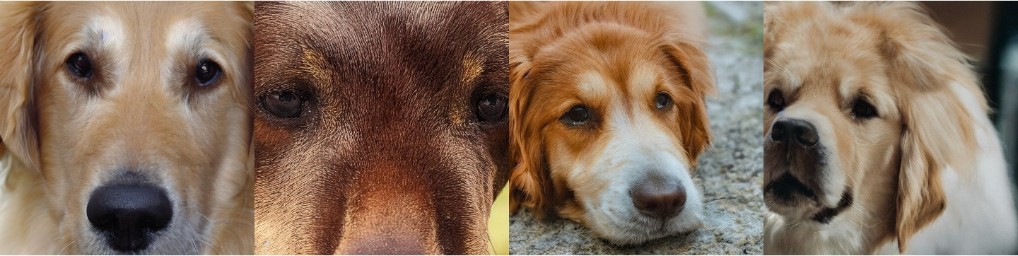} & \includegraphics[width=\linewidth]{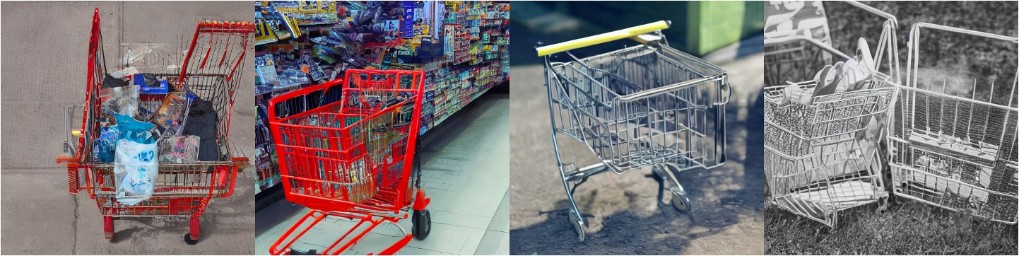}\\
    \end{tabular}
    \end{minipage}

      \begin{minipage}[b]{\textwidth}
    \centering
                  \small \textbf{(h) IN-Caption}
    \begin{tabular}  {@{\hspace{0em}}p{0.32\textwidth}@{\hspace{0.5em}}p{0.32\textwidth}@{\hspace{0.5em}}p{0.32\textwidth}}
    \small{goldfish, fish is swimming in the water
} &  \small{Golden Retriever, a dog is running on the grass} & \small{shopping cart, a street with a garbage can and a cart}\\
    \includegraphics[width=\linewidth]{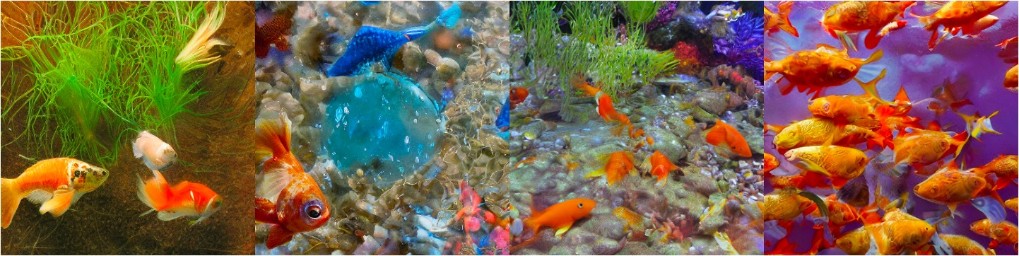}
    & \includegraphics[width=\linewidth]{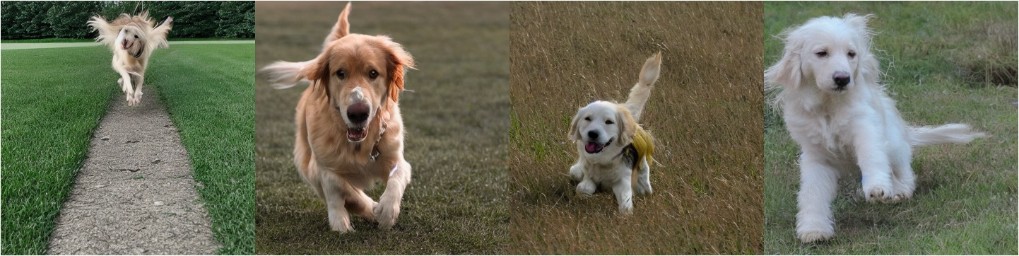} & \includegraphics[width=\linewidth]{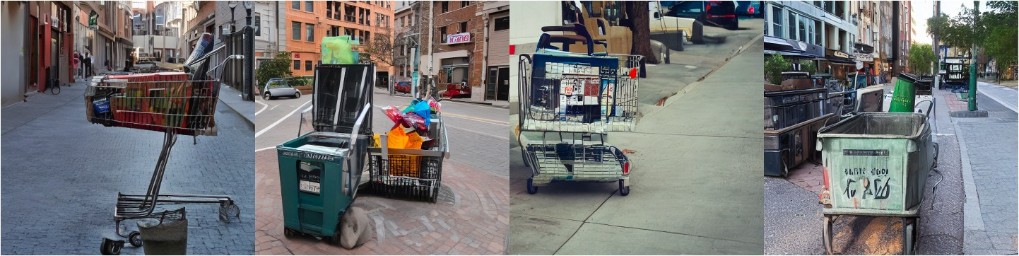}\\
    \end{tabular}
    \end{minipage}

  \caption{\small Synthetic images generated by Stable Diffusion with different text prompt configurations on three ImageNet categories: goldfish, Golden Retriever, and shopping cart. All of these visualizations use a guidance scale of 2.0.}
  \label{fig:appendix-prompt}
\end{figure*}

In Figure~\ref{fig:appendix-prompt}, we present a detailed visualization of the images generated with  different text prompt configurations for three different ImageNet categories: Goldfish, Golden Retriever, and shopping carts. The visualizations illustrate that incorporating more detailed information into the prompt tends to encourage the text-to-image model to generate more diverse images. However, this increased diversity may potentially compromise the accuracy of the category of interest in the generated images.

\section{More per-class analysis}
\subsection{Recognizablity Distribution}
\begin{figure}[h]
\centering
\includegraphics[width=1.0\linewidth]{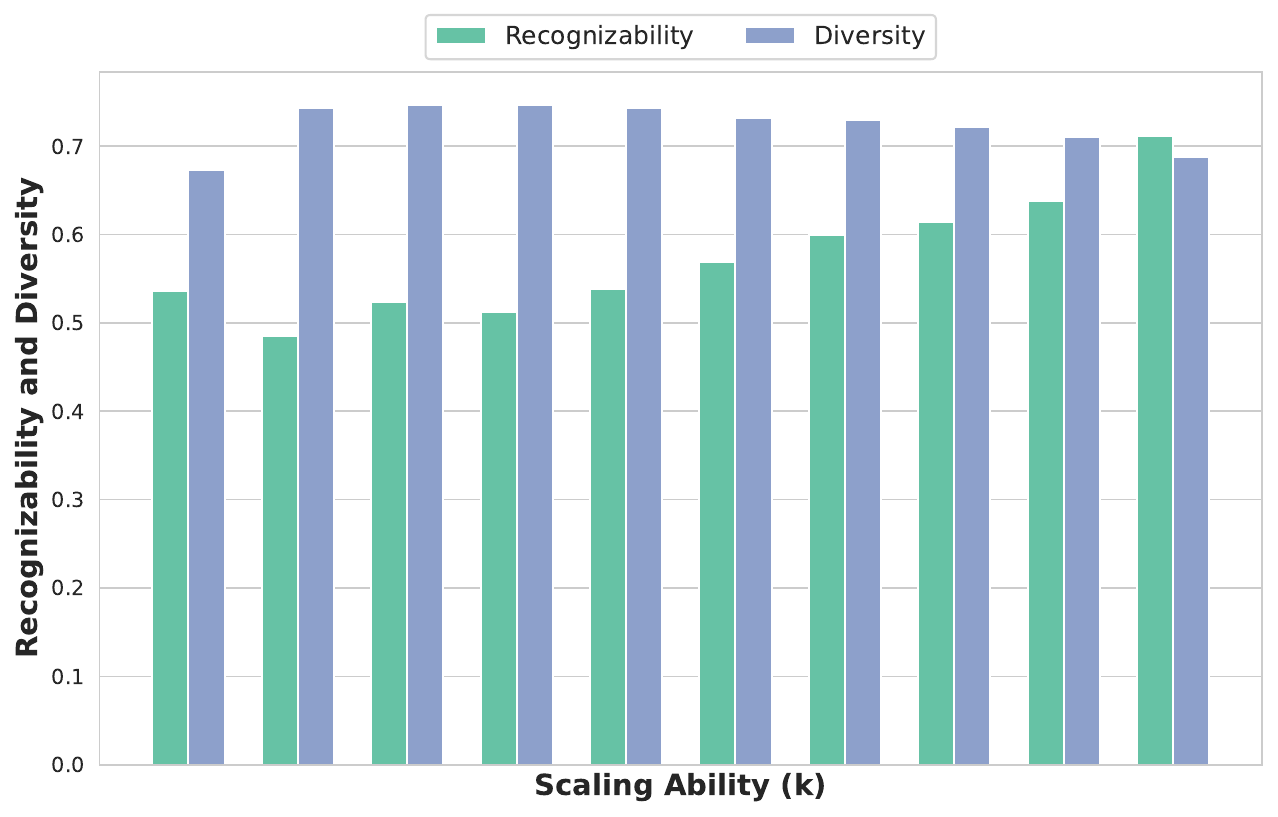}
\caption{
\small
Per class analysis on the changes in recognizability and diversity as the scaling ability ($k$) increases. Here we divide the $1000$ ImageNet classes into 10 groups based on their scaling ability ranging from the lowest to the highest.}
\label{fig:appendix-scalingbarplot}
\end{figure}

To delve deeper into how recognizability and diversity are distributed across the 1000 ImageNet classes and their influence on scaling ability ($k$), we categorized all classes into 10 different groups based on their scaling ability, ranging from lowest to highest. For each group, we calculated the average recognizability and diversity of the classes within it. The result of this analysis is illustrated in a detailed bar plot in Figure~\ref{fig:appendix-scalingbarplot}.

The analysis reveals the following trend: as the scaling ability of a group increases, the average diversity initially rises and then begins to decrease. This trend suggests that at the initial stages, enhanced diversity contributes to the generation of more varied synthetic images, helping the supervised classifier to learn more robust features during training. However, beyond a certain point, further increases in diversity can be harmful, potentially compromising the accuracy of the generated images or leading to the omission of key objects or the generation of wrong concepts.

In contrast, the average recognizability consistently increases as scaling ability increases, indicating a stronger correlation between the scaling ability and recognizability for each class. This consistent improvement shows the significance of recognizability as a more relevant metric for class-based analysis.


\subsection{More Results on `Scaling' Classes}
\label{sec:appendix-scaling-classes}
\begin{figure*}[h]
\centering
\includegraphics[width=0.93\linewidth]{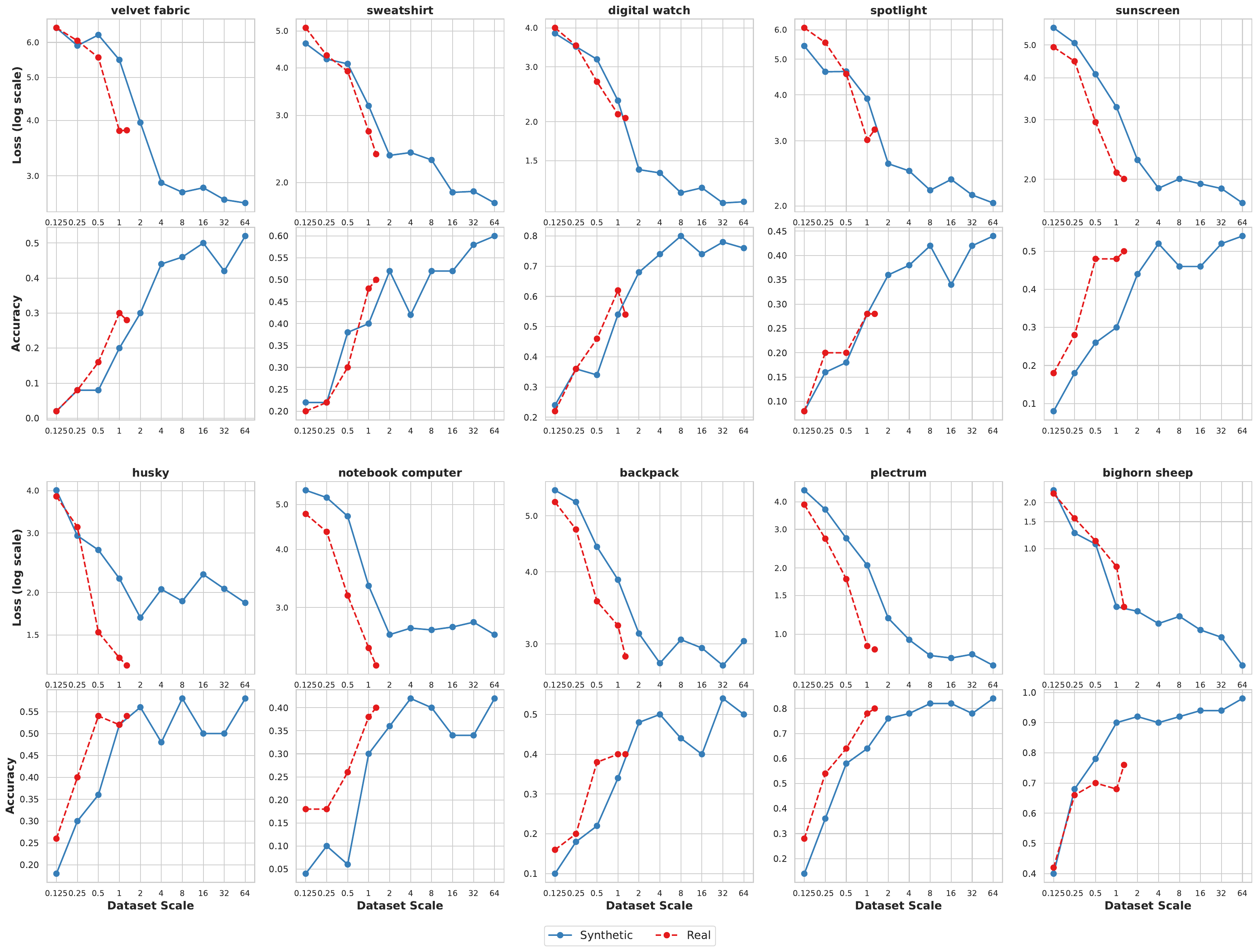}\\
\vspace*{-4mm}
\caption{
\small
More comparison on supervised models trained on real and synthetic images (from Stable Diffusion), for the `Scaling' classes.}
\label{fig:appendix-scaling_classes}

\vspace*{2mm}
  \includegraphics[width=.97\linewidth]{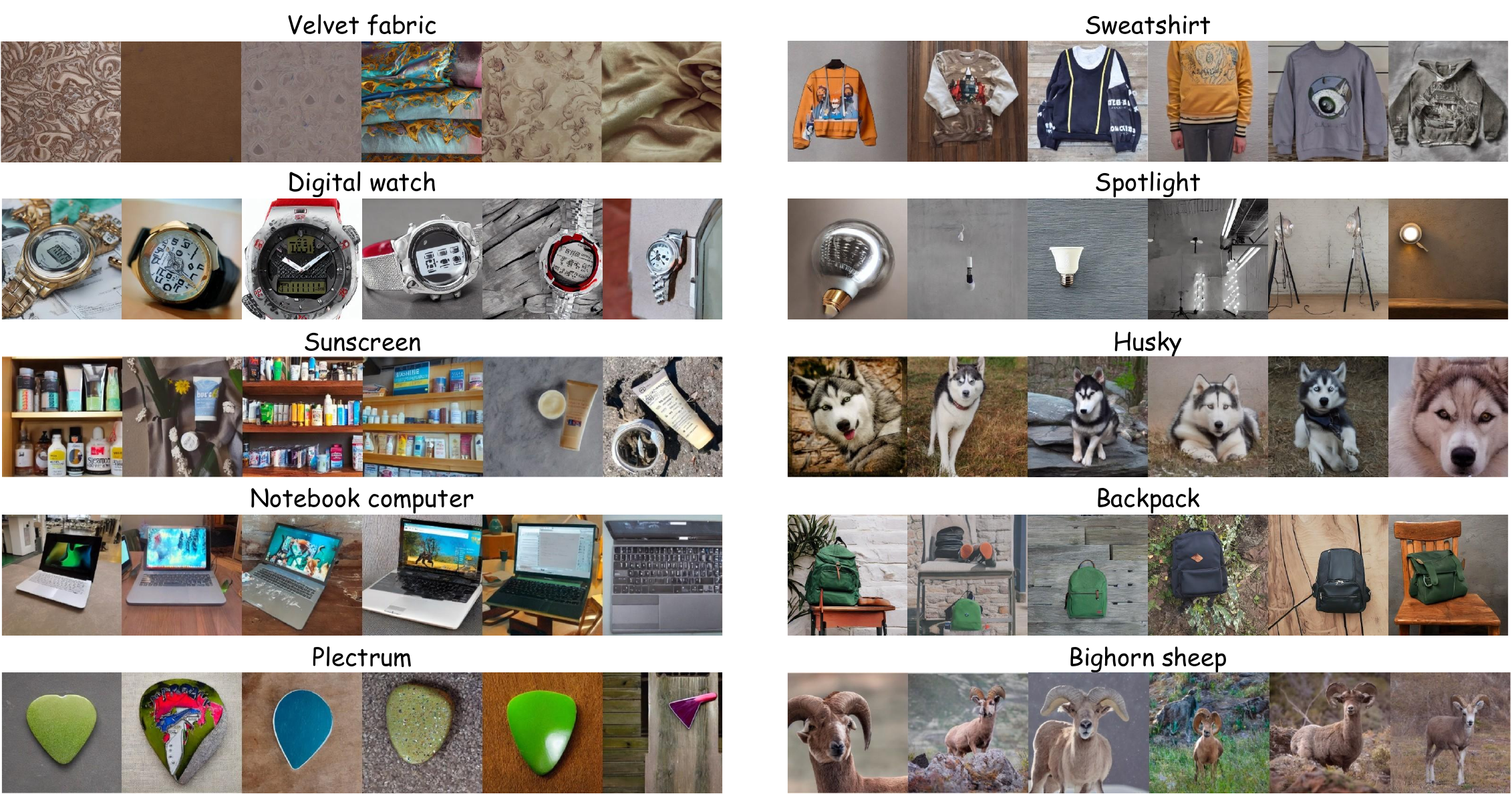}\\
\vspace*{-2mm}
\caption{
\small
Visualizations of the  synthetic images generated for `Scaling' classes, using Stable Diffusion with a guidance scale of 2.0.}
\label{fig:appendix-vis-scaling_classes}

\end{figure*}

\begin{figure*}[h]
\centering
\includegraphics[width=0.93\linewidth]{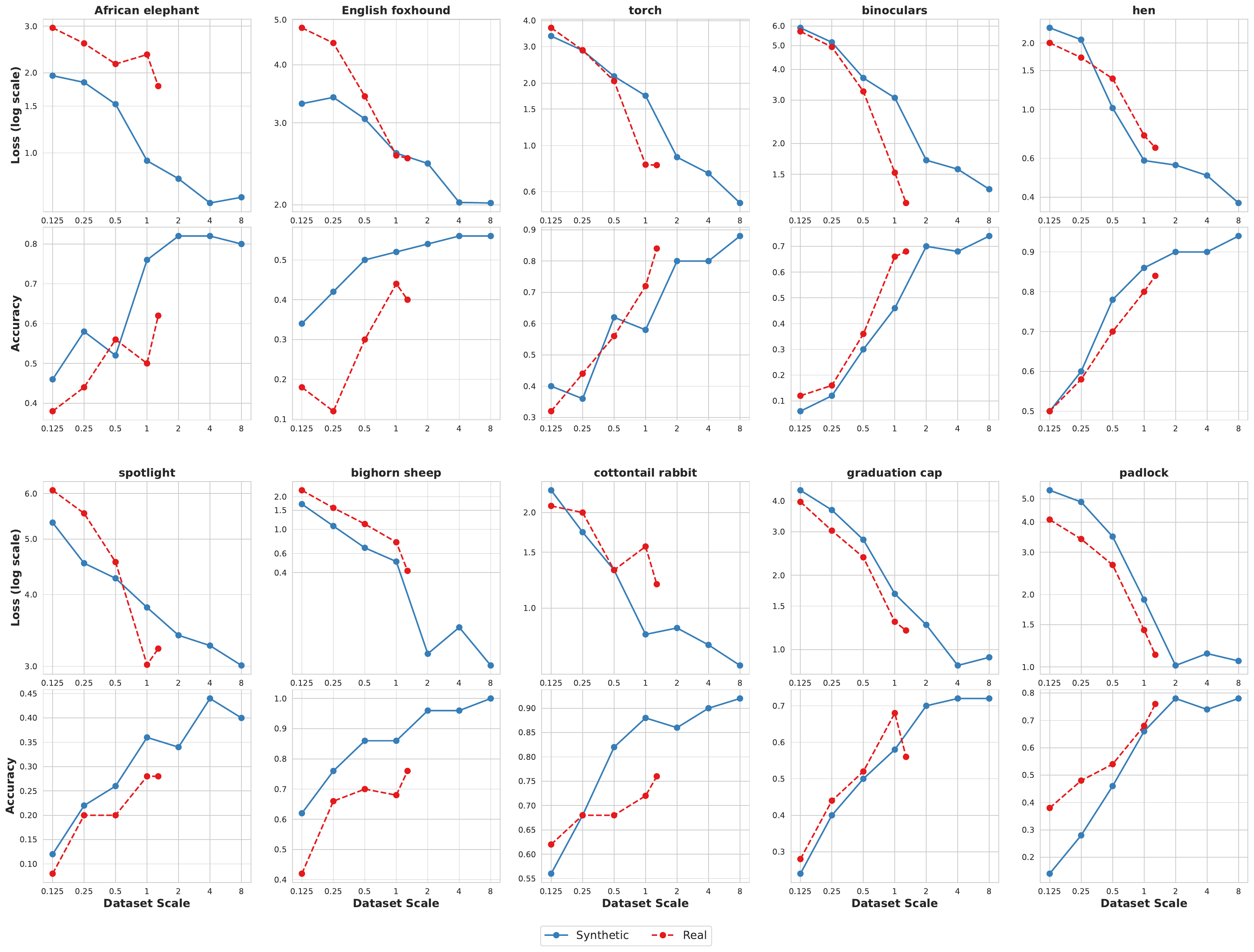}\\
\vspace*{-4mm}
\caption{
\small
More comparison on supervised models trained on real and synthetic images (from Imagen), for the `Scaling' classes.}
\label{fig:appendix-scaling_classes_imagen}

\vspace*{1mm}
  \includegraphics[width=.97\linewidth]{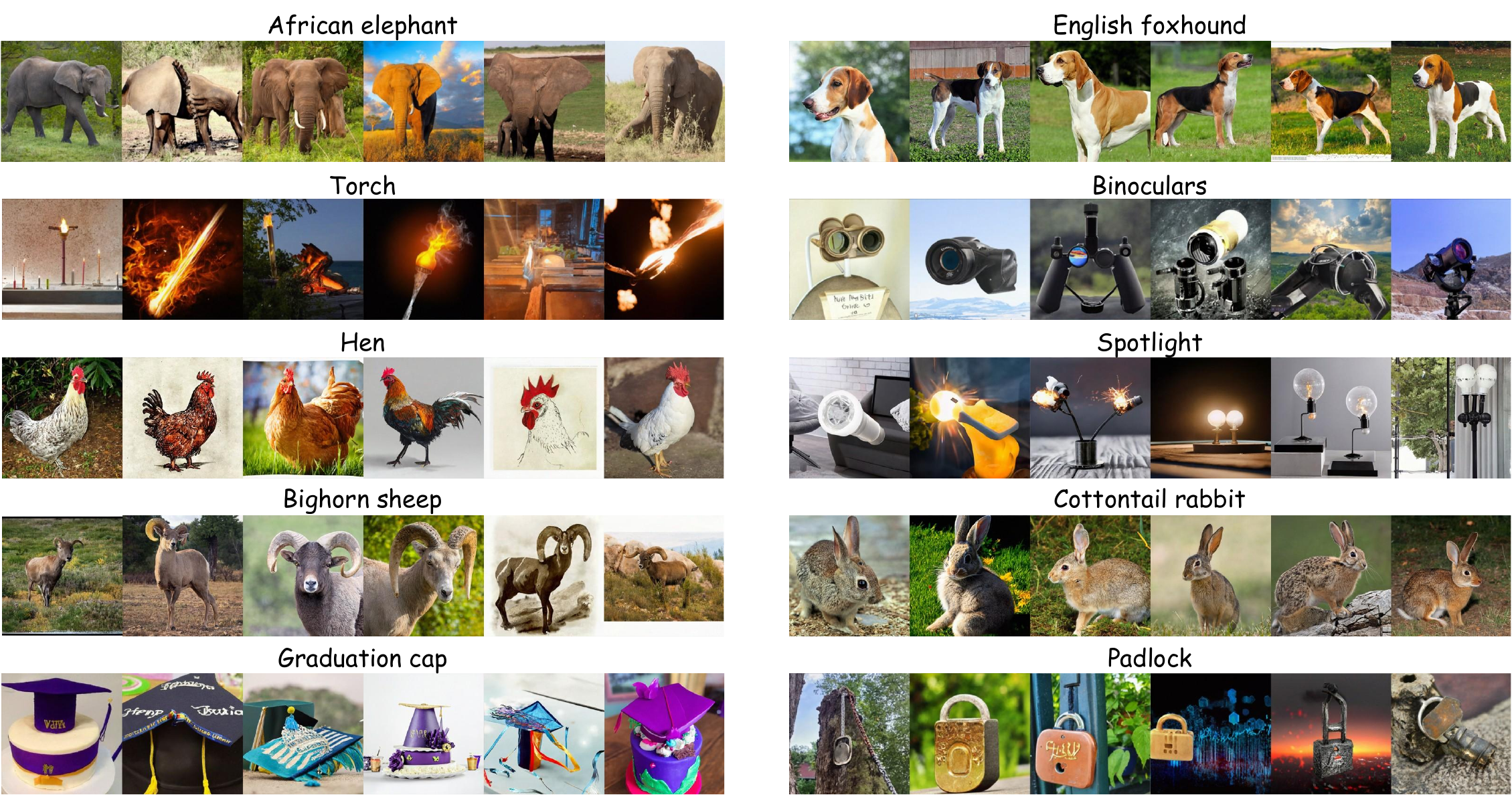}\\
\vspace*{-2mm}
\caption{
\small
Visualizations of the  synthetic images generated for `Scaling' classes, using Imagen~\cite{saharia2022photorealistic} with a guidance scale of 1.5.}
\label{fig:appendix-vis-scaling_classes_imagen}

\end{figure*}

\begin{figure*}[h]
\centering
\includegraphics[width=0.93\linewidth]{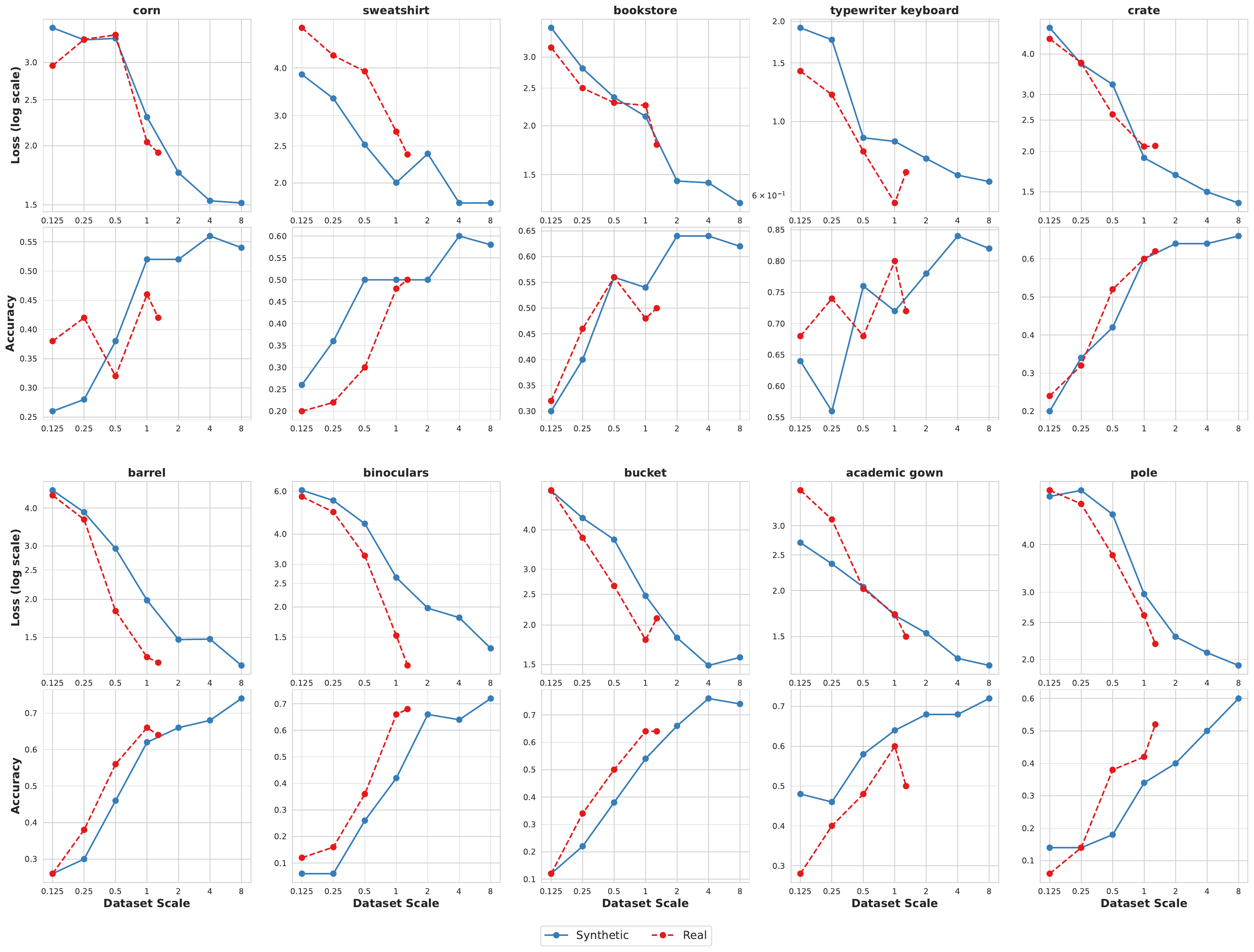}\\
\vspace*{-4mm}
\caption{
\small
More comparison on supervised models trained on real and synthetic images (from Muse), for the `Scaling' classes.}
\label{fig:appendix-scaling_classes_muse}

\vspace*{1mm}
  \includegraphics[width=.97\linewidth]{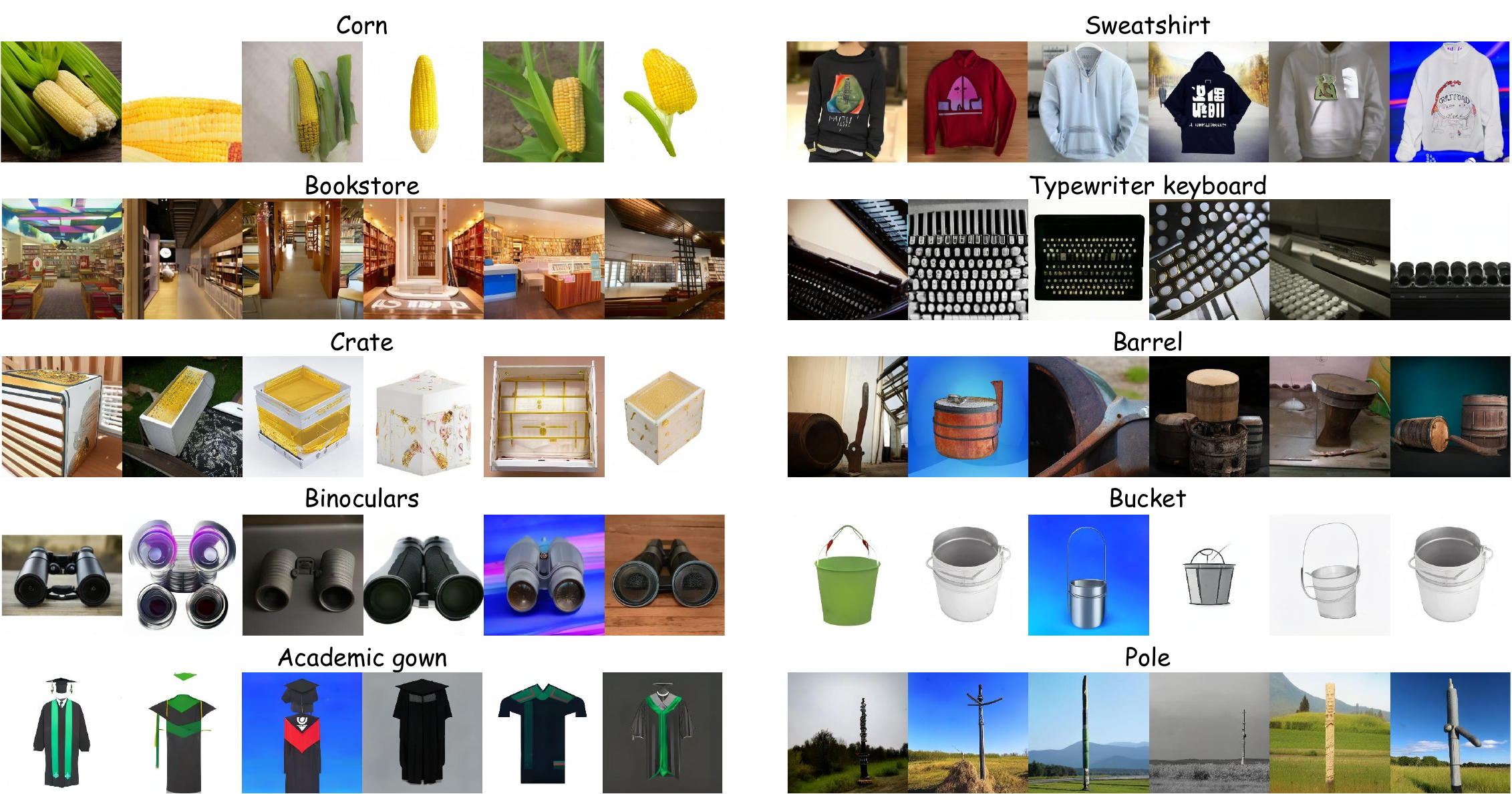}\\
\vspace*{-2mm}
\caption{
\small
Visualizations of the  synthetic images generated for `Scaling' classes, using Muse~\cite{chang2023muse} with a guidance scale of 0.3.}
\label{fig:appendix-vis-scaling_classes_muse}

\end{figure*}

In Figure~\ref{fig:appendix-scaling_classes}, we provide a detailed comparison of the scaling behavior for models trained on either real or synthetic images from Stable Diffusion, specifically focusing on the `scaling' classes as described in Section~\ref{sec:poor-class}. Additionally, Figure~\ref{fig:appendix-vis-scaling_classes} presents visualizations of synthetic images generated for these classes, using the same setup as described in Section~\ref{sec:perclass}.

We further explore 'Scaling' classes for supervised classifiers trained on images generated by the Imagen and Muse models. The scaling behaviors of these classes, in comparison to models trained with real images, along with their visualizations, are presented in Figure~\ref{fig:appendix-scaling_classes_imagen}, Figure~\ref{fig:appendix-vis-scaling_classes_imagen}, Figure~\ref{fig:appendix-scaling_classes_muse} and Figure~\ref{fig:appendix-vis-scaling_classes_muse}. Our analysis reveals that certain classes, such as sweatshirts, exhibit consistently good scaling across different text-to-image models. Meanwhile, there are classes that show particularly strong scaling performance with specific text-to-image models.

For the ten `scaling' classes selected in Stable Diffusion, we observed that models trained on synthetic images exhibit scaling abilities that are comparable to, and in some cases even superior to, those trained on real images. A notable example can be seen in the `bighorn sheep' and `spotlight' categories, where models trained on synthetic images already outperform those trained on real images at dataset scales below 1 million, and this advantage continues to grow as the scale increases, since there are only 1.3M real images.

This finding suggests that for certain concepts, text-to-image models are indeed capable of generating images that are more conducive to train supervised classifiers effectively. As text-to-image models continue to improve, we anticipate that such instances will become more frequent. Eventually, it's plausible that models trained on synthetic images could surpass the performance of those trained on real images across the entire ImageNet validation set.

\subsection{More `Poor' Classes}
\begin{table*}
\centering
\caption{\small{Lists of `poor' classes that has poor scaling ability and performance. Supervised models trained with synthetic images struggles in classifying them correctly. We list 40 categories for Stable Diffusion, Imagen and Muse, respectively.}}
\vspace{0mm}
\label{table:appendix-poor-clases}
\subfloat[
\small
Stable Diffusion
]{
\resizebox{.95\textwidth}{!}{
    \begin{tabular}{|p{\dimexpr 0.2\textwidth-2\tabcolsep-1.25\arrayrulewidth\relax}|
                    p{\dimexpr 0.2\textwidth-2\tabcolsep-1.25\arrayrulewidth\relax}|
                    p{\dimexpr 0.2\textwidth-2\tabcolsep-1.25\arrayrulewidth\relax}|
                    p{\dimexpr 0.2\textwidth-2\tabcolsep-1.25\arrayrulewidth\relax}|
                    p{\dimexpr 0.2\textwidth-2\tabcolsep-1.25\arrayrulewidth\relax}|}
        \hline
        fire salamander & Appenzeller Sennenhund & tiger cat & collie & Australian Terrier \\
        \hline
        African bush elephant & cassette player & canoe & European green lizard & night snake \\
        \hline
        mushroom & eastern hog-nosed snake & hot tub & wall clock & crayfish \\
        \hline
        espresso machine & water jug & toy terrier & Brittany dog & keyboard space bar \\
        \hline
        shower curtain & gymnastic horizontal bar & African rock python & letter opener & ladle \\
        \hline
        tape player & tea cup & paper towel & wok & flute \\
        \hline
        vine snake & black-footed ferret & cricket insect & European polecat & cradle \\
        \hline
        Lakeland Terrier & green mamba & cleaver & breastplate & monitor \\
        \hline
    \end{tabular}
}
}
\vspace*{5mm}
\subfloat[
\small
Imagen]{
\resizebox{.95\textwidth}{!}{
    \begin{tabular}{|p{\dimexpr 0.2\textwidth-2\tabcolsep-1.25\arrayrulewidth\relax}|
                    p{\dimexpr 0.2\textwidth-2\tabcolsep-1.25\arrayrulewidth\relax}|
                    p{\dimexpr 0.2\textwidth-2\tabcolsep-1.25\arrayrulewidth\relax}|
                    p{\dimexpr 0.2\textwidth-2\tabcolsep-1.25\arrayrulewidth\relax}|
                    p{\dimexpr 0.2\textwidth-2\tabcolsep-1.25\arrayrulewidth\relax}|}
        \hline
        kit fox & shower curtain & night snake & hot tub & minivan \\
\hline
desktop computer & keyboard space bar & European green lizard & espresso machine & black-footed ferret \\
\hline
water jug & flute & velvet fabric & mobile phone & digital clock \\
\hline
product packet / packaging & CRT monitor & eastern hog-nosed snake & tape player & bolete \\
\hline
tobacco shop & monastery & purse & mushroom & printer \\
\hline
letter opener & wall clock & toilet paper & monitor & sunglasses \\
\hline
overskirt & hard disk drive & ladle & can opener & tiger cat \\
\hline
combination lock & paper towel & plunger & tights & vine snake \\
\hline
\end{tabular}
}
}

\vspace*{5mm}
\subfloat[
\small
Muse]{
\resizebox{.95\textwidth}{!}{
    \begin{tabular}{|p{\dimexpr 0.2\textwidth-2\tabcolsep-1.25\arrayrulewidth\relax}|
                    p{\dimexpr 0.2\textwidth-2\tabcolsep-1.25\arrayrulewidth\relax}|
                    p{\dimexpr 0.2\textwidth-2\tabcolsep-1.25\arrayrulewidth\relax}|
                    p{\dimexpr 0.2\textwidth-2\tabcolsep-1.25\arrayrulewidth\relax}|
                    p{\dimexpr 0.2\textwidth-2\tabcolsep-1.25\arrayrulewidth\relax}|}
\hline
titi monkey & alligator lizard & European green lizard & cottontail rabbit & African rock python \\
\hline
stopwatch & gar fish & Irish Water Spaniel & European polecat & CRT monitor \\
\hline
toy terrier & keyboard space bar & night snake & Norfolk Terrier & Ibizan Hound \\
\hline
mobile phone & ground beetle & Tibetan Terrier & Norwich Terrier & purse \\
\hline
Treeing Walker Coonhound & Siberian Husky & eastern hog-nosed snake & Bouvier des Flandres dog & patas monkey \\
\hline
Australian Terrier & CD player & Briard & Affenpinscher & English Setter \\
\hline
cradle & red wolf or maned wolf & Geoffroy's spider monkey & Border Terrier & Lakeland Terrier \\
\hline
tape player & Cairn Terrier & Bluetick Coonhound & Entlebucher Sennenhund & Redbone Coonhound \\
\hline
\end{tabular}
}
}
\end{table*}

In Table~\ref{table:appendix-poor-clases}, we identify and list `poor' classes where supervised models, trained on synthetic images, face challenges in accurate classification. For each of the three text-to-image models — Stable Diffusion, Imagen, and Muse — we highlight 40 distinct categories that pose difficulties. Notably, certain categories, such as tiger cats and vine snakes, are common challenges across different text-to-image models. Future research in the development of text-to-image models could benefit from focusing on these categories. Improving the accuracy in generating images of these `poor' classes is crucial, as their current limitations are a key factor hindering the ability of synthetic images to have better scaling ability and performance than real images, in  the supervised learning contexts.

\begin{table*}[t]
\begin{center}
\caption{\small{Comparison of CLIP models trained on LAION-400M subsets of different data scales, using different hyper-parameter configurations. Hyper-parameter configuration `S' and `L' corresponds to (a) and (b) in Table~\ref{table:appendix-clip-hyperparam}, respectively. All models are trained on real images and text only, using ViT-B/32 as backbone architecture.
}}
\label{table:appendix-clip-hyper-choose}
\resizebox{\textwidth}{!}{
\begin{tabular}
{c@{\hspace{2.0em}}c@{\hspace{1.0em}}|@{\hspace{1.0em}}ccccccccccccccc|c|c}
\toprule[1.2pt]
\bf Scale&\rotatebox[origin=lb]{90}{\smash{\small Hyper-param}}&
\rotatebox[origin=lb]{90}{\smash{\small Food-101}} & \rotatebox[origin=lb]{90}{\smash{\small CIFAR-10}} & \rotatebox[origin=lb]{90}{\smash{\small CIFAR-100}} & \rotatebox[origin=lb]{90}{\smash{\small SUN397}} &
\rotatebox[origin=lb]{90}{\smash{\small Cars}} & \rotatebox[origin=lb]{90}{\smash{\small Aircraft}} & \rotatebox[origin=lb]{90}{\smash{\small DTD}} & \rotatebox[origin=lb]{90}{\smash{\small Pets}} & \rotatebox[origin=lb]{90}{\smash{\small Caltech-101}} &
\rotatebox[origin=lb]{90}{\smash{\small Flowers}} & \rotatebox[origin=lb]{90}{\smash{\small STL-10}} & \rotatebox[origin=lb]{90}{\smash{\small EuroSAT}} &
\rotatebox[origin=lb]{90}{\smash{\small RESISC45}} & \rotatebox[origin=lb]{90}{\smash{\small GTSRB}} & \rotatebox[origin=lb]{90}{\smash{\small Country211}}  & \rotatebox[origin=lb]{90}{\smash{\small \bf Average}} & \rotatebox[origin=lb]{90}{\smash{\small \bf ImageNet}}\\
\midrule
\multirow{2}{1.3em}{\rotatebox[origin=c]{0}{\small{10M}}}   & \small \bf S & 42.0 & 78.3 & 49.0 & 36.1 & 40.3 & 3.6 & 22.2 & 47.4 & 75.0 & 29.9 & 85.0 & 27.5 & 30.4 & 19.6 & 4.4 & \textbf{39.4} & \textbf{30.6} \\
& \small \bf L & 15.8 & 51.0 & 22.7 & 16.0 & 9.9 & 0.8 & 10.2 & 19.7 & 50.2 & 14.9 & 63.2 & 12.5 & 16.5 & 6.0 & 1.9 & 20.8 & 14.8 \\
\midrule
\multirow{2}{1.3em}{\rotatebox[origin=c]{0}{\small{50M}}}   & \small \bf S & 63.9 & 87.7 & 65.4 & 54.5 & 61.6 & 5.9 & 34.4 & 71.2 & 84.0 & 45.3 & 93.4 & 46.4 & 45.3 & 28.1 & 8.4 & \textbf{53.0} & \textbf{47.9} \\
& \small \bf L & 62.3 & 84.2 & 59.9 & 48.9 & 62.1 & 4.9 & 31.5 & 71.3 & 83.2 & 47.1 & 90.6 & 30.9 & 39.5 & 30.0 & 7.7 & 50.3 & 44.2 \\ 
\midrule
\multirow{2}{1.3em}{\rotatebox[origin=c]{0}{\small{100M}}}  & \small \bf S  & 69.5 & 88.9 & 68.6 & 58.4 & 67.3 & 6.9 & 43.4 & 75.4 & 85.7 & 49.9 & 93.3 & 46.9 & 54.1 & 41.1 & 9.6 & 57.3 & 51.8 \\
& \small \bf L & 72.6 & 89.4 & 68.0 & 57.6 & 72.6 & 7.1 & 41.0 & 80.9 & 87.3 & 55.9 & 93.8 & 36.4 & 52.3 & 41.5 & 10.4 & \textbf{57.8} & \textbf{54.2} \\

\bottomrule[1.2pt]
\end{tabular}}
\end{center}
\end{table*}

\begin{table*}[t]
\begin{center}
\caption{\small{Comparison of CLIP models trained on synthetic CC12M generated by Stable Diffusion with different CFG scales. Models are trained using ViT-B/16 as backbone architecture.
}}
\label{table:appendix-clip-cfg}
\resizebox{\textwidth}{!}{
\begin{tabular}
{c|ccccccccccccccc|c|c}
\toprule[1.2pt]
\bf CFG &
\rotatebox[origin=lb]{90}{\smash{\small Food-101}} & \rotatebox[origin=lb]{90}{\smash{\small CIFAR-10}} & \rotatebox[origin=lb]{90}{\smash{\small CIFAR-100}} & \rotatebox[origin=lb]{90}{\smash{\small SUN397}} &
\rotatebox[origin=lb]{90}{\smash{\small Cars}} & \rotatebox[origin=lb]{90}{\smash{\small Aircraft}} & \rotatebox[origin=lb]{90}{\smash{\small DTD}} & \rotatebox[origin=lb]{90}{\smash{\small Pets}} & \rotatebox[origin=lb]{90}{\smash{\small Caltech-101}} &
\rotatebox[origin=lb]{90}{\smash{\small Flowers}} & \rotatebox[origin=lb]{90}{\smash{\small STL-10}} & \rotatebox[origin=lb]{90}{\smash{\small EuroSAT}} &
\rotatebox[origin=lb]{90}{\smash{\small RESISC45}} & \rotatebox[origin=lb]{90}{\smash{\small GTSRB}} & \rotatebox[origin=lb]{90}{\smash{\small Country211}}  & \rotatebox[origin=lb]{90}{\smash{\small \bf Average}} & \rotatebox[origin=lb]{90}{\smash{\small \bf ImageNet}}\\
\midrule
1.25 & 41.9 & 37.8 & 16.3 & 39.5 & 22.8 & 3.2 & 20.9 & 56.3 & 70.9 & 22.4 & 83.8 & 11.7 & 31.5 & 7.2 & 4.7 & 31.4 & 34.9 \\ \midrule
1.5 & 43.1 & 32.8 & 16.6 & 42.4 & 26.6 & 3.7 & 23.4 & 58.8 & 68.8 & 24.3 & 86.4 & 10.8 & 33.3 & 7.4 & 5.6 & 32.3 & \textbf{35.7} \\\midrule
1.75 & 43.5 & 29.5 & 13.3 & 42.7 & 28.0 & 3.9 & 22.2 & 58.1 & 70.6 & 21.8 & 84.2 & 17.7 & 32.0 & 8.3 & 5.5 & 32.1 & 35.6 \\\midrule
2.5 & 41.3 & 47.0 & 15.1 & 41.9 & 28.8 & 4.7 & 23.9 & 57.8 & 71.2 & 24.4 & 87.4 & 16.8 & 31.0 & 6.1 & 5.7 & 33.5 & 34.3\\
\bottomrule[1.2pt]
\end{tabular}}
\end{center}
\end{table*}

\subsection{Per-class FID and LPIPS}
Following the same setup outlined in Section~\ref{sec:affect-scaling-ability} of the main paper, we also computed the correlations between scaling ability ($k$ in Equation~\ref{equ:scaling} of the main paper) and both FID and LPIPS scores. Unlike the previous analysis focusing on recognizability and diversity, this evaluation specifically studies the relationship of scaling ability with these two metrics.

\begin{figure}[h]
\centering
\includegraphics[width=1.0\linewidth]{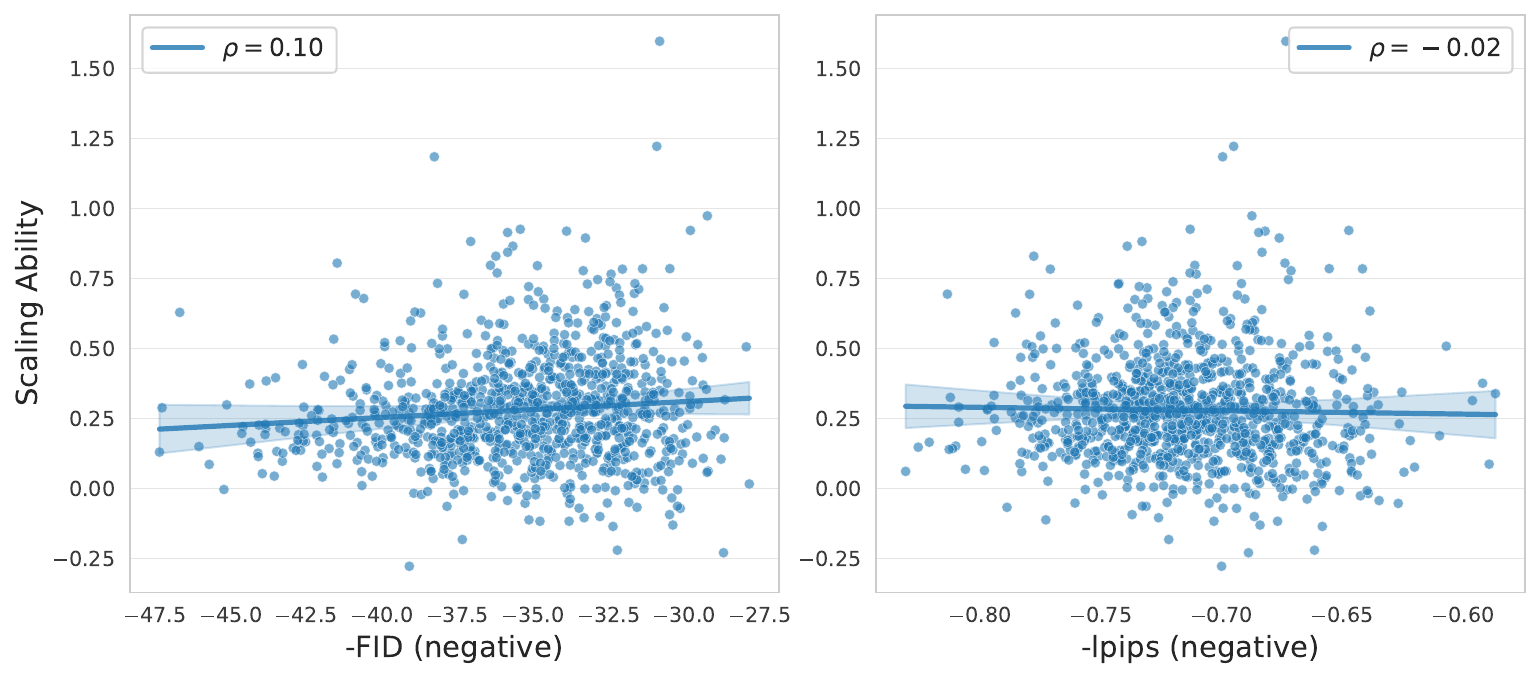}
\caption{
\small
Per class analysis on the relationship between scaling ability (defined as $k$ in Equation~\ref{equ:scaling} in the main paper) and both FID and LPIPS. Within each specific class, the plots indicate the correlation between the scaling ability and both metrics appears to be negligible.}
\label{fig:scaling_fid_lpips}
\end{figure}

To calculate the per-class LPIPS scores, we used the same method as detailed previously. However, for per-class FID computation, the existing synthetic test sets, containing only 50 images per class, were deemed insufficient, since FID score is sensitive to the number of images. Therefore we sample $1300$ images from the synthetic training images and compute the FID with images from real ImageNet training set for each class. Similar to our previous approach, we took the negative of the per-class FID and LPIPS scores for consistency, as lower scores indicate better performance. The correlations obtained are depicted in Figure~\ref{fig:scaling_fid_lpips}.

The results from this figure indicate a lack of strong correlation between scaling ability and either FID or LPIPS scores. This finding highlights the necessity for a more tailored metric that is specifically designed to assess the scaling ability of supervised classifiers trained on synthetic images.
\begin{table*}[t]
\begin{center}
\caption{\small{Zero-shot transfer performance on 15 downstream datasets and ImageNet. Models are trained on LAION-400M subsets with images from synthetic, real or synthetic+real. Dataset scale starts from 1M and increase exponentially. Combining synthetic images and real images can improve zero-shot classification performance under various cases, especially when data amount is limited.
}}
\vspace*{-2mm}
\label{table:appendix-clip-all}
\resizebox{\textwidth}{!}{
\begin{tabular}
{ccccccccccccccccc@{\hspace{1.0em}}|@{\hspace{1.0em}}c@{\hspace{1.0em}}|@{\hspace{1.0em}}c@{\hspace{1.0em}}}
\toprule[1.2pt]
\bf Scale&\bf Data&
\rotatebox[origin=lb]{90}{\smash{\small Food-101}} & \rotatebox[origin=lb]{90}{\smash{\small CIFAR-10}} & \rotatebox[origin=lb]{90}{\smash{\small CIFAR-100}} & \rotatebox[origin=lb]{90}{\smash{\small SUN397}} &
\rotatebox[origin=lb]{90}{\smash{\small Cars}} & \rotatebox[origin=lb]{90}{\smash{\small Aircraft}} & \rotatebox[origin=lb]{90}{\smash{\small DTD}} & \rotatebox[origin=lb]{90}{\smash{\small Pets}} & \rotatebox[origin=lb]{90}{\smash{\small Caltech-101}} &
\rotatebox[origin=lb]{90}{\smash{\small Flowers}} & \rotatebox[origin=lb]{90}{\smash{\small STL-10}} & \rotatebox[origin=lb]{90}{\smash{\small EuroSAT}} &
\rotatebox[origin=lb]{90}{\smash{\small RESISC45}} & \rotatebox[origin=lb]{90}{\smash{\small GTSRB}} & \rotatebox[origin=lb]{90}{\smash{\small Country211}}  & \rotatebox[origin=lb]{90}{\smash{\small \bf Average}}& \rotatebox[origin=lb]{90}{\smash{\small \bf ImageNet}}\\
\midrule
\multirow{3}{1.3em}{\rotatebox[origin=c]{0}{\small{1M}}}   & \small Syn & 5.2 & 12.8 & 3.3 & 5.9 & 1.7 & 0.9 & 5.5 & 6.7 & 17.8 & 3.5 & 29.4 & 9.0 & 9.7 & 5.4 & 1.2 & 7.9 & 4.1 \\
 & \small Real & 5.2 & 25.4 & 7.6 & 5.0 & 2.1 & 1.0 & 5.4 & 5.4 & 18.0 & 5.0 & 36.4 & 14.7 & 9.3 & 6.6 & 1.0 & 9.9 & 3.8 \\
 &\small Syn+Real & \textbf{10.9} & \textbf{32.2} & \textbf{13.0} & \textbf{13.1} & \textbf{4.6} & \textbf{1.4} & \textbf{9.4} & \textbf{12.0} & \textbf{36.0} & \textbf{8.9} & \textbf{62.5} & \textbf{19.8} & \textbf{14.7} & \textbf{7.5} & \textbf{1.9} & \textbf{16.5} & \textbf{9.4} \\ \midrule

\multirow{3}{1.3em}{\rotatebox[origin=c]{0}{\small{2M}}}   & \small Syn &11.0 & 15.3 & 3.8 & 14.5 & 6.2 & 1.7 & 10.3 & 15.6 & 36.2 & 7.2 & 36.3 & 15.6 & 14.5 & 3.4 & 1.7 & 12.9 & 10.7 \\
 & \small Real & 13.4 & 39.0 & 16.8 & 13 & 6.6 & 1.3 & 10.5 & 13.0 & 40.1 & 12.4 & 57.1 & \textbf{17.0} & 14.9 & \textbf{6.5} & 1.7 & 17.6 & 10.6 \\
 &\small Syn+Real & \textbf{22.2} & \textbf{59.7} & \textbf{27.0} & \textbf{23.4} & \textbf{18.2} & \textbf{2.1} & \textbf{15.4} & \textbf{24.8} & \textbf{55.7} & \textbf{13.1} & \textbf{73.9} & \textbf{22.4} & \textbf{20.6} & \textbf{4.2} & \textbf{3.0} & \textbf{25.7} & \textbf{19.8} \\ \midrule
 
 \multirow{3}{1.3em}{\rotatebox[origin=c]{0}{\small{4M}}}   & \small Syn & 19.6 & 19.7 & 7.1 & 23.0 & 22.4 & 2.1 & 17.0 & 30.1 & 53.4 & 13.9 & 64.4 & 12.8 & 21.1 & 5.3 & 3.1 & 21.0 & 19.8 \\
 & \small Real & 30.8 & 63.8 & 33.4 & 26.2 & 27.7 & 1.8 & 18.8 & 33.9 & 61.7 & 18.9 & 79.2 & \textbf{40.2} & 21.5 & 10.0 & 3.4 & 31.4 & 21.7 \\
 &\small Syn+Real & \textbf{40.3} & \textbf{67.1} & \textbf{39.3} & \textbf{35.8} & \textbf{40.9} & \textbf{2.3} & \textbf{22.9} & \textbf{45.4} & \textbf{70.1} & \textbf{23.1} & \textbf{88.2} & 33.5 & \textbf{27.7} & \textbf{12.3} & \textbf{4.5} & \textbf{36.9} & \textbf{30.6}
 \\ \midrule
 \multirow{3}{1.3em}{\rotatebox[origin=c]{0}{\small{8M}}} &\small Syn & 34.2 & 23.8 & 9.5 & 32.6 & 39.9 & 3.5 & 20.3 & 46.3 & 63.0 & 20.7 & 78.7 & 9.8 & 19.1 & 4.9 & 4.3 & 27.4 & 29.0 \\
 &\small Real & 48.7 & 79.6 & 47.9 & 38.5 & 48.9 & 3.9 & 25.5 & 52.8 & 74.9 & \textbf{31.2} & 88.0 & 27.4 & 32.8 & \textbf{16.7} & 5.2 & 41.5 & 34.2 \\
 &\small Syn+Real & \textbf{54.5} & \textbf{82.9} & \textbf{53.1} & \textbf{46.3} & \textbf{57.3} & \textbf{5.1} & \textbf{29.6} & \textbf{61.4} & \textbf{78.2} & 31.1 & \textbf{92.5} & \textbf{29.6} & \textbf{41.1} & 14.5 & \textbf{6.5} & \textbf{45.6} &\textbf{40.7} \\ \midrule

\multirow{3}{1.8em}{\rotatebox[origin=c]{0}{\small{16M}}}   & \small Syn &44.2 & 32.4 & 11.5 & 41.6 & 51.3 & 4.9 & 27.4 & 58.3 & 72.1 & 24.8 & 83.6 & 16.7 & 29.5 & 4.6 & 5.9 & 33.9  & 37.7\\
 & \small Real & 62.9 & 85.2 & 58.1 & 49.0 & 60.6 & \textbf{5.0} & 30.4 & 61.9 & 81.5 & \textbf{40.9} & 93.1 & \textbf{43.2} & 39.4 & \textbf{28.0} & 7.4 & 49.8 & 43.8 \\
 &\small Syn+Real & \textbf{64.8} & \textbf{87.5} & \textbf{61.0} & \textbf{53.7} & \textbf{63.3} & \textbf{4.9} & \textbf{36.5} & \textbf{67.7} & \textbf{82.8} & \textbf{38.6} & \textbf{94.5} & \textbf{37.6} & \textbf{48.2} & \textbf{28.6} & \textbf{8.2} & \textbf{51.9}& \textbf{48.2}\\ \midrule

\multirow{3}{1.8em}{\rotatebox[origin=c]{0}{\small{32M}}}   & \small Syn & 54.2 & 33.3 & 18.3 & 47.3 & 58.9 & 4.4 & 30.3 & 65.3 & 75.3 & 29.4 & 88.5 & 15.5 & 35.5 & 8.5 & 7.3 & 38.1 & 43.8 \\
 & \small Real & 70.4 & 86.3 & 64.7 & 55.7 & 67.2 & 4.9 & 35.7 & 70.1 & 82.8 & \textbf{46.0} & 94.9 & \textbf{43.4} & 48.6 & \textbf{36.6} & 9.1 & 54.4 & 50.5 \\
 &\small Syn+Real & \textbf{71.4} & \textbf{89.4} & \textbf{65.3} & \textbf{57.9} & \textbf{69.7} & \textbf{6.5} & \textbf{41.8} & \textbf{72.9} & \textbf{83.2} & 41.3 & \textbf{95.2} & 38.7 & \textbf{55.1} & 29.9 & \textbf{10.2} & \textbf{55.2} & \textbf{52.9}\\ \midrule

\multirow{3}{1.8em}{\rotatebox[origin=c]{0}{\small{64M}}}   & \small Syn & 59.7 & 44.1 & 20.9 & 51.5 & 62.7 & \textbf{7.7} & 37.9 & 71.1 & 79.2 & 35.8 & 92.1 & 15.1 & 39.0 & 12.5 & 8.6 & 42.5 & 48.0 \\
 & \small Real & 75.2 & \textbf{90.9} & \textbf{69.5} & 59.4 & 71.5 & 7.3 & 42.8 & 75.0 & 87.0 & \textbf{50.6} & 95.7 & \textbf{46.8} & 51.8 & \textbf{39.8} & 11.2 & 58.3 & 55.1 \\
 &\small Syn+Real & \textbf{74.6} & 90.8 & 67.8 & \textbf{61.1} & \textbf{73.3} & 6.3 & \textbf{50.2} & \textbf{76.3} & \textbf{87.8} & 47.6 & \textbf{95.8} & 45.5 & \textbf{58.3} & 37.5 & \textbf{11.6} & \textbf{59.0} & \textbf{56.4}\\ \midrule

\multirow{3}{2.3em}{\rotatebox[origin=c]{0}{\small{128M}}} &\small Syn & 63.7 & 45.1 & 15.9 & 52.3 & 67.1 & 9.3 & 37.8 & 75.7 & 80.5 & 39.1 & 93.2 & 8.0 & 35.7 & 10.1 & 9.5 & 42.9 & 51.2 \\
 &\small Real & \textbf{81.9} & 90.5 & \textbf{70.9} & 62.5 & 78.7 & 10.7 & 46.0 & \textbf{85.9} & 88.7 & \textbf{60.4} & 96.0 & \textbf{48.3} & 57.8 & 42.7 & \textbf{14.2} & 62.3 & 61.4 \\
 &\small Syn+Real & 81.6 & \textbf{91.0} & 70.4 & \textbf{64.0} & \textbf{79.4} & \textbf{11.9} & \textbf{52.5} & 85.1 & \textbf{90.2} & 59.5 & \textbf{97.0} & 47.3 & \textbf{61.1} & \textbf{45.3} & 14.1 & \textbf{63.4}& \textbf{62.9} \\ \midrule

\multirow{3}{2.3em}{\rotatebox[origin=c]{0}{\small{256M}}}   & \small Syn & 68.6 & 46.2 & 21.8 & 54.7 & 70.4 & 10.9 & 42.9 & 80.2 & 81.5 & 44.6 & 95.2 & 20.2 & 39.1 & 12.8 & 10.5 & 46.6 & 54.4 \\
 & \small Real & \textbf{84.6} & \textbf{92.8} & \textbf{73.5} & \textbf{66.5} & \textbf{82.4} & 12.3 & 52.7 & \textbf{89.9} & 91.3 & \textbf{65.7} & 96.9 & \textbf{39.2} & 64.4 & \textbf{47.3} & \textbf{16.9} & \textbf{65.1} & \textbf{65.4} \\
 &\small Syn+Real &\textbf{83.8} & \textbf{92.4} & \textbf{73.3} & \textbf{66.0} & \textbf{82.3} & \textbf{14.6} & \textbf{55.0} & \textbf{86.7} & \textbf{91.4} & 58.6 & \textbf{97.8} & \textbf{47.7} & \textbf{65.2} & 42.5 & \textbf{15.3} & 64.8 & \textbf{65.4} \\ \midrule

 \multirow{3}{2.3em}{\rotatebox[origin=c]{0}{\small{371M}}} &\small Syn & 70.1 & 51.9 & 26.2 & 55.5 & 70.8 & 12.3 & 41.5 & 79.6 & 83.6 & 45.5 & 95.7 & 28.8 & 39.3 & 20.6 & 10.9 & 48.8 & 55.7 \\
 &\small Real & \textbf{85.7} & \textbf{93.9} & \textbf{75.6} & \textbf{67.5} & \textbf{83.3} & 14.2 & 50.1 & \textbf{88.8} & 91.1 & \textbf{67.0} & 97.0 & 43.9 & \textbf{66.6} & 42.8 & \textbf{17.5} & 65.7 &\textbf{66.8} \\
 &\small Syn+Real & 84.6 & 92.4 & 73.2 & 67.1 & 82.0 & \textbf{17.2} & \textbf{56.8} & 86.4 & \textbf{91.7} & 61.6 & \textbf{97.3} & \textbf{52.2} & 65.9 & \textbf{46.7} & 16.0 & \textbf{66.1} & 66.6 \\

\bottomrule[1.2pt]
\end{tabular}}
\end{center}
\vspace{-4mm}
\end{table*}

\section{More Results on CLIP Scaling}
\label{sec:appendix-clip}
\subsection{Comparison on Hyper-parameters}
\label{sec:appendix-clip-hyperselect}
In Table~\ref{table:appendix-clip-hyperparam}, we detail the use of two distinct sets of hyper-parameters for CLIP training, tailored to different dataset scales. Config (a) in the table, labeled as `S' here, is designed for smaller dataset scales with fewer than 100 million images. Conversely, config (b), represented as `L' here, is intended for larger dataset scales equal to or exceeding 100 million images. To validate the necessity of these configurations, we present an empirical study in Table~\ref{table:appendix-clip-hyper-choose}.

Here we train CLIP models on subsets of the LAION-400M dataset with 10M, 50M, and 100M samples, exclusively utilizing real images and applying the two different hyper-parameter sets. Our findings indicate that for scales of 10M and 50M, the `S' hyper-parameter configuration yields superior results, with the performance difference being reduced at the 50M scale. In contrast, at the 100M scale, the `L' configuration demonstrates enhanced performance. Therefore, based on these empirical results, we opted to utilize the `S' hyper-parameter set for smaller data scales and the `L' set for larger scales.

\subsection{Comparison on different CFGs}
\label{sec:appendix-clip-cfg}
To identify the most effective CFG scale for generating synthetic images to train CLIP models, we utilized the Stable Diffusion to create synthetic images for the CC12M dataset~\cite{changpinyo2021conceptual} at four different CFG scales: 1.25, 1.5, 1.75, and 2.5. Following the generation of these images, CLIP models were trained using the synthetic images and their corresponding texts. The efficacy of these trained models was then evaluated through zero-shot classification on ImageNet and various downstream classification datasets.

The detailed comparison of these different CFG scales are presented in Table~\ref{table:appendix-clip-cfg}. Based on these results, we determined that a CFG scale of 1.5 delivers the best zero-shot classification performance on ImageNet. Consequently, we chose CFG$=1.5$ for the majority of our CLIP experiments.

\begin{figure}[t]
\centering
\includegraphics[width=1.0\linewidth]{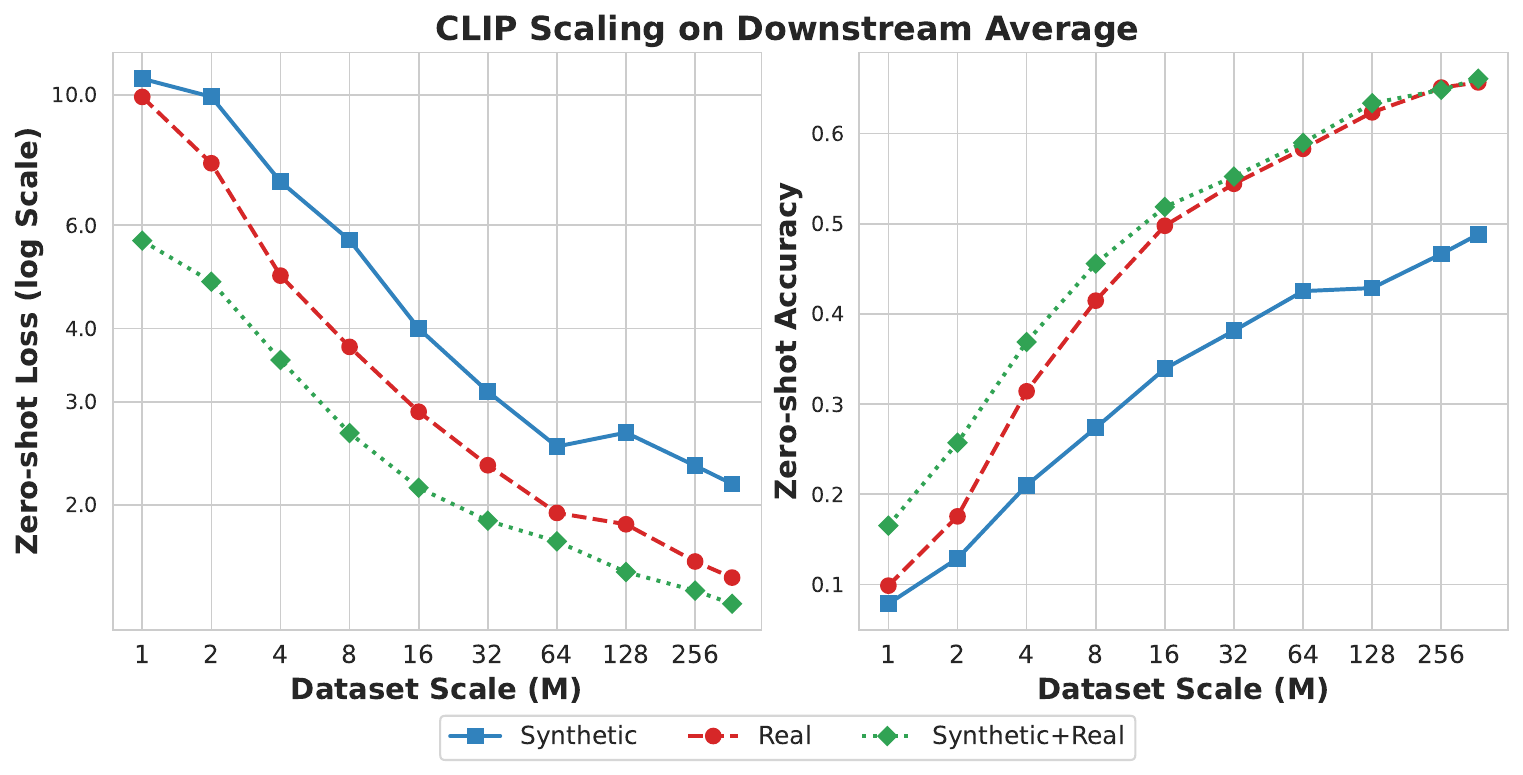}
\caption{
\small
The average scaling behavior on zero-shot classification for CLIP models over all 15 downstream datasets. Models are trained on LAION-400M subsets with synthetic, real, or synthetic+real images.}
\label{fig:appendix-clip_scaling_ds_avg}
\end{figure}
\begin{figure*}[h]
\centering
\includegraphics[width=1.0\linewidth]{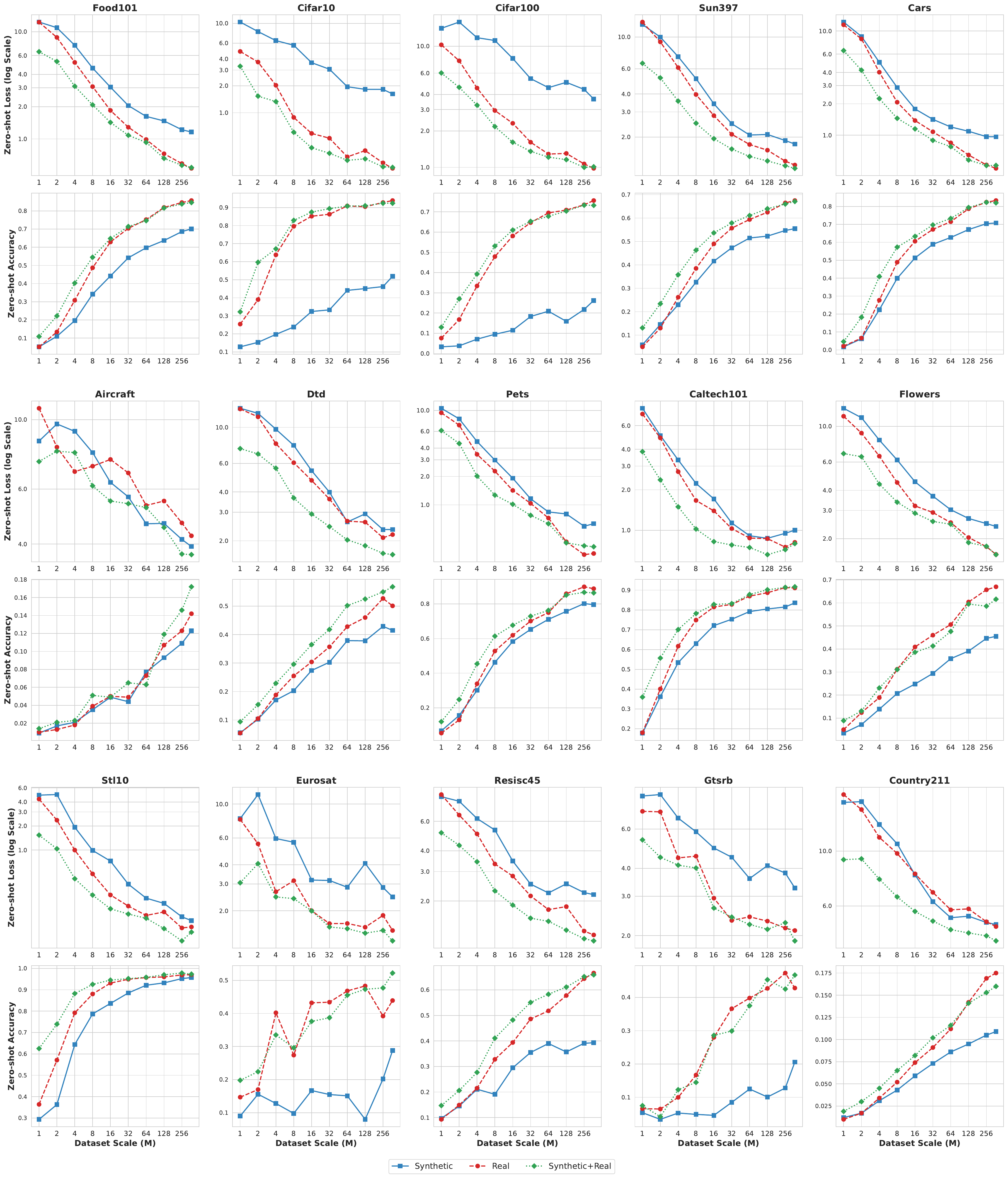}
\caption{
\small
Detailed scaling behavior comparison between CLIP models trained on synthetic, real, or the combination of synthetic and real images, on 15 different downstream tasks. The models are evaluated under zero-shot classification.}
\label{fig:appendix-clipscalingds}
\vspace{-5mm}
\end{figure*}
\subsection{Detailed experiment results for all scales}
Table~\ref{table:appendix-clip-all} provides detailed scaling behavior for CLIP models trained utilizing either synthetic, real, or a combination of synthetic and real images. We also present the scaling behavior comparison in detailed plots for each specific downstream dataset in Figure~\ref{fig:appendix-clipscalingds}. Figure~\ref{fig:appendix-clip_scaling_ds_avg} shows the average scaling behavior over all 15 downstream datasets. The models were trained on subsets of the LAION-400M dataset, beginning with 1 million samples and scaling up exponentially to the entire set of 371M.  Our findings indicate synthetic images does not scale as good as real onees,  yet integrating synthetic images with real images in the training of CLIP models can be advantageous, particularly in scenarios where the dataset size is relatively limited.
\end{appendices}

\end{document}